\documentclass[preprint,12pt,authoryear]{elsarticle}

\usepackage{amssymb}
\usepackage{pdflscape}
\usepackage[colorlinks=true, linkcolor=blue, citecolor=blue, urlcolor=blue]{hyperref}
\usepackage{booktabs}  
\usepackage{amsmath}
\usepackage{caption}
\usepackage[ruled,vlined]{algorithm2e}
\usepackage{placeins}
\usepackage{microtype}
\usepackage{subcaption}
\usepackage{multirow}
\usepackage[table]{xcolor}
\usepackage{float}
\usepackage[english]{babel}
\usepackage{hyperref}

\journal{Nuclear Physics B}

\begin{document}

\hypersetup{pageanchor=false}

\begin{frontmatter}

  \title{AgriKD: Cross-Architecture Knowledge Distillation for Efficient Leaf Disease Classification}

  \author{Minh-Dung Le}
  \author{Minh-Duc Hoang}
  \author{Hoang-Vu Truong}

  \author{Thi-Thu-Hong Phan\corref{cor1}}

  \affiliation{organization={AIT laboratory, Faculty  of Artificial Intelligence, \\ FPT University},
    city={Da Nang},
    postcode={550000},
  country={Viet Nam}}

  \cortext[cor1]{Corresponding author}

  \ead{hongptt11@fe.edu.vn}
  \author{}

  \begin{abstract}
    Automated leaf disease classification plays a critical role in precision agriculture, particularly in resource-constrained environments requiring efficient real-time inference. While Vision Transformers (ViTs) demonstrate strong performance by modeling global dependencies, their high computational and memory demands limit practical deployment. Existing approaches, therefore, struggle to effectively transfer transformer-derived representations to lightweight models.
    In this paper, we propose AgriKD, a cross-architecture knowledge distillation framework for efficient edge deployment that transfers knowledge from a ViT teacher to a compact convolutional student. To bridge the representational gap between Transformer and CNN architectures, AgriKD aligns transformer representations and attention-derived features with convolutional feature maps through a multi-level distillation strategy integrating output-level, feature-level, and relational supervision.
    Experiments on multiple leaf disease datasets show that the distilled student achieves performance comparable to the teacher, while significantly improving efficiency. Specifically, the proposed framework reduces model parameters by approximately 172 times, computational cost by 47.57 times, and inference latency by 18--22 times.
    Furthermore, the optimized model is deployed across multiple runtime formats, including ONNX, TFLite Float16, and TensorRT FP16, achieving consistent predictive performance with negligible accuracy degradation. Real-world deployment on NVIDIA Jetson edge devices and a mobile application demonstrates reliable real-time inference, highlighting the practicality of AgriKD for AI-powered agricultural applications in resource-constrained environments.
  \end{abstract}


\end{frontmatter}

\hypersetup{pageanchor=true}

\section{Introduction}\label{sec:introduction}

Leaf diseases pose a significant threat to agricultural productivity, and early detection is essential to reduce crop losses and minimize excessive pesticide use. In practice, automated leaf disease recognition systems must balance classification accuracy with deployment efficiency, as they are often required to operate on resource-constrained edge devices.

Deep learning has significantly advanced leaf disease recognition. Convolutional neural networks (CNNs) have achieved strong performance across agricultural datasets~\citep{abade2021plant,dhaka2021survey,anis2026survey}, while ViTs have demonstrated superior capability in modeling long-range dependencies through self-attention, enabling richer global representations~\citep{dosovitskiy2020image,borhani2022deep,thai2023formerleaf}. However, both high-capacity CNNs and ViTs are computationally expensive for edge deployment, whereas lightweight models such as MobileNetV2 improve efficiency at the cost of reduced representational capacity.

Knowledge distillation (KD) provides a practical solution by transferring knowledge from a high-capacity teacher to a compact student~\citep{hinton2015distilling}. In particular, transformer-based models are attractive teachers due to their ability to encode global contextual information through mechanisms such as self-attention and class tokens. This makes cross-architecture distillation from ViT teachers to CNN students a promising approach for balancing accuracy and efficiency.

Despite these advantages, existing approaches have not fully exploited the representational strengths of transformer-based teachers. While ViTs encode rich global information, current distillation methods often fail to effectively transfer these representations to lightweight CNN students. In many cases, knowledge transfer relies on limited supervision, such as logits or attention alone, which is insufficient to capture the full spectrum of transformer knowledge.

To address this limitation, we propose AgriKD, a cross-architecture knowledge distillation framework that enhances the transfer of transformer representations to lightweight CNN models. The proposed approach integrates multiple distillation objectives, where each objective captures a different aspect of the teacher knowledge, including output distributions, feature representations, and relational structure. This enables more effective transfer of transformer-derived representations while maintaining deployment efficiency.

The main contributions of this work are summarized as follows:
\begin{enumerate}
  \item We propose a cross-architecture distillation framework that transfers knowledge from a ViT-Base teacher to a lightweight MobileNetV2 student with explicit alignment between transformer tokens and convolutional feature maps.
  \item We design a multi-level distillation strategy that combines output-level, feature-level, and relational supervision to better preserve transformer-derived representations.
  \item We conduct comprehensive experiments across multiple leaf disease datasets to validate the effectiveness and robustness of the proposed approach.
  \item We deploy the proposed model in real-world edge environments and evaluate its efficiency, inference latency, and runtime consistency across platforms such as TensorRT and TFLite.
\end{enumerate}

The remainder of this paper is organized as follows. Section~\ref{sec:related} reviews related work. Section~\ref{sec:method} presents the proposed AgriKD framework. Section~\ref{sec:experiments} describes the datasets and experimental setup, followed by the results and analysis in Section~\ref{sec:results}. Section~\ref{sec:Comparative} provides comparisons with existing studies, and Section~\ref{sec:deployment} demonstrates real-world deployment. Section~\ref{sec:discussion} discusses key findings and limitations, and Section~\ref{sec:conclusion} concludes the paper.

\section{Related Work}\label{sec:related}
\subsection{Deep Learning for Leaf Disease Recognition}

Deep learning has significantly advanced automated leaf disease recognition, particularly through convolutional architectures that learn discriminative lesion, texture, and color patterns from images. Survey studies report that CNN-based pipelines remain the dominant paradigm in agricultural computer vision, providing strong accuracy and adaptability through transfer learning, even with limited annotated data~\citep{abade2021plant,dhaka2021survey,anis2026survey}.
More recently, hybrid CNN--Transformer architectures have been introduced to enhance representation capability by combining local feature extraction with global contextual modeling. While these models improve classification performance, their increased computational complexity limits their applicability in real-world deployment scenarios~\citep{mehdipour2026novel}. Consequently, deployment-oriented factors such as inference latency, memory footprint, and model size remain insufficiently addressed in existing studies.

\subsection{Lightweight CNNs for Edge Deployment}

Lightweight CNNs play a key role in practical agricultural AI due to their compatibility with mobile devices and resource-constrained edge platforms. Among them, MobileNetV2 is widely adopted because its depthwise separable convolutions and inverted residual design significantly reduce computational cost while maintaining competitive performance~\citep{sandler2018mobilenetv2}.
Its effectiveness has been demonstrated in leaf disease classification, where transfer learning with MobileNetV2 achieves strong performance even under class imbalance~\citep{jlassi2024potato}. Subsequent studies further enhance its capability through architectural improvements, including optimized bottleneck structures and attention mechanisms tailored for efficient deployment~\citep{lu2023improved,duhan2025rtr_lite_mobilenetv2}.
Despite these advances, lightweight CNNs remain limited in representation capacity, particularly when disease patterns are subtle or data distributions are complex. This limitation motivates the use of knowledge distillation to improve performance while preserving deployment efficiency.

\subsection{Vision Transformers and Cross-Architecture Transfer}

Vision Transformers model long-range dependencies through self-attention over patch sequences, enabling strong global feature representations. In agricultural vision, transformer-based models have demonstrated competitive or superior performance compared with convolutional approaches, particularly when disease patterns are spatially distributed or visually subtle~\citep{dosovitskiy2020image,borhani2022deep,thai2023formerleaf,hossain2024deep,noman2025vix}. These advantages make ViTs effective teacher models in knowledge distillation frameworks. However, their high computational cost limits their direct applicability in resource-constrained deployment scenarios.

Transferring knowledge from a transformer to a CNN introduces a substantial representational mismatch due to differences in architectural inductive biases. While ViTs operate on sequences of patch tokens, CNNs learn hierarchical feature maps with strong locality priors, making direct feature alignment less effective~\citep{hao2023one}. To address this challenge, prior work has explored projection-based and relation-based distillation strategies. Projection-based approaches align intermediate representations through learned mappings, although some methods may weaken spatial correspondence between teacher and student features~\citep{hao2023one}. In contrast, spatially preserving approaches explicitly maintain positional relationships across architectures~\citep{Liu2022CAKD}, while relation-based methods improve stability by preserving correlation structures instead of enforcing exact prediction matching~\citep{Huang2022DIST}. Despite these advances, most existing studies are evaluated on natural-image benchmarks, and their effectiveness for agricultural leaf disease classification under real-world deployment constraints remains insufficiently explored.

\subsection{Knowledge Distillation in Agricultural Vision}

KD remains underexplored in agricultural computer vision, with existing work following two main directions. First, CNN-to-CNN compression has been investigated to improve efficiency while maintaining accuracy. For example, \citet{phan2025multi} proposed a multi-objective distillation framework integrating hard-label, feature-level, response-level, and self-distillation objectives, demonstrating improved performance across agricultural image classification tasks. However, such approaches are inherently constrained by the representational ceiling of convolutional teachers, limiting the potential gains from knowledge transfer.

Second, cross-architecture distillation has recently been explored as a promising direction to transfer knowledge from transformer-based teachers to lightweight CNN students. In this setting, \citet{mugisha2025hybridknowledgetransferattention} distilled a Swin Transformer into MobileNetV3 for tomato disease classification using joint attention and logit-based supervision, demonstrating the feasibility of compressing transformer representations for edge deployment in agricultural IoT systems. Nevertheless, existing approaches typically rely on logits and attention-based supervision, which may be insufficient to fully transfer the global representations learned by transformer models.

Despite these advances, current research has not fully exploited the representational advantages of transformer-based teachers in cross-architecture distillation. Although feature-projection techniques for ViT-to-CNN alignment have been developed in general computer vision~\citep{Ren_2022_CVPR,Liu2022CAKD}, their effectiveness in transferring transformer-style global representations to CNN students in agricultural settings remains limited.
More importantly, existing approaches typically rely on a small number of distillation objectives, and it remains largely unexplored whether combining supervision across output, feature, and relational levels can more effectively bridge the representational gap between transformer and CNN architectures.
Moreover, deployment-oriented evaluation remains limited, particularly in verifying consistency across practical runtime environments.

These limitations highlight the need for a unified framework that can effectively transfer transformer representations across architectures while maintaining deployment efficiency.

\section{Methodology}
\label{sec:method}
\subsection{Problem Formulation}

Let $\mathcal{X}$ denote the input space of leaf images and $\mathcal{Y}=\{1,\dots,C\}$ denote the label space of $C$ leaf disease categories. Given a dataset $\mathcal{D}=\{(x_i,y_i)\}_{i=1}^{N}$, where $x_i \in \mathcal{X}$ and $y_i \in \mathcal{Y}$, our goal is to train a compact student model $f_S$ that remains competitive with a stronger teacher model $f_T$ while being suitable for deployment on resource-constrained devices.

Formally, the student is optimized by minimizing a multi-component distillation objective:

\begin{equation}
  \min_{\theta_S}\;
  \mathcal{L}_{\mathrm{total}}
  \bigl(f_S(x;\theta_S),\, f_T(x),\, y\bigr),
  \label{eq:objective}
\end{equation}
where $\theta_S$ denotes the trainable parameters of the student and $\mathcal{L}_{\mathrm{total}}$ combines the classification loss with multiple distillation terms.

In our setting, the teacher $f_T$ and the student $f_S$ are heterogeneous architectures, with the teacher providing rich, high-capacity representations and the student designed for efficient deployment. For each input image, the teacher provides class-level predictions and intermediate representations, while the student produces class logits and spatial feature maps. The distillation framework is designed to transfer knowledge across these heterogeneous architectures while preserving classification effectiveness and reducing model complexity, computational cost, and inference latency.

In practice, this objective is difficult to optimize because the teacher and the student represent information differently. This motivates the design of the proposed framework to address this mismatch, as described in the following section.

\subsection{Proposed Framework}
\label{sec:architecture}

\subsubsection{Overview}

\begin{landscape}
  \begin{figure}[p]
    \centering
    \vspace*{-2cm}
    \includegraphics[
      width=1.55\textheight,
      height=\textwidth,
      keepaspectratio
    ]{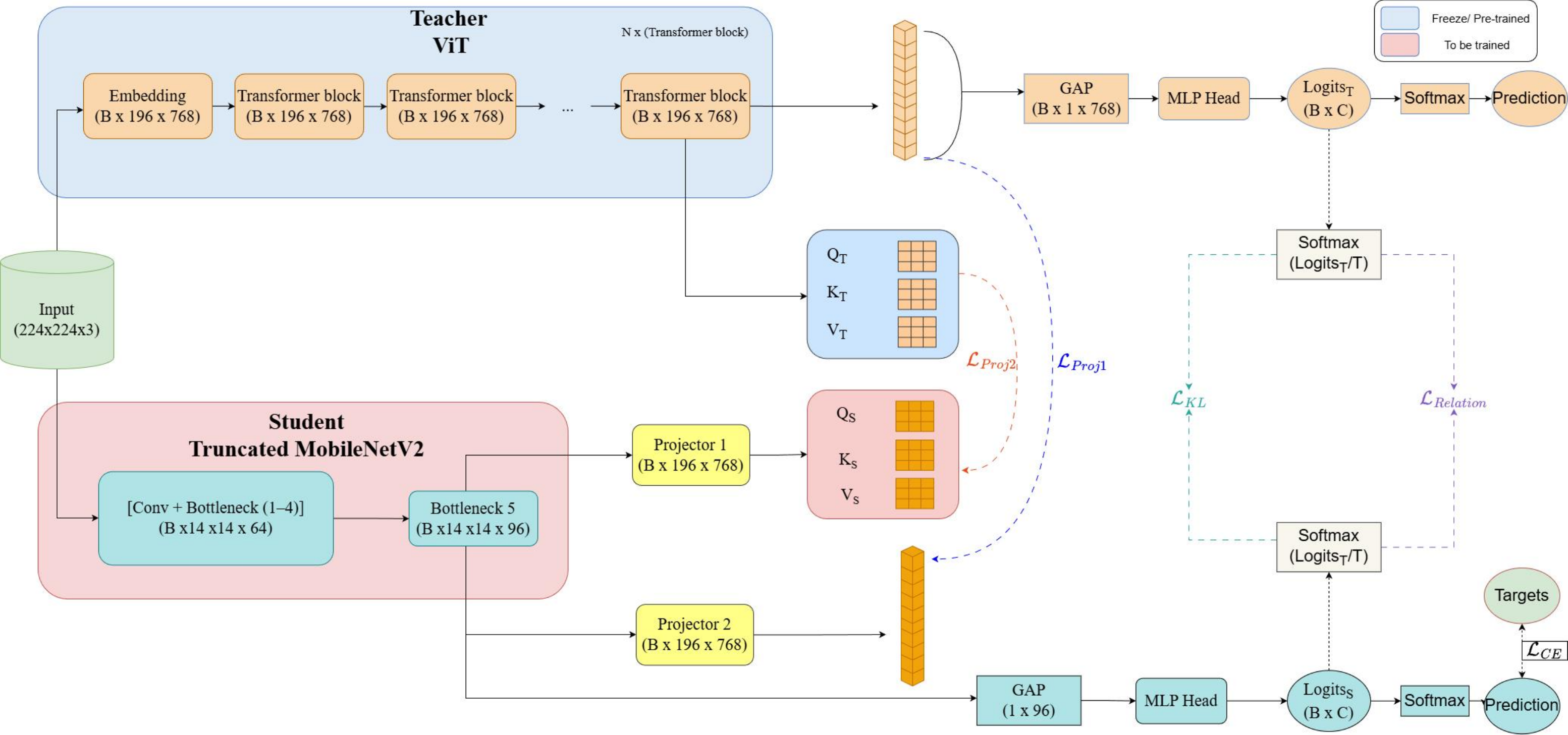}
    \caption{Overview of the proposed AgriKD framework: a pre-trained ViT-Base teacher distills knowledge into a truncated MobileNetV2 student via multi-component distillation}
    \label{fig:overview_pipeline}
  \end{figure}
\end{landscape}

Figure~\ref{fig:overview_pipeline} illustrates the overall pipeline of the proposed framework. The training process follows an offline knowledge distillation setting, where a high-capacity teacher model is first pre-trained and then frozen during the student training phase. Given an input image, the pre-trained teacher generates guidance information, while the student processes the same input to learn from these objectives.

The teacher produces class-level predictions together with intermediate representations, including attention-related outputs and token embeddings, whereas the student generates class logits and spatial feature maps. To enable knowledge transfer across architectures, the student is optimized using a multi-component distillation objective that operates at multiple levels.

At the output level, logit-based distillation aligns the predicted class distributions, while relation-based distillation preserves structural dependencies among predictions. At the feature level, two projection modules are introduced to bridge the representational gap by aligning attention-related representations and spatial features.

Through this design, the student is guided to approximate both the predictive behavior and internal representations of the teacher, while maintaining a lightweight architecture suitable for efficient deployment.

\subsubsection{Teacher Architecture: ViT-Base}

The teacher $f_T$ is ViT-Base~\citep{dosovitskiy2020image}, a Vision Transformer with 12 encoder blocks, embedding dimension $d = 768$, and approximately 86M parameters. ViT-Base is selected as the teacher based on its superior performance in baseline experiments, making it a strong source of high-capacity knowledge for distillation. In addition, its transformer architecture is particularly suitable for this role, as self-attention can capture global contextual relationships and long-range dependencies that are difficult for lightweight CNNs to model directly, especially in leaf disease recognition tasks~ \citep{borhani2022deep,thai2023formerleaf,hossain2024deep}.

Each input image is partitioned into non-overlapping $16 \times 16$ patches, forming a sequence of spatial tokens processed by the transformer encoder. The model is initialized from an ImageNet pre-trained checkpoint and fine-tuned on each target dataset, after which its weights are frozen during the distillation stage. During training, the teacher serves as a fixed source of supervision. For each mini-batch, it produces class-level logits $z_T \in \mathbb{R}^{B\times C}$, attention representations $\mathrm{Attn}_T$ derived from the Query, Key, and Value matrices $(Q_T, K_T, V_T)$, value features $V_T$, and intermediate token embeddings $h_T \in \mathbb{R}^{B \times 196 \times 768}$ (excluding the \texttt{[CLS]} token). These outputs are used to guide the student at both prediction and feature levels.

\subsubsection{Student Architecture: Truncated MobileNetV2}

The student network $f_S$ is derived from MobileNetV2~\citep{sandler2018mobilenetv2} by retaining the initial convolutional stem and the first five inverted residual (IR) blocks. MobileNetV2 is adopted as the student backbone due to its favorable trade-off between compactness, computational efficiency, and classification performance, making it well-suited for deployment-oriented leaf disease recognition.

The truncation is performed at the output of the 5th IR block, yielding a feature map of size \(14 \times 14 \times 96\). This design ensures direct spatial alignment with the ViT teacher, which represents each \(224 \times 224\) input as \(14 \times 14 = 196\) patch tokens. As a result, the student feature map can be reshaped into 196 spatial positions without requiring additional resizing or interpolation.

While the 4th IR block also produces a \(14 \times 14\) feature map, its lower channel dimensionality (\(64\)) provides a less expressive representation and would require a more aggressive projection to match the teacher embedding dimension. In contrast, the \(14 \times 14 \times 96\) representation offers improved semantic capacity while preserving spatial correspondence.

Deeper layers of MobileNetV2 reduce the spatial resolution to \(7 \times 7\), which breaks the one-to-one alignment with teacher tokens and would require artificial spatial expansion, potentially degrading the fidelity of position-wise distillation. Therefore, the 5th IR block provides a balanced trade-off between spatial alignment, representation capacity, and computational efficiency for cross-architecture knowledge distillation.

\subsection{Distillation Objective}
\label{sec:losses}

\subsubsection{Teacher Objective}

The teacher model is fine-tuned on each target dataset using supervised learning. Let $z_T \in \mathbb{R}^{B \times C}$ denote the teacher logits for a mini-batch of size $B$ over $C$ classes, and let $y$ denote the ground-truth labels. The training objective is the standard cross-entropy loss:
\begin{equation}
  \mathcal{L}_{T} = \mathcal{L}_{CE}(z_T, y)
\end{equation}
where
\begin{equation}
  \mathcal{L}_{CE} = -\frac{1}{B}\sum_{i=1}^{B}\sum_{c=1}^{C} y_{i,c} \log \big( \sigma(z_T^{(i,c)}) \big).
\end{equation}
and $\sigma(\cdot)$ denotes the softmax function.

The teacher is initialized from ImageNet pre-trained weights and fine-tuned prior to distillation. After training, its parameters are frozen, and its logits and intermediate representations are used as supervisory information for training the student.

\subsubsection{Student Objective}

AgriKD combines a standard classification loss with four distillation terms into a unified training objective.

\vspace{0.2cm}
\textbf{i) Logit-based Distillation}
\vspace{0.2cm}

The logit distillation term encourages the student to match the softened output distribution of the teacher~\citep{hinton2015distilling}:
\begin{equation}
\mathcal{L}_{\mathrm{KD}}
=
\mathrm{KL}\!\left(
  \sigma(z_T / T) \,\Big\|\, \sigma(z_S / T)
\right),
\label{eq:lkd}
\end{equation}
where $T > 0$ is the temperature and $\sigma(\cdot)$ denotes the softmax function applied to logits.

\vspace{0.2 cm}
\textbf{{ii) Relation-based Distillation}}
\vspace{0.2 cm}

Instead of enforcing exact probability matching, this term preserves the relational structure of the teacher's output distribution by measuring correlation agreement at both the inter-instance and inter-class levels~\citep{Huang2022DIST}. Let $P_T, P_S \in \mathbb{R}^{B\times C}$ denote the teacher and student softmax probability matrices, and let $\rho(\cdot,\cdot)$ denote the Pearson correlation coefficient.

The inter-instance term measures how consistently each sample's predicted distribution correlates between teacher and student across the batch:
\begin{equation}
\mathcal{L}_{\mathrm{inter}}
=
\frac{1}{B} \sum_{i=1}^{B}
\bigl(1 - \rho(P_T^{(i,:)},\, P_S^{(i,:)})\bigr).
\label{eq:linter}
\end{equation}

The inter-class term measures how consistently each class's predicted score correlates between teacher and student across all instances:
\begin{equation}
\mathcal{L}_{\mathrm{intra}}
=
\frac{1}{C} \sum_{j=1}^{C}
\bigl(1 - \rho(P_T^{(:,j)},\, P_S^{(:,j)})\bigr).
\label{eq:lintra}
\end{equation}

\begin{figure}[h]
\centering
\includegraphics[width=0.82\columnwidth]{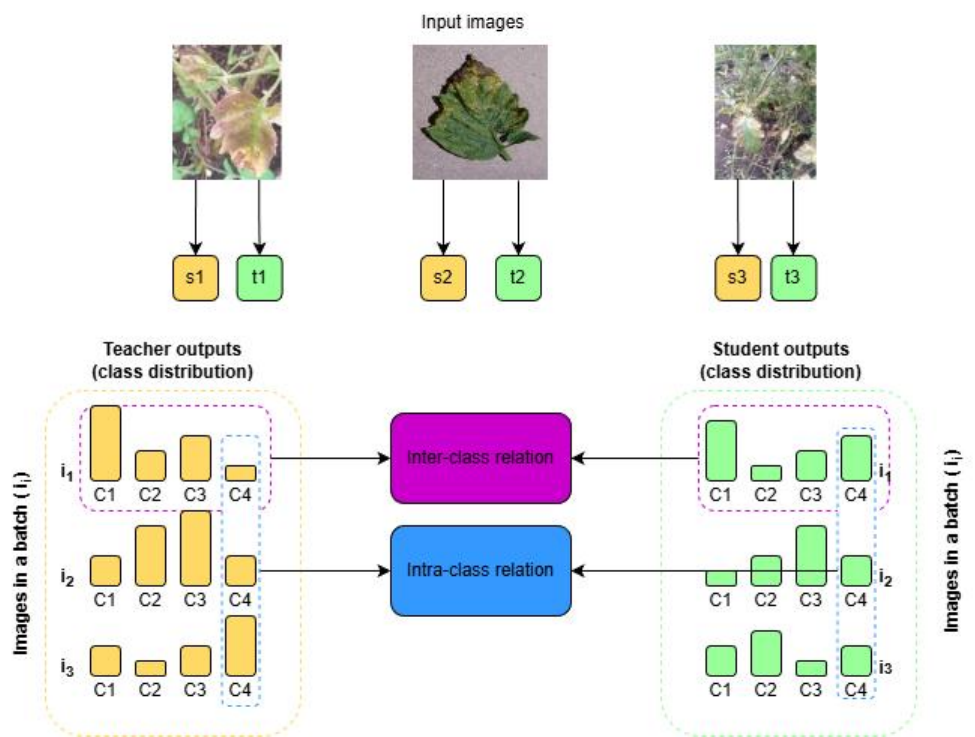}
\caption{Illustration of the relation-based distillation component. The loss preserves structural agreement between teacher and student predictions through inter-instance and inter-class correlation matching.}
\label{fig:relation_based}
\end{figure}

The overall relation loss is defined as a weighted combination of the two components:
\begin{equation}
\mathcal{L}_{\mathrm{rel}}
=
\beta_1 \mathcal{L}_{\mathrm{inter}}
+
\beta_2 \mathcal{L}_{\mathrm{intra}},
\label{eq:lrel}
\end{equation}
where $\beta_1$ and $\beta_2$ control the relative contribution of inter-instance and inter-class correlation preservation, respectively. Figure~\ref{fig:relation_based} illustrates the computation of this loss.

\FloatBarrier

\vspace{0.2 cm}
\textbf{{iii) Projection 1: Partially Cross-Attention (PCA) Projector}}
\vspace{0.2 cm}

\begin{figure}[h]
\centering
\includegraphics[width=0.98\columnwidth]{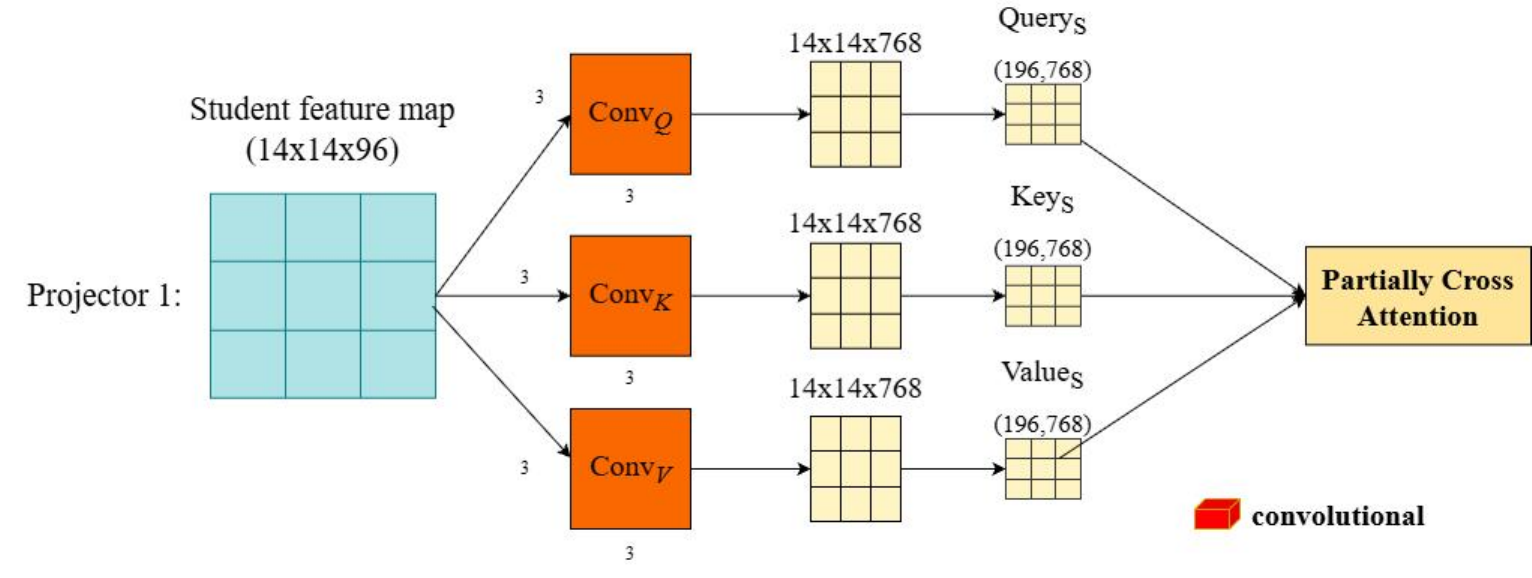}
\caption{Projection~1: Partially Cross-Attention (PCA) projector for aligning student convolutional features with teacher attention representations.}
\label{fig:proj1}
\end{figure}

To bridge the representational gap between the CNN student and the ViT teacher,
we adopt the partially cross-attention (PCA) mechanism proposed by \citet{Liu2022CAKD}.
Unlike direct feature matching, PCA enables the student to approximate the
teacher’s global relational structure by constructing a transformer-style attention
mechanism within the student feature space.
Figure~\ref{fig:proj1} illustrates the overall design of the PCA projector.

Three $3\times3$ convolutional layers project the student feature map $\mathbf{F}_S \in \mathbb{R}^{B \times C' \times 14 \times 14}$ into query, key, and value tensors:
\begin{equation}
\begin{aligned}
Q^{(s)} &= \mathrm{Conv}_Q(\mathbf{F}_S),
K^{(s)} &= \mathrm{Conv}_K(\mathbf{F}_S),
V^{(s)} &= \mathrm{Conv}_V(\mathbf{F}_S)
\end{aligned}
\label{eq:qkv}
\end{equation}
Each mapping converts $C'$ input channels to $d$ output channels and is reshaped to $\mathbb{R}^{B \times 196 \times d}$ after spatial flattening. To bridge the representational gap between the CNN student and ViT teacher, a partial cross-attention mask stochastically replaces elements of the student matrices with the corresponding teacher values:
\begin{equation}
\tilde{M}^{(s)}_{b,i,j} =
\begin{cases}
M^{(t)}_{b,i,j} & \text{if } \xi_{b,i,j} < p \\
M^{(s)}_{b,i,j} & \text{otherwise}
\end{cases}
\label{eq:mask}
\end{equation}
where $M \in \{Q, K, V\}$ and $\xi_{b,i,j} \sim \mathrm{Uniform}(0,1)$. The student attention output is computed as:
\begin{equation}
\mathrm{PCAttn}^{(s)} =
\mathrm{softmax}\!\left(
\frac{\hat{Q}^{(s)}\hat{K}^{(s)\top}}{\sqrt{d}}
\right)\hat{V}^{(s)}
\label{eq:pcattn}
\end{equation}
The Projection~1 loss combines attention matching and value correlation matching:
\begin{equation}
\mathcal{L}_{\mathrm{proj1}} =
\underbrace{\left\|\mathrm{Attn}^{(t)} - \mathrm{PCAttn}^{(s)}\right\|_2^2}_{\text{attention matching}}
+
\underbrace{\left\| \frac{V^{(t)} \odot V^{(t)}}{\sqrt{d}} - \frac{V^{(s)} \odot V^{(s)}}{\sqrt{d}} \right\|_2^2}_{\text{value correlation matching}}
\label{eq:lproj1}
\end{equation}
where $\mathrm{Attn}^{(t)}$ is the teacher's attention output from the corresponding transformer block and $\odot$ denotes element-wise multiplication. This objective encourages the student to reproduce both transformer-style attention patterns and value-level self-correlations within a convolutional feature space.

\vspace{0.2 cm}
\textbf{{iv) Projection 2: Group-wise Linear (GWL) Projector}}
\vspace{0.2 cm}

\begin{figure}[h]
\centering
\includegraphics[width=0.8\textwidth]{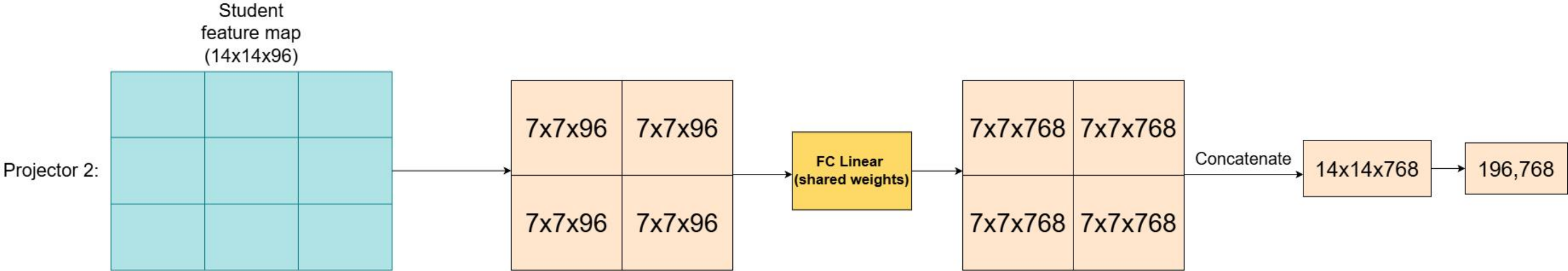}
\caption{Projection~2: Group-wise Linear (GWL) projector for matching grouped student spatial features to teacher token representations.}
\label{fig:proj2}
\end{figure}

The second feature-projection term aligns the student's spatial features with the teacher's intermediate token representations through group-wise linear projection~\citep{Liu2022CAKD}. As illustrated in Figure~\ref{fig:proj2}, the student feature map $\mathbf{F}_S \in \mathbb{R}^{B \times 14 \times 14 \times C'}$ is partitioned into four non-overlapping $7 \times 7$ quadrants corresponding to the top-left, top-right, bottom-left, and bottom-right spatial regions:
\begin{equation}
G_k = \mathbf{F}_{S,R_k}
\in \mathbb{R}^{B \times 49 \times C'},
\quad k \in \{1,2,3,4\}
\label{eq:groups}
\end{equation}
Each group is projected from $C'$ channels to the transformer embedding dimension $d$ via a shared fully-connected linear layer with weight $W \in \mathbb{R}^{C' \times d}$ and bias $b \in \mathbb{R}^{d}$:
\begin{equation}
G'_k = G_k W + b
\in \mathbb{R}^{B \times 49 \times d},
\quad k \in \{1,2,3,4\}
\label{eq:gwl}
\end{equation}
The use of shared weights across all four groups encourages spatially consistent projection while reducing the number of learnable parameters. The four projected groups are spatially reassembled to produce the projected student feature:
\begin{equation}
h'^{(s)} = \!\left(
\mathrm{Concat}(G'_1, G'_2, G'_3, G'_4)
\right)
\in \mathbb{R}^{B \times 196 \times d}
\label{eq:hsprime}
\end{equation}
where the concatenation restores the full $14 \times 14 = 196$ spatial positions, each now embedded in $\mathbb{R}^d$. Let $h^{(t)} \in \mathbb{R}^{B \times 196 \times 768}$ denote the teacher's intermediate token representations from the corresponding transformer block. The Projection~2 loss is the mean squared error between the teacher and projected student features:
\begin{equation}
\mathcal{L}_{\mathrm{proj2}} =
\left\| h^{(t)} - h'^{(s)} \right\|_2^2
\label{eq:lproj2}
\end{equation}

\vspace{0.2cm}
\textbf{v) Total Distillation Objective}
\vspace{0.2cm}

The student is trained by minimizing a unified objective that integrates a standard classification loss with four distillation components:
\begin{equation}
\mathcal{L}_{\mathrm{total}}
=
\lambda_1 \mathcal{L}_{\mathrm{CE}}
+ \lambda_2 \mathcal{L}_{\mathrm{KD}}
+ \lambda_3 \mathcal{L}_{\mathrm{rel}}
+ \lambda_4 \mathcal{L}_{\mathrm{proj1}}
+ \lambda_5 \mathcal{L}_{\mathrm{proj2}},
\label{eq:ltotal}
\end{equation}
where $\mathcal{L}_{\mathrm{CE}}$ denotes the cross-entropy classification loss, and $\lambda_1,\dots,\lambda_5 \geq 0$ are weighting coefficients that balance the contribution of each term.

The coefficients $\lambda_1,\dots,\lambda_5$ are initialized using a simple heuristic based on the relative contribution of each loss component. This strategy avoids manual tuning while providing a data-driven initialization, and is further validated through ablation in Section~\ref{subsubsec:kd-heuristic}. For each component, a reference score $s_k$ is estimated from its individual impact on validation performance over a small subset of the training data. The weights are then normalized as:
\begin{equation}
\lambda_k = \frac{s_k}{\sum_{j=1}^{5} s_j},
\quad k = 1, \dots, 5.
\label{eq:heuristic}
\end{equation}
This provides a reasonable initialization of the multi-loss balance without requiring manual tuning.

\subsection{Imbalance Handling}
\label{sec:Imbalance}

To mitigate class imbalance, we consider two complementary strategies: Weighted Random Sampling (WRS) and Focal Loss (FL)~\citep{lin2017focal}.

WRS operates at the data level by adjusting the sampling probability of each sample inversely proportional to its class frequency:
\begin{equation}
w_i = \frac{1}{n_{c_i}},
\end{equation}
where $n_{c_i}$ denotes the number of training samples in class $c_i$. This increases the likelihood of selecting minority-class samples during mini-batch construction.

At the loss level, Focal Loss emphasizes hard examples by down-weighting well-classified samples:
\begin{equation}
\mathrm{FL} = -\alpha_t (1 - p_t)^\gamma \log(p_t),
\end{equation}
where $\alpha_t$ is a class-balancing factor and $\gamma$ controls the focusing strength.

The application of these methods varies across datasets and is determined empirically.

\subsection{Training Procedure}

This section describes the training process of the proposed distillation framework.
The teacher model is first pre-trained and then kept frozen during the distillation phase to provide stable supervision information.
For each mini-batch, the teacher produces both logits and intermediate feature representations, while the student processes the same inputs and is optimized by minimizing the composite loss defined in Eq.~(\ref{eq:ltotal}).

The overall training workflow of AgriKD is summarized in Algorithm~\ref{alg:agrikd}.

\begin{algorithm}[h]
\caption{AgriKD Training Procedure with Cross-Validation}
\label{alg:agrikd}
\KwIn{Full dataset $\mathcal{D}$, number of folds $K$, pre-trained teacher checkpoints $\{f_T^{(k)}\}_{k=1}^{K}$, temperature $T$, loss weights $\lambda_1,\dots,\lambda_5$, number of epochs $E$}
\KwOut{Fold-wise trained students $\{f_S^{(k)}\}_{k=1}^{K}$}

Initialize $\lambda_1,\dots,\lambda_5$ via heuristic weighting (Eq.~\ref{eq:heuristic})\;

Partition $\mathcal{D}$ into $K$ stratified folds $\{\mathcal{D}^{(1)}, \dots, \mathcal{D}^{(K)}\}$\;

\For{fold $k = 1$ \KwTo $K$}{
$\mathcal{D}_{\mathrm{test}}^{(k)} \leftarrow \mathcal{D}^{(k)}$\tcp*{1 fold held out for testing}

Split $\mathcal{D} \setminus \mathcal{D}^{(k)}$ into $\mathcal{D}_{\mathrm{train}}^{(k)}$ and $\mathcal{D}_{\mathrm{val}}^{(k)}$\tcp*{remaining $K{-}1$ folds split into train/val}

Load frozen teacher $f_T^{(k)}$ (fine-tuned on $\mathcal{D}_{\mathrm{train}}^{(k)}$)\;

Initialize student $f_S^{(k)}$ with parameters $\theta_S$\;

\For{epoch $e = 1$ \KwTo $E$}{
\For{each mini-batch $(x, y) \sim \mathcal{D}_{\mathrm{train}}^{(k)}$}{
$z_T,\; \mathrm{Attn}_T,\; V_T,\; h_T \leftarrow f_T^{(k)}(x)$\;
$z_S,\; \mathbf{F}_S \leftarrow f_S^{(k)}(x)$\;

$\mathcal{L}_{\mathrm{CE}} \leftarrow \mathrm{CrossEntropy}(z_S, y)$\;
$\mathcal{L}_{\mathrm{KD}} \leftarrow \text{Eq.~\ref{eq:lkd}}$\;
$\mathcal{L}_{\mathrm{rel}} \leftarrow \text{Eq.~\ref{eq:lrel}}$\;
$\mathcal{L}_{\mathrm{proj1}} \leftarrow \text{Eq.~\ref{eq:lproj1}}$\;
$\mathcal{L}_{\mathrm{proj2}} \leftarrow \text{Eq.~\ref{eq:lproj2}}$\;

$\mathcal{L}_{\mathrm{total}} \leftarrow \sum_{m=1}^{5} \lambda_m \mathcal{L}_m$\;
$\theta_S \leftarrow \theta_S - \nabla_{\theta_S}\mathcal{L}_{\mathrm{total}}$\;
}
Monitor $f_S^{(k)}$ on $\mathcal{D}_{\mathrm{val}}^{(k)}$ for early stopping\;
}
Evaluate best $f_S^{(k)}$ on $\mathcal{D}_{\mathrm{test}}^{(k)}$\;
}
\Return $\{f_S^{(k)}\}_{k=1}^{K}$\;
\end{algorithm}

\section{Experiments}
\label{sec:experiments}

\subsection{Datasets}
\label{sec:datasets}

We evaluate the proposed framework on three leaf disease image datasets covering different crops, class distributions, and acquisition conditions. These datasets enable a comprehensive assessment of the method under varying levels of visual complexity and class imbalance.

\subsubsection{Tomato Leaf Disease}
\label{Tomato}

The Tomato Leaf Disease dataset~\citep{Solapure2024Tomato} contains 1,609 images across 10 classes, including nine disease categories and one healthy class. The dataset exhibits moderate class imbalance, with a majority-to-minority ratio of approximately 4.04 (400 vs. 99 samples).

Image resolutions vary from $256\times256$ to $4096\times4096$, and all images are resized to $224\times224$ for model training. The class distribution is summarized in Table~\ref{tab:tomato_distribution}, with representative samples shown in Figure~\ref{fig:tomato_samples}.

\begin{center}
\footnotesize
\captionof{table}{Class distribution of the Tomato Leaf Disease dataset.}
\label{tab:tomato_distribution}
\begin{tabular}{lc}
\hline
\textbf{Class} & \textbf{Images} \\
\hline
Bacterial Spot & 108 \\
Early Blight & 298 \\
Late Blight & 111 \\
Leaf Mold & 400 \\
Septoria Leaf Spot & 106 \\
Spider Mites & 121 \\
Target Spot & 99 \\
Tomato Yellow Leaf Curl Virus & 100 \\
Tomato Mosaic Virus & 109 \\
Healthy & 157 \\
\hline
\textbf{Total} & \textbf{1609} \\
\hline
\end{tabular}
\end{center}

\begin{center}
\includegraphics[width=0.75\linewidth]{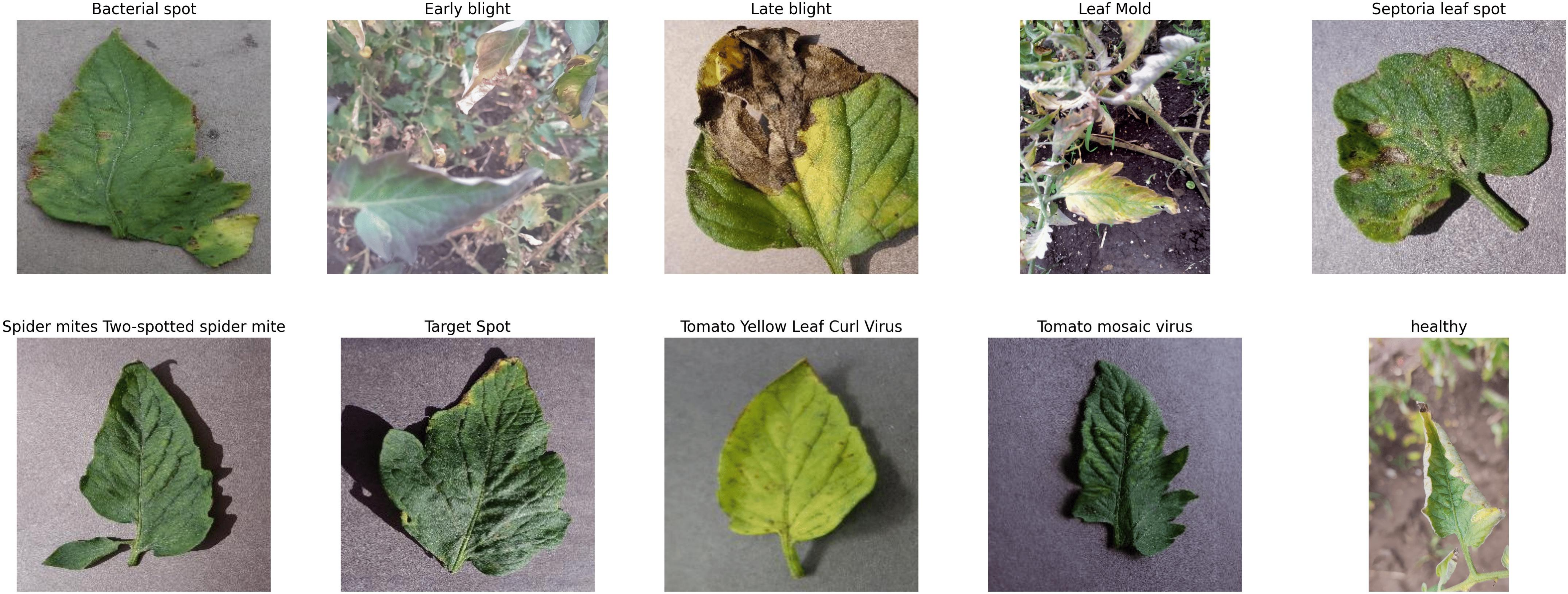}
\captionof{figure}{Representative samples from each class in the Tomato Leaf Disease dataset.}
\label{fig:tomato_samples}
\end{center}

\FloatBarrier
\subsubsection{Burmese Grape Leaf Disease}
\label{sec:Burmese}

The Burmese Grape Leaf Disease dataset~\citep{Rahman2025Burmese} contains 3,103 images across five classes, including four disease categories and one healthy class. The dataset exhibits moderate class imbalance, with a majority-to-minority ratio of approximately 3.4 (1006 vs. 296 samples).

Since the images are captured under relatively controlled conditions, a foreground-based preprocessing step is applied to isolate the leaf region before resizing to $224\times224$. This step reduces background variation while preserving disease-relevant visual patterns. The class distribution is summarized in Table~\ref{tab:burmese_distribution}, and representative samples along with the preprocessing pipeline are shown in Figure~\ref{fig:burmese_overview}.
\begin{center}
\footnotesize
\captionof{table}{Class distribution of the Burmese Grape Leaf Disease dataset.}
\label{tab:burmese_distribution}
\begin{tabular}{lc}
\hline
\textbf{Class} & \textbf{Images} \\
\hline
Healthy & 1006 \\
Anthracnose (Brown Spot) & 447 \\
Insect Damage & 990 \\
Powdery Mildew & 296 \\
Leaf Spot (Yellow) & 364 \\
\hline
\textbf{Total} & \textbf{3103} \\
\hline
\end{tabular}
\end{center}

\begin{center}
\captionsetup{type=figure}
\begin{subfigure}[h]{0.95\linewidth}
\centering
\includegraphics[width=\linewidth]{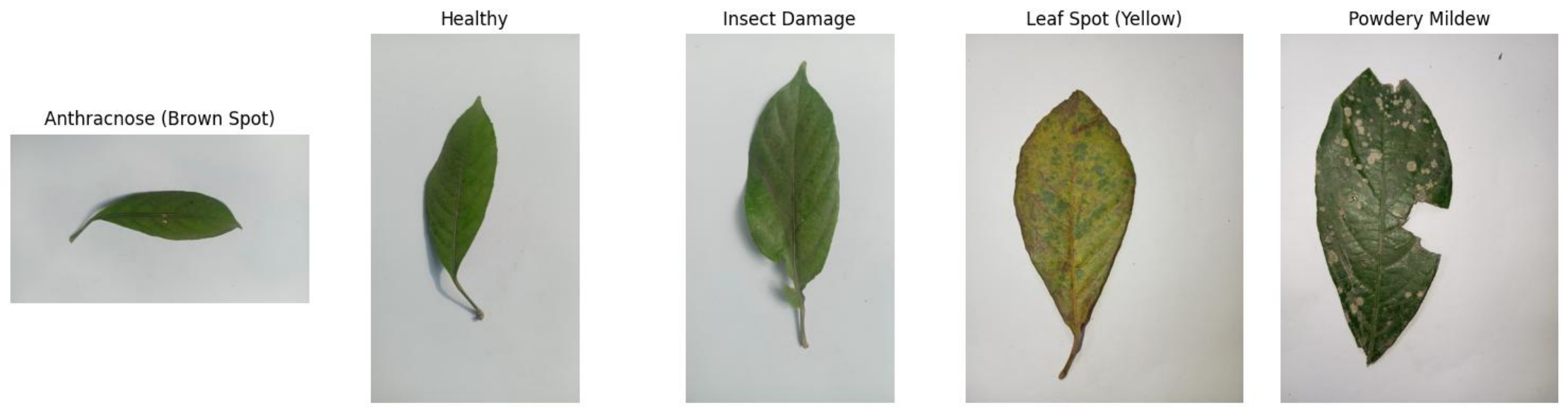}
\caption{Representative samples from each class.}
\label{fig:burmese_samples}
\end{subfigure}

\vspace{1em}

\begin{subfigure}[h]{0.95\linewidth}
\centering
\includegraphics[width=\linewidth]{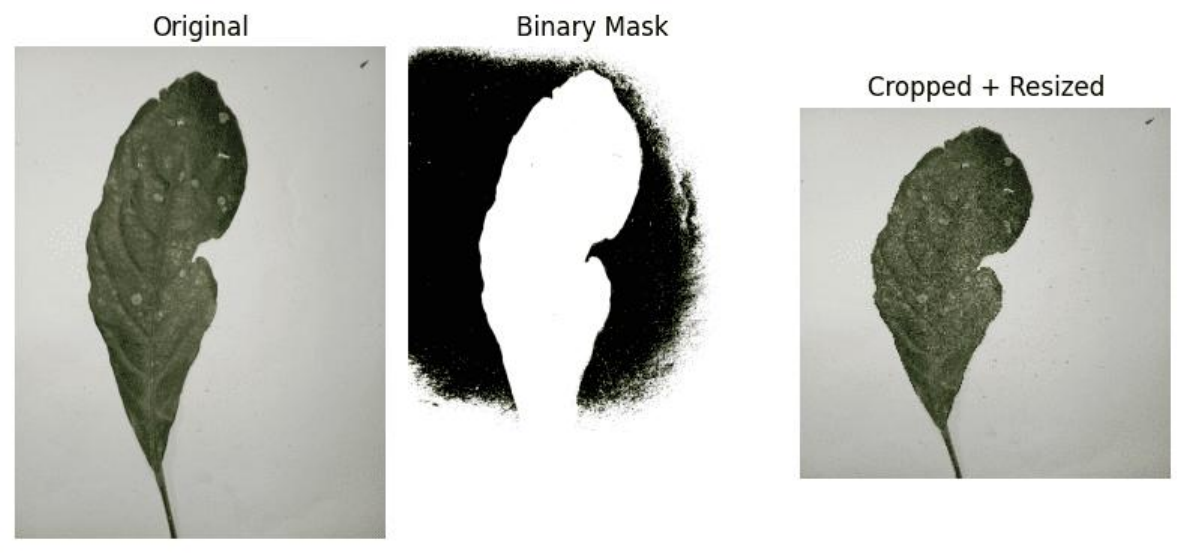}
\caption{Foreground extraction and resizing pipeline.}
\label{fig:burmese_preprocess}
\end{subfigure}

\vspace{0.5em}
\captionof{figure}{Burmese Grape Leaf Disease dataset overview.}
\label{fig:burmese_overview}
\end{center}

\FloatBarrier
\subsubsection{Potato Leaf Disease}
\label{sec:Potato}
The Potato Leaf Disease dataset~\citep{Shabrina2023Potato} contains 3,076 images across seven classes, including six disease categories and one healthy class. Among the three datasets, it exhibits the highest class imbalance, with a majority-to-minority ratio of $748/68 \approx 11.0$.

All images have a fixed resolution of $1500\times1500$ pixels and are resized to $224\times224$ using a standard preprocessing pipeline. Unlike the Burmese Grape dataset, no foreground extraction is applied, as the images are captured under more diverse and less controlled conditions, where threshold-based cropping may introduce segmentation errors.

The class distribution is summarized in Table~\ref{tab:potato_distribution}, and representative samples are shown in Figure~\ref{fig:potato_samples}.
\begin{center}
\footnotesize
\captionof{table}{Class distribution of the Potato Leaf Disease dataset.}
\label{tab:potato_distribution}
\begin{tabular}{lc}
\hline
\textbf{Class} & \textbf{Images} \\
\hline
Bacteria & 569 \\
Fungi & 748 \\
Healthy & 201 \\
Nematode & 68 \\
Pest & 611 \\
Phytophthora & 347 \\
Virus & 532 \\
\hline
\textbf{Total} & \textbf{3076} \\
\hline
\end{tabular}
\end{center}
\vspace{-0.5cm}
\begin{center}
\includegraphics[width=\linewidth]{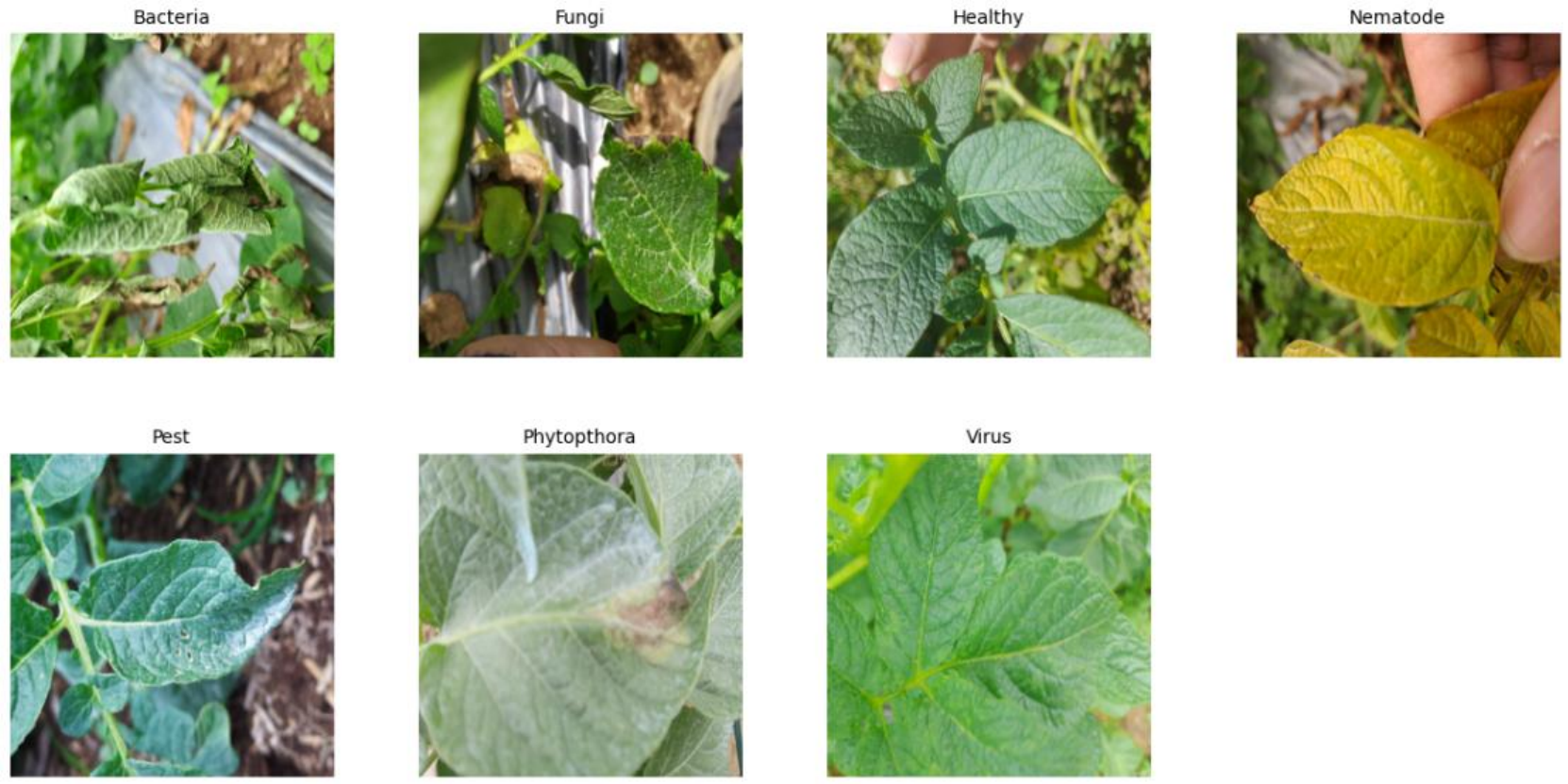}
\captionof{figure}{Representative samples from each class in the Potato Leaf Disease dataset.}
\label{fig:potato_samples}
\end{center}

In this study,  stratified five-fold cross-validation is applied to all three datasets.
\FloatBarrier

\subsection{Experiment Settings}
\label{subsec:experiment-settings}

All experiments are implemented in PyTorch, with all models initialized from ImageNet-1K pre-trained weights. The input resolution is fixed at 224 $\times$ 224 across all datasets. During training, standard data augmentation techniques, including random horizontal flipping, rotation, and color jittering, are applied. No augmentation is used for validation and testing. For the Burmese Grape dataset, foreground-extraction preprocessing is applied prior to resizing.

\subsubsection{Baseline Configuration}
Each dataset is trained using independently tuned hyperparameters. All baseline models utilize a linear warmup followed by cosine annealing learning rate scheduling. Optimizers are selected based on preliminary experiments to ensure stable convergence: Adam is used for most architectures, while AdamW is adopted for ViT-Base to handle weight decay more effectively. To ensure fair comparison, the same optimizer is consistently used across models within each specific dataset. Detailed configurations for these baseline models are summarized in Table~\ref{tab:config-baseline}.

\begin{center}
\captionof{table}{Hyperparameter configuration for baseline models
on Tomato, Burmese Grape, and Potato datasets}
\label{tab:config-baseline}
\resizebox{\columnwidth}{!}{%
\begin{tabular}{ll c c c c c c c c c c}
\hline\hline
\textbf{Dataset} & \textbf{Model} & \textbf{BS} & \textbf{Epochs}
& \textbf{LR} & \textbf{WD} & \textbf{Warmup}
& $\boldsymbol{\eta_{\min}}$ & \textbf{ES} & \textbf{Dropout}
& \textbf{FC Head} & \textbf{Optimizer} \\
\hline
\multirow{6}{*}{Tomato}
& VGG16          & 64 & 150 & $5{\times}10^{-4}$   & $1{\times}10^{-7}$ & 9 & $1{\times}10^{-6}$ & 15 & 0.5 & [256, 128] & Adam \\
& ResNet101      & 64 & 150 & $5{\times}10^{-4}$   & $1{\times}10^{-7}$ & 9 & $1{\times}10^{-6}$ & 15 & 0.5 & [256, 128] & Adam \\
& MobileNetV2    & 32 & 150 & $1.5{\times}10^{-4}$ & $1{\times}10^{-4}$ & 9 & $1{\times}10^{-5}$ & 15 & 0.4 & [512]      & Adam \\
& DenseNet121    & 64 & 100 & $1.5{\times}10^{-5}$ & $5{\times}10^{-2}$ & 6 & $1{\times}10^{-6}$ & 10 & 0.2 & [512]      & Adam \\
& EfficientNetB0 & 64 & 150 & $5{\times}10^{-4}$   & $1{\times}10^{-7}$ & 9 & $1{\times}10^{-6}$ & 15 & 0.5 & [256, 128] & Adam \\
& ViT-Base       & 32 &  50 & $9{\times}10^{-7}$   & $1{\times}10^{-1}$ & 5 & $1{\times}10^{-5}$ & 10 & 0.4 & [512]      & AdamW \\
\hline
\multirow{6}{*}{Burmese Grape}
& VGG16          & 16 & 100 & $2{\times}10^{-5}$ & $5{\times}10^{-4}$ & 6 & $1{\times}10^{-6}$ & 15 & 0.5 & [256, 128] & Adam \\
& ResNet101      & 64 &  50 & $2{\times}10^{-5}$ & $1{\times}10^{-7}$ & 3 & $1{\times}10^{-6}$ & 10 & 0.5 & [256, 128] & Adam \\
& MobileNetV2    & 64 & 100 & $1{\times}10^{-3}$ & $1{\times}10^{-7}$ & 6 & $1{\times}10^{-6}$ & 20 & 0.5 & [256, 128] & Adam \\
& DenseNet121    & 64 & 100 & $1{\times}10^{-5}$ & $1{\times}10^{-7}$ & 6 & $1{\times}10^{-6}$ & 15 & 0.5 & [256, 128] & Adam \\
& EfficientNetB0 & 64 & 100 & $2{\times}10^{-3}$ & $1{\times}10^{-6}$ & 6 & $1{\times}10^{-6}$ & 15 & 0.5 & [256, 128] & Adam \\
& ViT-Base       & 64 &  50 & $2{\times}10^{-5}$ & $1{\times}10^{-7}$ & 3 & $1{\times}10^{-6}$ & 10 & 0.5 & [256, 128] & AdamW \\
\hline
\multirow{6}{*}{Potato}
& VGG16          & 32 &  70 & $5{\times}10^{-4}$   & $1{\times}10^{-6}$ & 7 & $1{\times}10^{-5}$ & 10 & 0.5 & [256, 128] & Adam \\
& ResNet101      & 64 &  50 & $6{\times}10^{-4}$   & $1{\times}10^{-6}$ & 5 & $1{\times}10^{-5}$ & 10 & 0.6 & [256, 128] & Adam \\
& MobileNetV2    & 32 & 150 & $1.5{\times}10^{-4}$ & $1{\times}10^{-4}$ & 9 & $1{\times}10^{-5}$ & 15 & 0.4 & [512]      & Adam \\
& DenseNet121    & 32 &  50 & $5{\times}10^{-4}$   & $1{\times}10^{-6}$ & 5 & $1{\times}10^{-5}$ & 10 & 0.5 & [256, 128] & Adam \\
& EfficientNetB0 & 64 &  50 & $6{\times}10^{-4}$   & $1{\times}10^{-6}$ & 5 & $1{\times}10^{-5}$ & 10 & 0.5 & [256, 128] & Adam \\
& ViT-Base       & 64 &  50 & $2{\times}10^{-5}$   & $1{\times}10^{-7}$ & 3 & $1{\times}10^{-6}$ & 10 & 0.5 & [256, 128] & Adam \\
\hline\hline
\end{tabular}}
\begin{minipage}{\columnwidth}
\smallskip\footnotesize
BS: batch size; WD: weight decay; ES: early stopping patience;
FC Head: number of neurons in the fully connected classifier layers.
All models use linear warmup followed by cosine annealing scheduling.
Warmup values are reported in epochs.
\end{minipage}
\end{center}

\subsubsection{Knowledge Distillation Configuration}
Table~\ref{tab:kd-config-all} summarizes the dataset-specific hyperparameters used for Knowledge Distillation (KD) training. These experiments utilize the teacher checkpoints selected in Section~\ref{subsec:teacher-results}: ViT-Base teachers trained without imbalance-handling for Tomato and Burmese Grape, and the ViT-Base teacher trained with CE + WRS for Potato.

Shared settings across all three datasets include a truncated MobileNetV2 student (output of the 5th inverted residual block, 14 $\times$ 14 $\times$ 96), a student FC head of [512, 256], dropout of 0.30, and label smoothing of 0.1. The distillation process uses a temperature T = 4 and relation-loss coefficients $\beta_1 = \beta_2 = 1$. The Adam optimizer is employed for all KD runs, with teacher blocks fixed at transformer output position 1 and QKV position 12. All KD experiments follow the same Linear Warmup + Cosine Annealing scheduler, with dataset-specific warmup durations and minimum learning rates ($\eta_{\min}$) as detailed in Table~\ref{tab:kd-config-all}.

\begin{center}
\captionof{table}{KD training hyperparameters across datasets}
\label{tab:kd-config-all}
\resizebox{\columnwidth}{!}{%
\begin{tabular}{lcccccccc}
\hline\hline
\textbf{Dataset} & \textbf{BS} & \textbf{Epochs} & \textbf{LR} & \textbf{ES} & \textbf{Proj1 Dropout} & \textbf{Proj1 $p$} & \textbf{Warmup} & \textbf{$\eta_{\min}$} \\
\hline
Tomato & 16 & 60 & $3{\times}10^{-4}$ & 10 & 0.20 & 0.50 & 10 & $5{\times}10^{-7}$ \\
Burmese Grape & 64 & 150 & $9{\times}10^{-4}$ & 30 & 0.50 & 0.50 & 15 & $1{\times}10^{-8}$ \\
Potato & 32 & 50 & $8{\times}10^{-4}$ & 10 & 0.50 & 0.50 & 5 & $1{\times}10^{-8}$ \\
\hline\hline
\end{tabular}}
\begin{minipage}{\columnwidth}
\smallskip\footnotesize
BS: batch size; ES: early stopping; Proj Dropout: dropout for QKV projectors; Proj $p$: masking probability. Proj2 dropout is fixed at 0.40 for all datasets.
\end{minipage}
\end{center}

\section{Results}

\label{sec:results}
This section presents the experimental results of the proposed framework. We first analyze the baseline model selection process, including classification performance, model complexity, and sensitivity to imbalance-handling methods. We then evaluate teacher robustness under cross-validation before analyzing the effectiveness of the proposed KD framework across all datasets.

\subsection{Baseline Performance and Teacher--Student Selection}
\label{sec:baseline_models}

To systematically explore teacher--student configurations, we consider a diverse set of candidate models spanning multiple architectural paradigms, including classical convolutional neural networks such as VGG16~\citep{simonyan2014very}, ResNet101~\citep{he2016deep}, and DenseNet121~\citep{huang2017densely}, efficient architectures such as EfficientNetB0~\citep{tan2019efficientnet} and MobileNetV2~\citep{sandler2018mobilenetv2}, and transformer-based models such as ViT-Base~\citep{dosovitskiy2020image}.
These models differ in design philosophy, capacity, and computational efficiency, enabling a comprehensive evaluation of knowledge distillation across heterogeneous architectures.

Table~\ref{tab:baseline-all} presents the performance of all candidate architectures on the three datasets under the standard training setting, where no imbalance-handling techniques are applied. All models are evaluated using with dataset-specific hyperparameter configurations detailed in Table~\ref{tab:config-baseline}.

\begin{center}
\captionof{table}{Baseline model performance across all three datasets without imbalance handling. Best result per dataset shown in bold.}
\label{tab:baseline-all}
\resizebox{\columnwidth}{!}{%
\begin{tabular}{l cc cc cc}
\hline\hline
& \multicolumn{2}{c}{\textbf{Tomato}}
& \multicolumn{2}{c}{\textbf{Burmese Grape}}
& \multicolumn{2}{c}{\textbf{Potato}} \\
\cmidrule(lr){2-3}\cmidrule(lr){4-5}\cmidrule(lr){6-7}
\textbf{Model}
& \textbf{Acc (\%)} & \textbf{F1 (\%)}
& \textbf{Acc (\%)} & \textbf{F1 (\%)}
& \textbf{Acc (\%)} & \textbf{F1 (\%)} \\
\hline
ViT-Base       & \textbf{84.40} & \textbf{88.03} & \textbf{86.75} & \textbf{86.96} & \textbf{89.74} & \textbf{88.89} \\
VGG16          & 81.20 & 83.72 & 83.55 & 83.10 & 88.03 & 86.59 \\
DenseNet121    & 80.80 & 86.10 & 83.33 & 81.50 & 88.90 & 86.94 \\
EfficientNetB0 & 80.00 & 83.08 & 83.97 & 83.97 & 86.32 & 84.78 \\
ResNet101      & 77.20 & 82.20 & 83.33 & 83.40 & 86.75 & 85.16 \\
MobileNetV2    & 76.40 & 86.70 & 83.55 & 82.68 & 85.26 & 83.23 \\
\hline\hline
\end{tabular}}
\end{center}

As shown in Table~\ref{tab:baseline-all}, ViT-Base consistently achieves the best performance across all datasets in both accuracy and macro F1-score. Specifically, it outperforms the second-best model by +3.20\% in accuracy and +1.93\% in F1 on Tomato, +2.78\% and +2.99\% on Burmese Grape, and +0.84\% and +1.95\% on Potato, respectively.
In contrast, MobileNetV2 yields the lowest accuracy on Tomato (76.40\%) and Potato (85.26\%), and competitive but lower F1-scores (e.g., 83.23\% on Potato), reflecting its limited capacity compared to larger models.

This clear performance gap between ViT-Base and MobileNetV2 motivates their selection as the teacher and student models, respectively, enabling a meaningful evaluation of knowledge distillation effectiveness.

Beyond predictive performance, the selection of the student model focuses on computational efficiency. Table~\ref{tab:model_complexity} reports the parameter count and model size of all candidate architectures, providing the efficiency dimension for model selection.

\begin{table}[H]
\centering
\footnotesize
\setlength{\tabcolsep}{4pt}
\renewcommand{\arraystretch}{1.1}
\caption{Model complexity comparison of baseline architectures}
\label{tab:model_complexity}
\begin{tabular}{lcc}
\hline\hline
\textbf{Model} & \textbf{Parameters (M)} & \textbf{Size (MB)} \\
\hline
ViT-Base       & 86.03 & 328.00 \\
ResNet101      & 43.06 & 164.00 \\
VGG16          & 14.88 & 80.70 \\
DenseNet121    & 7.25  & 28.20 \\
EfficientNetB0 & 4.37  & 16.90 \\
MobileNetV2    & \textbf{2.59} & \textbf{10.00} \\
\hline\hline
\end{tabular}
\end{table}

From Table~\ref{tab:model_complexity}, MobileNetV2 is the most lightweight architecture, with only 2.59M parameters and a model size of 10.00\,MB, making it significantly smaller than all other candidates (e.g., $\sim$33$\times$ smaller than ViT-Base).

Therefore, while the teacher model is selected based on predictive performance, the student model is chosen based on computational efficiency. MobileNetV2 is adopted as the student due to its compact design, enabling a clear evaluation of performance–efficiency trade-offs in knowledge distillation.

\subsection{Imbalance Sensitivity Analysis}
\label{sec:Imbalance_Results}

To assess whether class imbalance affects model ranking, we evaluate each model under two imbalance-handling strategies, namely weighted random sampling (WRS) and focal loss (FL), using the same hold-out split. Table~\ref{tab:imbalance-main} reports macro F1-score across the baseline, WRS, and FL settings.

The impact of imbalance handling varies across datasets.
On Tomato and Burmese Grape, neither WRS nor FL consistently improves performance over baseline training.
In most cases, baseline training yields the best or comparable F1-scores, and the variation across imbalance-handling strategies remains small.
In particular, ViT-Base consistently achieves the highest performance under the baseline setting, with only minor fluctuations across WRS and FL.
This indicates that the moderate imbalance levels in Tomato (4.04:1) and Burmese Grape (3.40:1) are insufficient to alter model ranking or justify additional imbalance-handling mechanisms.

In contrast, the Potato dataset exhibits a clear and consistent benefit from imbalance-aware training.
WRS improves performance across several models, including ViT-Base (88.89\% to 90.42\%), MobileNetV2 (83.23\% to 88.31\%), EfficientNetB0 (84.78\% to 89.15\%), and ResNet101 (85.16\% to 86.90\%).
These improvements are substantially larger than those observed on the other datasets, suggesting that the higher imbalance level in Potato (11.00:1) makes data-level rebalancing more effective.
In comparison, FL shows less consistent behaviour, with performance gains depending on the model.

\begin{center}
\footnotesize
\setlength{\tabcolsep}{5pt}
\renewcommand{\arraystretch}{1.15}

\captionof{table}{Imbalance-handling results across datasets using F1-score (\%) under the hold-out evaluation protocol. Best result within each model--dataset group is shown in bold.}
\label{tab:imbalance-main}

\resizebox{\columnwidth}{!}{%
\begin{tabular}{l ccc ccc ccc}
\hline\hline

& \multicolumn{3}{c}{\textbf{Tomato}}
& \multicolumn{3}{c}{\textbf{Burmese Grape}}
& \multicolumn{3}{c}{\textbf{Potato}} \\

\cmidrule(lr){2-4}
\cmidrule(lr){5-7}
\cmidrule(lr){8-10}

\textbf{Model}
& \textbf{Base}
& \textbf{WRS}
& \textbf{FL}

& \textbf{Base}
& \textbf{WRS}
& \textbf{FL}

& \textbf{Base}
& \textbf{WRS}
& \textbf{FL}

\\

\hline

ViT-Base
& \textbf{88.03} & 86.61 & 87.55
& \textbf{86.96} & 86.63 & 86.22
& 88.89 & \textbf{90.42} & 89.90
\\

MobileNetV2
& \textbf{86.70} & 80.01 & 83.23
& \textbf{82.68} & 80.26 & 82.08
& 83.23 & \textbf{88.31} & 84.62
\\

DenseNet121
& \textbf{86.10} & 83.40 & 84.61
& 81.50 & 81.25 & \textbf{81.93}
& 86.94 & 86.48 & \textbf{88.46}
\\

EfficientNetB0
& 83.08 & \textbf{87.07} & 86.28
& \textbf{83.97} & 82.81 & 80.55
& 84.78 & \textbf{89.15} & 84.17
\\

VGG16
& 83.72 & \textbf{85.21} & 82.66
& \textbf{83.10} & 82.74 & 81.25
& \textbf{86.59} & 83.72 & 79.63
\\

ResNet101
& 82.20 & 81.92 & \textbf{84.22}
& 83.40 & 83.29 & \textbf{83.69}
& 85.16 & \textbf{86.90} & 84.28
\\

\hline\hline

\end{tabular}}
\end{center}

These results indicate that the effectiveness of imbalance handling is dataset-dependent.
While it has limited impact under moderate imbalance, it becomes beneficial under more severe imbalance conditions.
Accordingly, subsequent experiments retain baseline training for Tomato and Burmese Grape, while additionally considering WRS for Potato.

\FloatBarrier

\subsection{Teacher Model Refinement}
\label{subsec:teacher-results}

After selecting ViT-Base as the teacher architecture, we refine the final
teacher training configuration under stratified five-fold cross-validation.
The imbalance-sensitivity analysis in Section~\ref{sec:Imbalance_Results}
indicates that imbalance handling does not materially improve the ViT-Base
teacher on Tomato and Burmese Grape, where the baseline setting remains the
best or most stable choice. Therefore, the teacher is trained with the baseline
cross-entropy setting on these two datasets.

For Potato, the hold-out imbalance analysis shows that WRS improves the
ViT-Base F1-score from 88.89\% to 90.42\%, outperforming both baseline
training and FL. This result motivates selecting WRS as the imbalance-aware
training configuration for the Potato teacher. We therefore train the Potato
teacher under stratified five-fold cross-validation using CE + WRS.

Table~\ref{tab:teacher-cv-results} reports the teacher cross-validation
results. On Tomato and Burmese Grape, the baseline ViT-Base teacher achieves
macro F1-scores of 86.98 $\pm$ 1.88\% and 83.08 $\pm$ 0.92\%, respectively.
On Potato, the CE + WRS teacher achieves 88.06 $\pm$ 1.21\% macro F1-score,
showing stable performance under severe class imbalance.

Consequently, the final teacher configurations are defined as follows:
baseline cross-entropy training is used for Tomato and Burmese Grape, while
CE + WRS is used for Potato. These refined teacher models are then used as
the fixed supervision sources for all subsequent knowledge distillation
experiments.

\begin{table}[H]
\centering
\caption{Teacher with cross-validation (mean $\pm$ std) across datasets}
\label{tab:teacher-cv-results}
\footnotesize
\setlength{\tabcolsep}{8pt}
\renewcommand{\arraystretch}{1.2}
\begin{tabular}{llcc}
\hline\hline
Dataset & Training Setting & Accuracy (\%) & F1-Score (\%) \\
\hline
Tomato Leaf & Baseline
& 83.84 $\pm$ 1.85 & 86.98 $\pm$ 1.88 \\
Burmese Leaf & Baseline
& 84.30 $\pm$ 1.06 & 83.08 $\pm$ 0.92 \\
Potato Leaf & CE + WRS
& 89.43 $\pm$ 1.23 & 88.06 $\pm$ 1.21 \\
\hline\hline
\end{tabular}
\end{table}

\subsection{Knowledge Distillation Results}
\label{subsec:kd-results}

This section presents the experimental results for the student model trained within the proposed KD framework. Performance is evaluated using the previously refined teacher models across all three datasets. To ensure statistical robustness, all results are reported via stratified five-fold cross-validation, with the macro F1-score serving as the primary evaluation criterion.

\subsubsection{Loss Weight Initialisation}
\label{subsubsec:kd-heuristic}

To optimize the loss function for the student model, we conducted a heuristic setup to determine the optimal loss coefficients $\lambda_1, \ldots, \lambda_5$. Each individual loss component was evaluated in isolation on a small held-out subset to measure its specific contribution to the validation F1-score. These contribution scores were then normalized to establish the dataset-specific weights used throughout all subsequent experiments.

Table~\ref{tab:kd-heuristic-weights} reports the per-component F1-scores and the resulting $\lambda$ values. Across all three datasets, the logits and relation terms consistently receive the largest weights ($\lambda \approx 0.30$--$0.33$), confirming their roles as the primary supervisory objectives. In contrast, the two projection losses receive substantially smaller weights ($\lambda \approx 0.01$--$0.06$), which is consistent with their auxiliary function in feature-map alignment rather than acting as primary distillation objectives.


\begin{table}[H]
\centering
\caption{Heuristic weighting of KD loss coefficients based on per-component validation F1-score contribution.}
\label{tab:kd-heuristic-weights}
\footnotesize
\setlength{\tabcolsep}{6pt}
\renewcommand{\arraystretch}{1.15}

\resizebox{\columnwidth}{!}{%
\begin{tabular}{lcccccccccc}
\hline\hline

\multirow{2}{*}{\textbf{Dataset}} &
\multicolumn{2}{c}{\textbf{Cross Entropy}} &
\multicolumn{2}{c}{\textbf{Projection 1}} &
\multicolumn{2}{c}{\textbf{Projection 2}} &
\multicolumn{2}{c}{\textbf{Logits KD}} &
\multicolumn{2}{c}{\textbf{Relation}} \\

\cline{2-11}

& \textbf{F1} & $\lambda$
& \textbf{F1} & $\lambda$
& \textbf{F1} & $\lambda$
& \textbf{F1} & $\lambda$
& \textbf{F1} & $\lambda$ \\

\hline

Tomato
& 82.36 & 0.3038
& 3.62 & 0.0134
& 9.04 & 0.0333
& 87.21 & 0.3216
& 88.91 & 0.3279 \\

Burmese Grape
& 82.85 & 0.2935
& 17.77 & 0.0630
& 14.15 & 0.0501
& 83.74 & 0.2967
& 83.74 & 0.2967 \\

Potato
& 84.65 & 0.3127
& 3.97 & 0.0147
& 5.38 & 0.0199
& 88.63 & 0.3274
& 88.07 & 0.3253 \\

\hline\hline
\end{tabular}
}

\end{table}

\subsubsection{Distilled Student Performance}
\label{subsubsec:kd-cv}

Figures~\ref{fig:compare-acc} and~\ref{fig:compare-f1} show the accuracy and macro F1-score of the distilled student and the teacher across three datasets.

On Tomato and Burmese Grape, the student consistently outperforms the teacher on both metrics.
For accuracy, the improvements are +1.87\% (Tomato) and +1.36\% (Burmese Grape), indicating better overall prediction correctness.
For macro F1-score, the gains of +1.25\% and +1.30\% further confirm that the improvement is not limited to dominant classes but is distributed across categories.

On the Potato dataset, however, the student shows a slight decrease relative to the teacher, with -0.41\% in accuracy and -1.03\% in macro F1-score.
Notably, the drop is more pronounced in F1-score than in accuracy, suggesting that the performance gap is mainly associated with minority or harder classes rather than overall prediction correctness.
This is consistent with the stronger class imbalance and higher visual variability of the Potato dataset.

Across all datasets, the trends in accuracy and macro F1-score are largely aligned, but macro F1-score provides additional insight into class-wise performance under imbalance.
In particular, the consistent gains on both metrics for Tomato and Burmese Grape indicate that distillation improves both overall accuracy and class-balanced performance, while the larger degradation in F1-score on Potato highlights the remaining challenge in handling imbalanced data with a compact student model.

The results suggest that the effectiveness of distillation is dataset-dependent: it provides clear benefits in balanced settings and maintains competitive performance under more challenging conditions, albeit with some degradation on minority classes.

\begin{center}
\includegraphics[width=1.1\linewidth]{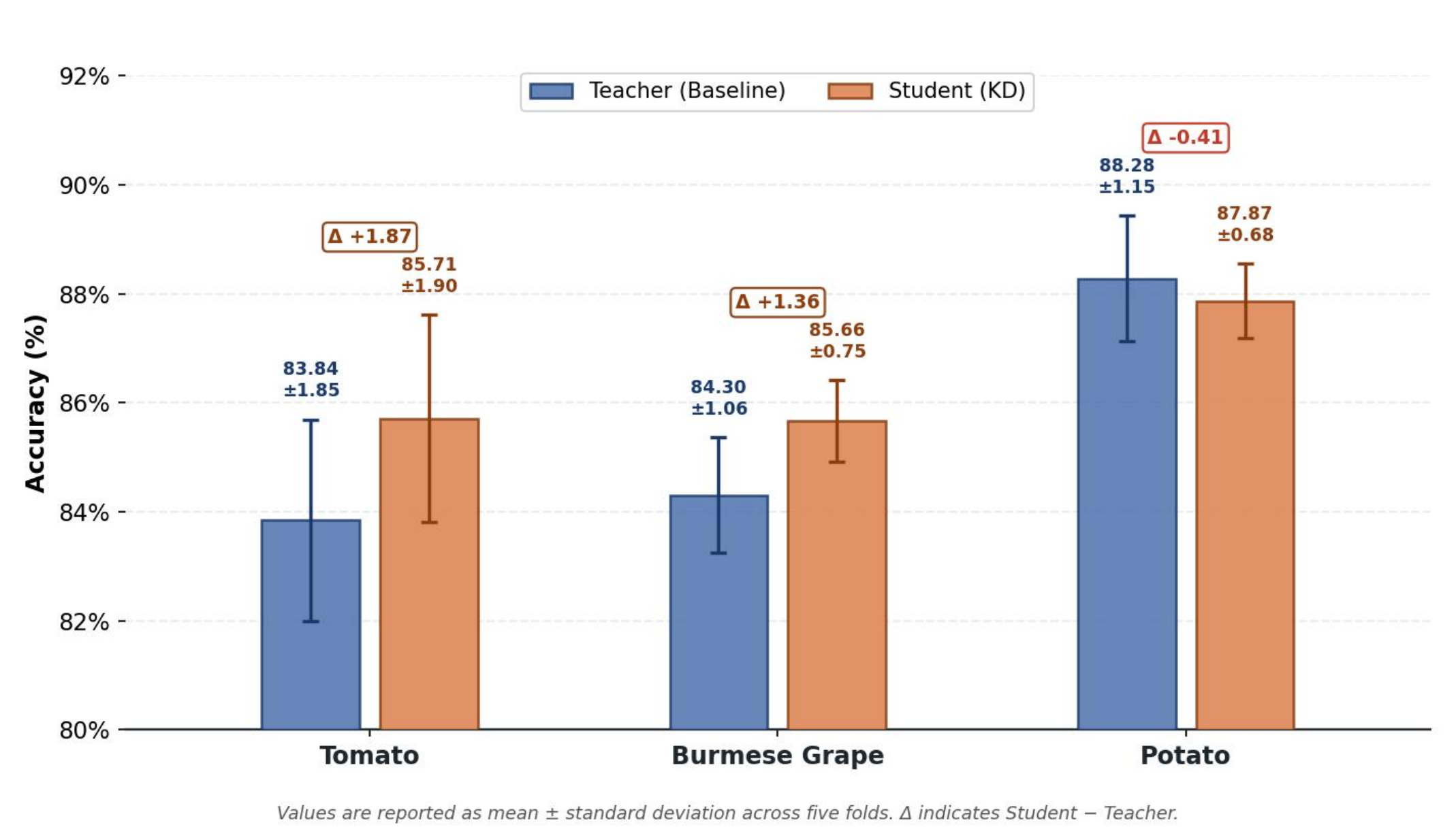}
\captionof{figure}{Accuracy comparison between teacher and distilled student across the three datasets.}
\label{fig:compare-acc}
\end{center}

\begin{center}
\includegraphics[width=1.1\linewidth]{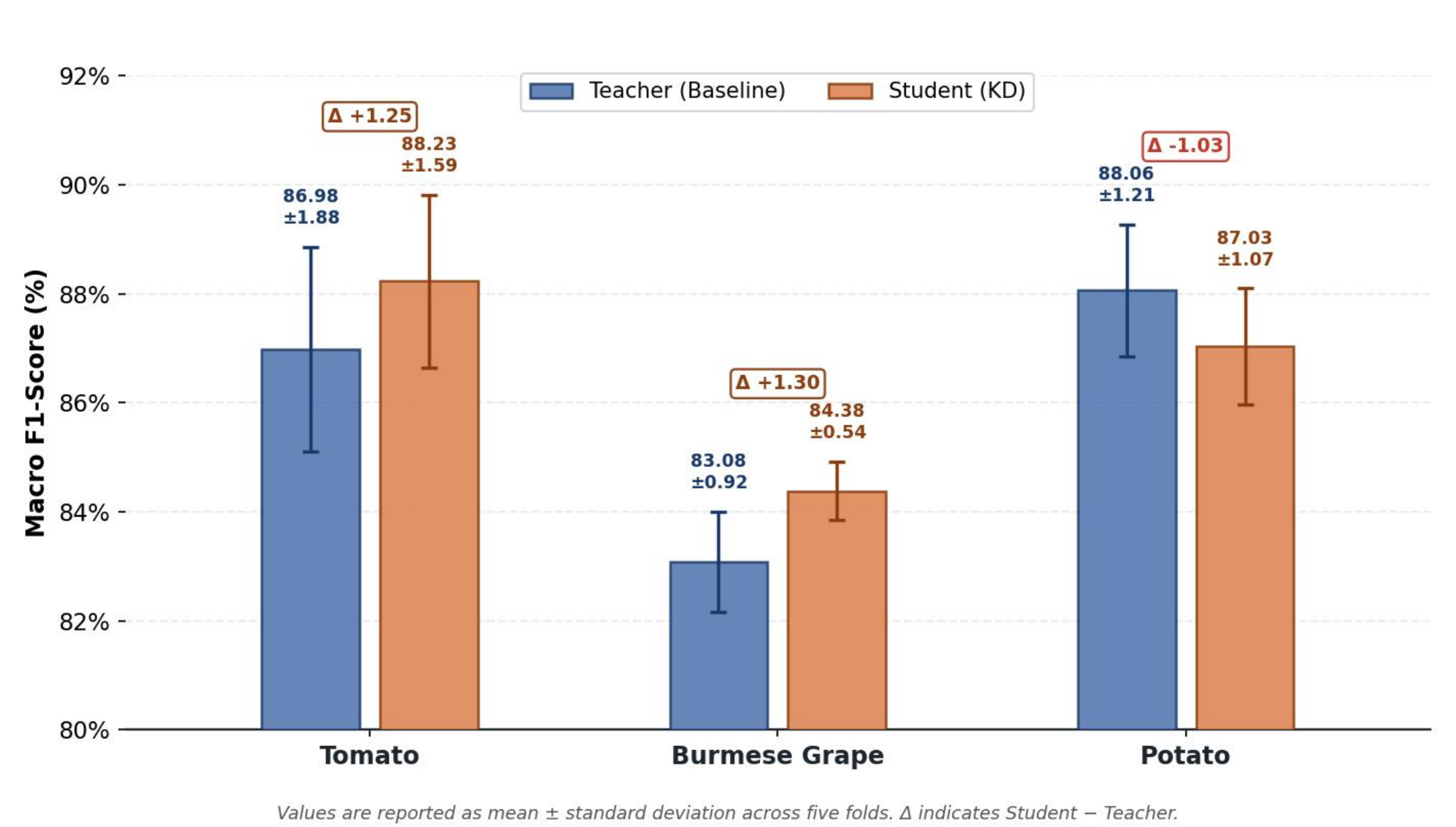}
\captionof{figure}{Macro F1-score comparison between teacher and distilled student across the three datasets.}
\label{fig:compare-f1}
\end{center}

\FloatBarrier

\subsubsection{Model Efficiency Analysis}
\label{subsec:efficiency}

Table~\ref{tab:model-efficiency} compares the ViT-Base teacher and the distilled student across three datasets. The student achieves substantial efficiency gains, reducing the parameter count by approximately 172$\times$ and computational cost by 47.57$\times$ in terms of GFLOPs. These reductions are consistent across all datasets, indicating that the proposed distillation framework produces a stable and architecture-independent compression effect.

In addition to model size and complexity, inference latency is significantly improved. The student achieves speed-ups ranging from 18.16$\times$ on Potato to 22.04$\times$ on Tomato, reducing inference time from approximately 227 ms to around 10–12.5 ms. This level of acceleration enables near real-time inference on CPU-based systems, which is particularly important for deployment in resource-constrained agricultural environments.

Despite this aggressive compression, the student maintains competitive predictive performance, achieving 85.77$\pm$0.78\% accuracy on Tomato, 86.12$\pm$0.46\% on Burmese Grape, and 87.97$\pm$0.48\% on Potato under cross-validation. The relatively small standard deviations further indicate stable training behavior across folds.

These results demonstrate that AgriKD effectively balances efficiency and accuracy, significantly reducing computational requirements while preserving strong predictive capability, making it well-suited for practical on-device agricultural applications.

\begin{center}
\captionof{table}{Model efficiency comparison: ViT-Base teacher versus distilled student across datasets}
\label{tab:model-efficiency}
\resizebox{\columnwidth}{!}{%
\begin{tabular}{l l c c c}
\hline\hline
\textbf{Dataset} & \textbf{Model} & \textbf{Params (M)} & \textbf{GFLOPs} & \textbf{CPU Inference (ms)} \\
\hline
\multirow{2}{*}{Tomato}
& ViT-Base Teacher & 86.199 & 17.60 & 227.00 \\
& Student          & 0.501  & 0.37  & 10.30  \\
\hline
\multirow{2}{*}{Burmese Grape}
& ViT-Base Teacher & 86.029 & 17.60 & 227.00 \\
& Student          & 0.499  & 0.37  & 12.00  \\
\hline
\multirow{2}{*}{Potato}
& ViT-Base Teacher & 86.031 & 17.60 & 227.00 \\
& Student          & 0.500  & 0.37  & 12.50  \\
\hline
\rowcolor[gray]{.95}
\multicolumn{2}{l}{\textbf{Avg. Reduction (Teacher $\rightarrow$ Student)}}
& \textbf{172.17$\times$} & \textbf{47.57$\times$} & \textbf{19.57$\times$} \\
\hline\hline
\end{tabular}}
\end{center}

\subsection{Ablation of Distillation Objectives}

To analyse how different distillation objectives contribute to the student model, we perform a loss-level ablation study.
Starting from a cross-entropy (CE) baseline, we progressively incorporate three types of supervision:

(i) logits-based (response-level),

(ii) relation-based (correlation-level), and

(iii) feature-based (representation-level) distillation.

We conduct the ablation on two representative datasets with different characteristics:
Tomato, which is relatively balanced, and Potato, which exhibits stronger class imbalance.
This selection allows us to examine whether the contribution of each distillation objective is consistent across different data distributions.

Tables~\ref{tab:tomato-ablation-new} and \ref{tab:potato-ablation-new} report the ablation results on the Tomato and Potato datasets, respectively.
Rows are ordered by progressively adding distillation objectives to isolate both individual and joint contributions.

The ablation is structured around three families of supervision:

CE + Logit corresponds to response-based distillation,

CE + Relation captures relation-based distillation, and

CE + Proj1 + Proj2 represents feature-based distillation.

We further evaluate pairwise combinations and the full objective to examine interaction effects.

\begin{center}
\captionof{table}{Ablation study on the Tomato Leaf Disease dataset}
\label{tab:tomato-ablation-new}
\resizebox{\columnwidth}{!}{%
\begin{tabular}{lcc}
\hline\hline
\textbf{Loss Components}
& \textbf{Accuracy (\%)}
& \textbf{F1-Score (\%)} \\
\hline
Cross Entropy (CE)
& 78.93 & 81.23 \\
CE + Proj1 + Proj2 (Feature-based)
& 83.88 & 86.00 \\
CE + Relation (Relation-based)
& 85.12 & 87.59 \\
CE + Logits (Logits-based)
& 85.12 & 88.20 \\
CE + Logits + Relation
& 85.95 & 88.82 \\
CE + Logits + Proj1 + Proj2
& 85.95 & 89.18 \\
CE + Logits + Relation + Proj1 + Proj2 (Full)
& \textbf{87.19} & \textbf{90.35} \\
\hline\hline
\end{tabular}}
\end{center}

\begin{center}
\captionof{table}{Ablation study on the Potato Leaf Disease dataset}
\label{tab:potato-ablation-new}
\resizebox{\columnwidth}{!}{%
\begin{tabular}{lcc}
\hline\hline
\textbf{Loss Components}
& \textbf{Accuracy (\%)}
& \textbf{F1-Score (\%)} \\
\hline
Cross Entropy (CE)
& 86.15 & 85.71 \\
CE + Proj1 + Proj2 (Feature-based)
& 87.45 & 86.66 \\
CE + Relation (Relation-based)
& 88.10 & 87.17 \\
CE + Logits (Logits-based)
& 88.53 & 88.60 \\
CE + Logits + Relation
& 88.96 & 89.23 \\
CE + Logits + Proj1 + Proj2
& 88.74 & 89.07 \\
CE + Logits + Relation + Proj1 + Proj2 (Full)
& \textbf{90.69} & \textbf{90.25} \\
\hline\hline
\end{tabular}}
\end{center}

Four consistent findings emerge from the ablation results, revealing how different distillation objectives contribute to the student model.

\paragraph{Logits-based distillation provides the strongest prediction-level guidance}

Adding logits-based distillation to the cross-entropy anchor produces the largest single-component gain across both datasets: $+6.97$ percentage points in F1-score on Tomato (81.23\% $\to$ 88.20\%) and $+2.89$ percentage points on Potato (85.71\% $\to$ 88.60\%). This result is expected because vanilla KD transfers more than the ground-truth class label: the softened teacher distribution encodes similarity among disease categories, including visually related classes that may share subtle lesion patterns, color changes, or texture degradation. Such dark knowledge is particularly useful in leaf disease classification, where disease symptoms can be fine-grained and partially overlapping. The larger gain on Tomato is consistent with its higher number of classes, where inter-class similarity information provides richer supervisory information than hard labels alone.

\paragraph{Relation-based distillation preserves the teacher's confusion structure}
Even in isolation, CE + Relation achieves 87.59\% F1-score on Tomato and 87.17\% on Potato, demonstrating that relational structure carries independent discriminative value. Mechanistically, the relation loss does not force the student to exactly reproduce the teacher's probability values; instead, it preserves the correlation structure of the teacher's predictions across classes and samples. This is important when the compact CNN student cannot fully match the ViT teacher distribution, but can still learn how the teacher ranks classes and how different samples relate to each disease category. When added on top of logits-based distillation, the relation loss contributes consistent additional gains on both datasets, suggesting that preserving inter-class and intra-class correlation provides complementary supervision beyond soft-label matching. This advantage becomes particularly important in cross-architecture distillation, where the representational gap between ViT and CNN models makes exact probability matching less reliable.

\paragraph{Projection losses refine spatial alignment across heterogeneous architectures} When applied without logits or relation losses, CE + Proj1 + Proj2 achieves 86.00\% F1-score on Tomato and 86.66\% on Potato, indicating that feature-level alignment is useful but less effective as a standalone supervisory objective than prediction-level distillation. This behavior is consistent with the role of the projection losses: they address the spatial mismatch between the ViT teacher's patch-token representations and the CNN student's convolutional feature maps. For leaf disease recognition, this alignment is meaningful because diagnostic evidence such as lesions, necrotic regions, mildew patterns, and discoloration is spatially localized. However, feature projection alone does not provide the same direct class-level guidance as logits or relation losses. When added to the logits + relation combination, the projection terms provide the final performance gains by refining the student's spatial representation after the prediction-level structure has already been established. This is especially critical in the ViT-to-CNN setting, where transformer tokens encode global context while convolutional features emphasize locality, requiring explicit alignment to bridge their representational differences.

\paragraph{The full objective combines complementary gains} The full four-component objective achieves the best F1-score on both datasets: 90.35\% on Tomato and 90.25\% on Potato, corresponding to improvements of $+9.12$ and $+4.54$ percentage points over cross-entropy-only training, respectively. Across the progressive ablation sequence, each added component produces a positive gain, indicating that the different supervision objectives contribute complementary information rather than acting as interchangeable objectives. In particular, logits-based distillation provides the strongest prediction-level guidance, relation-based distillation further preserves the teacher's correlation structure, and projection-based losses refine the spatial alignment between ViT token representations and CNN feature maps. These results support the design principle of combining response-based, relation-based, and feature-based supervision within a unified cross-architecture distillation objective. This complementarity is particularly beneficial in cross-architecture distillation, where no single supervision objective is sufficient to bridge the gap between transformer-based and convolutional representations.
\FloatBarrier

\subsection{Visual Interpretability via Grad-CAM}
\label{subsec:gradcam}

To further examine whether the distilled student learns diagnostically meaningful representations, we visualise model attention using Grad-CAM.

\vspace{0.2cm}
\textbf{
}{i) Grad-CAM analysis on Burmese dataset}
\vspace{0.2cm}

\begin{figure*}[h]
\centering
\setlength{\tabcolsep}{3pt}
\renewcommand{\arraystretch}{1.2}
\begin{tabular}{cccc}
\includegraphics[width=0.23\textwidth]{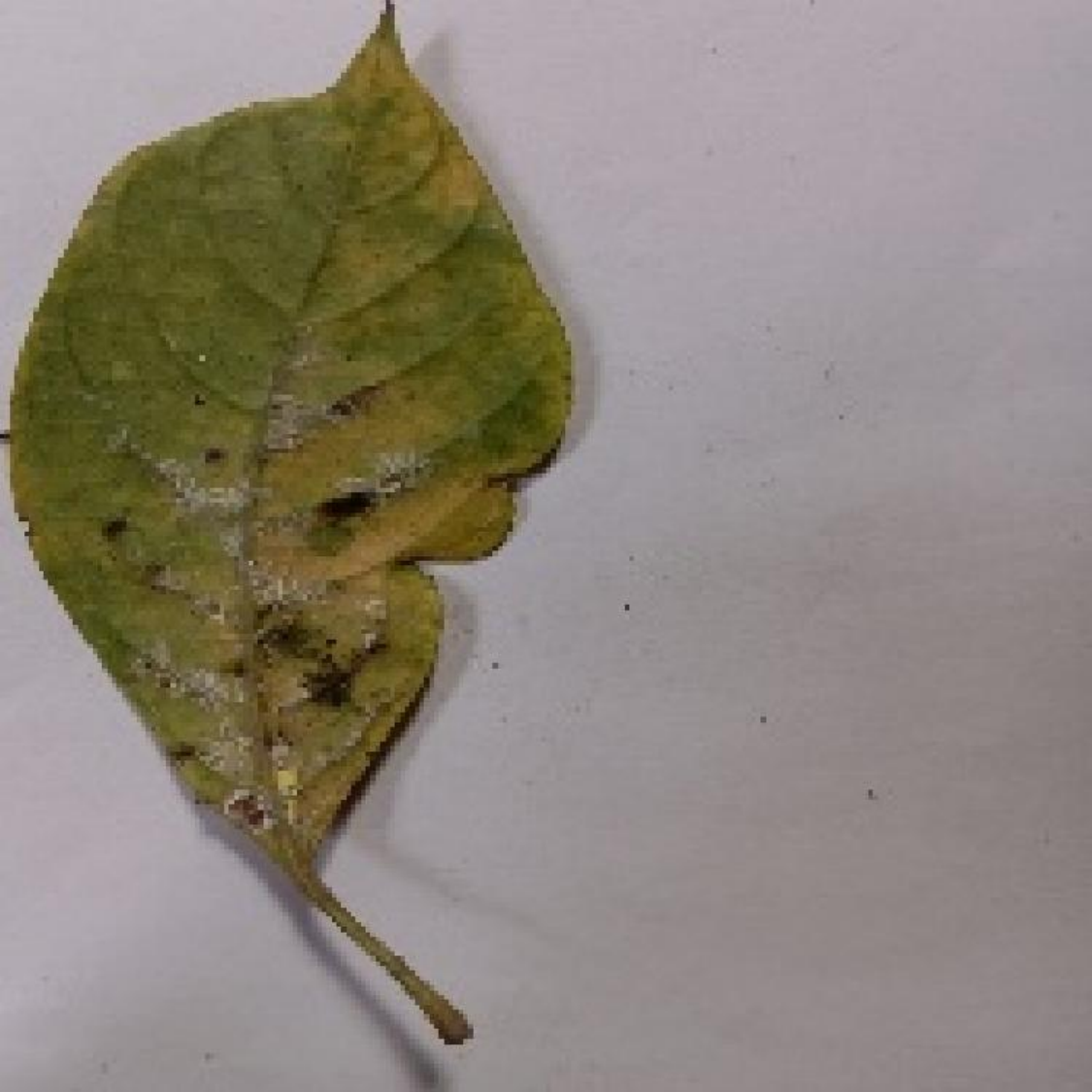} &
\includegraphics[width=0.23\textwidth]{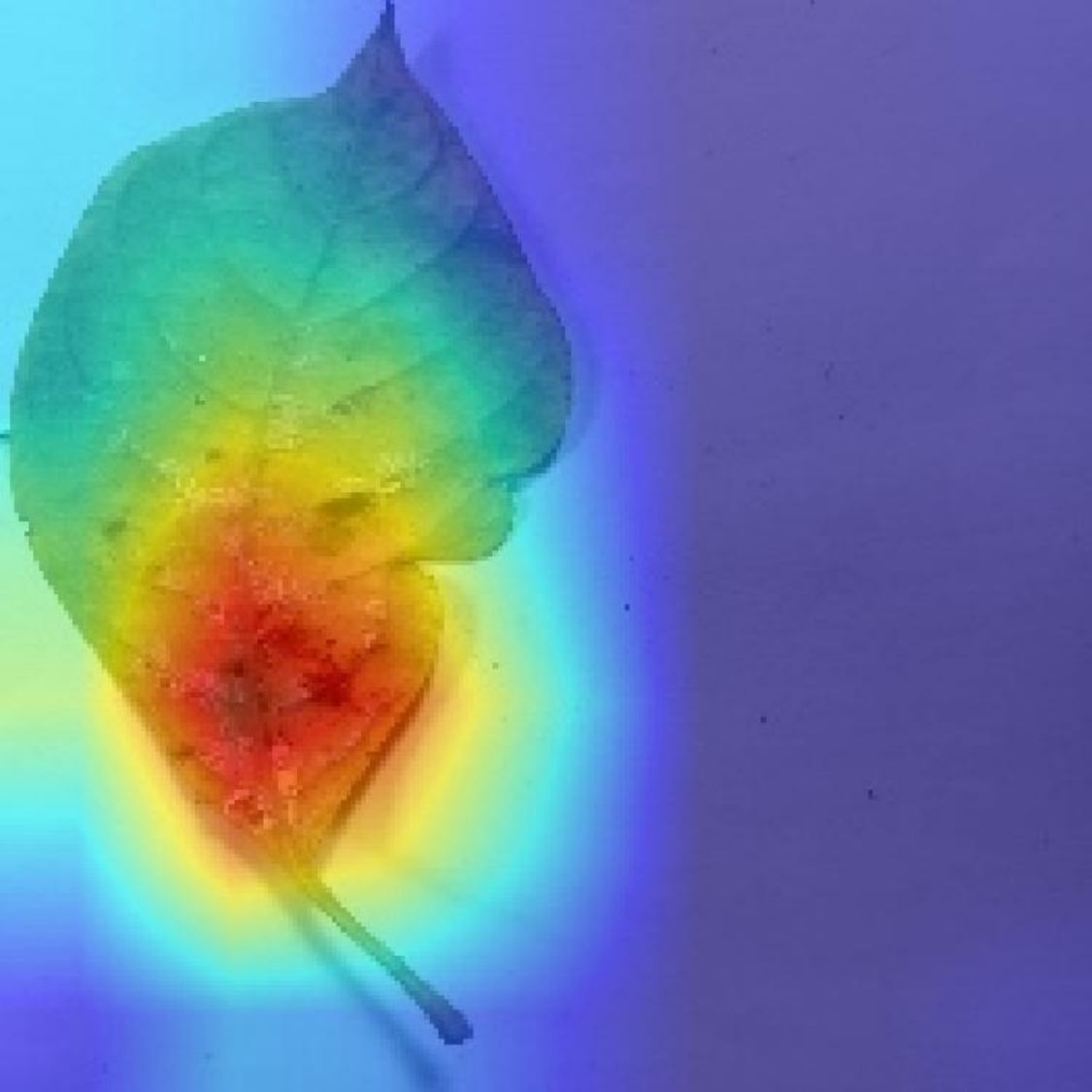} &
\includegraphics[width=0.23\textwidth]{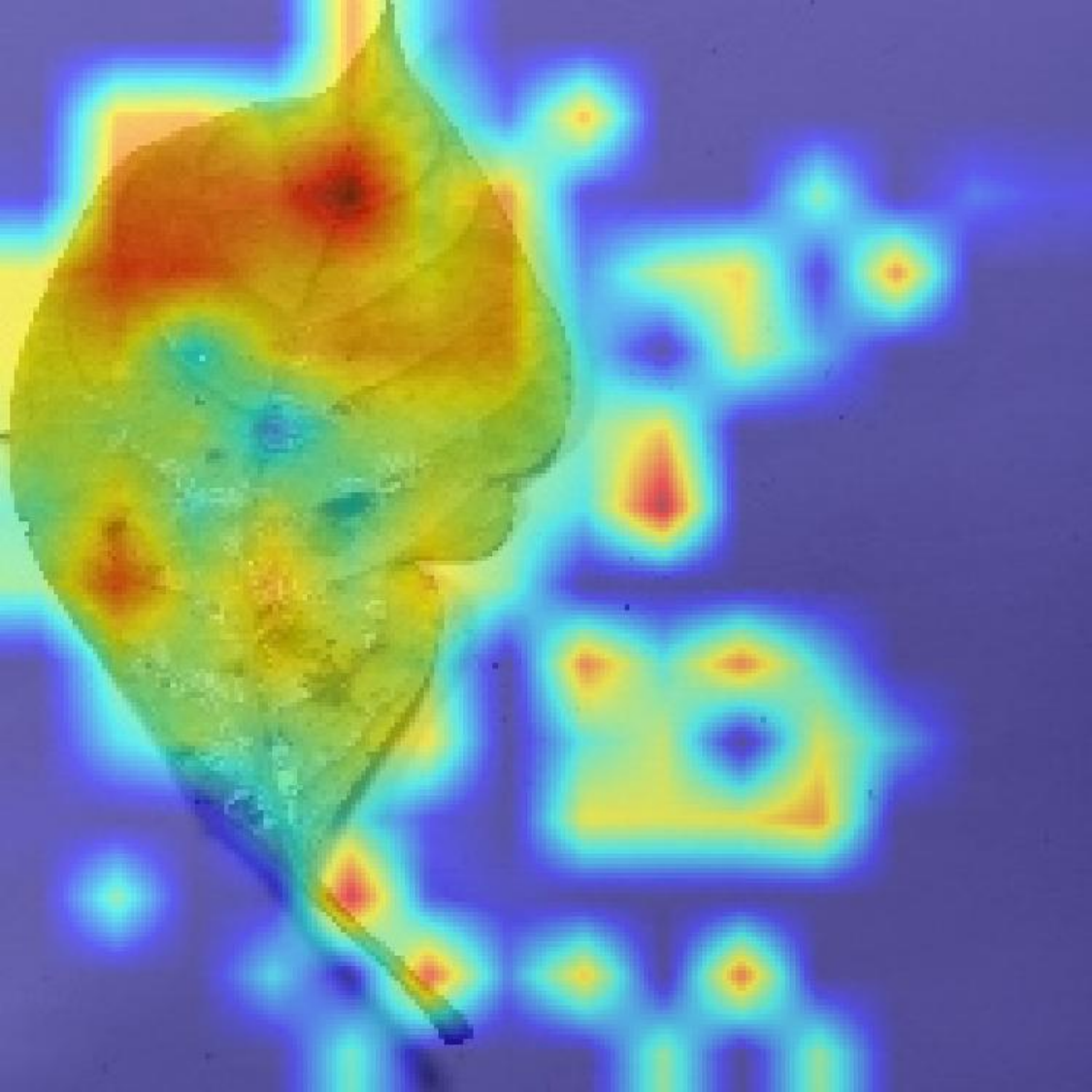} &
\includegraphics[width=0.23\textwidth]{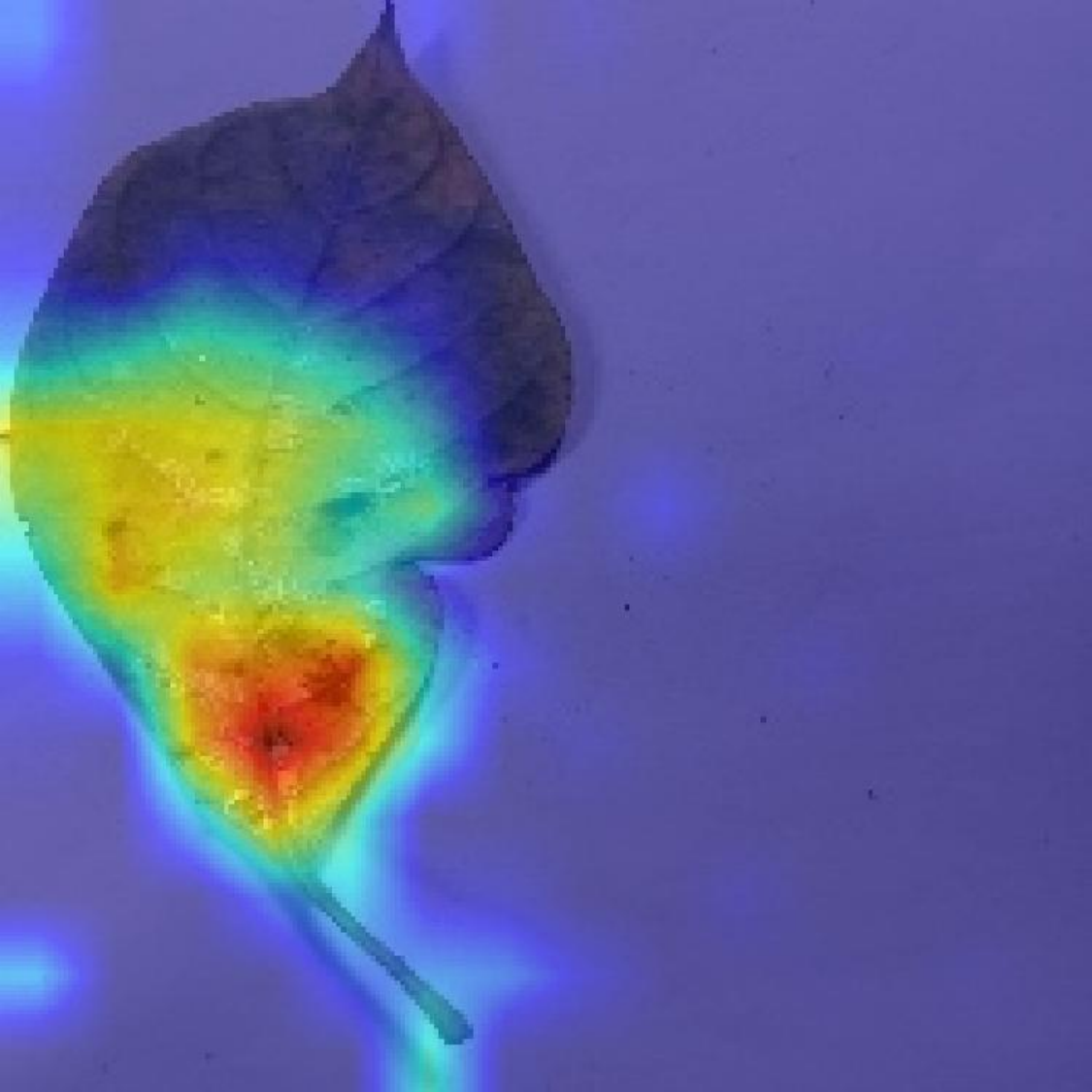} \\[-2pt]
\small Original (Powdery Mildew) & \small MobileNetV2 & \small Teacher & \small Student \\[8pt]
\includegraphics[width=0.23\textwidth]{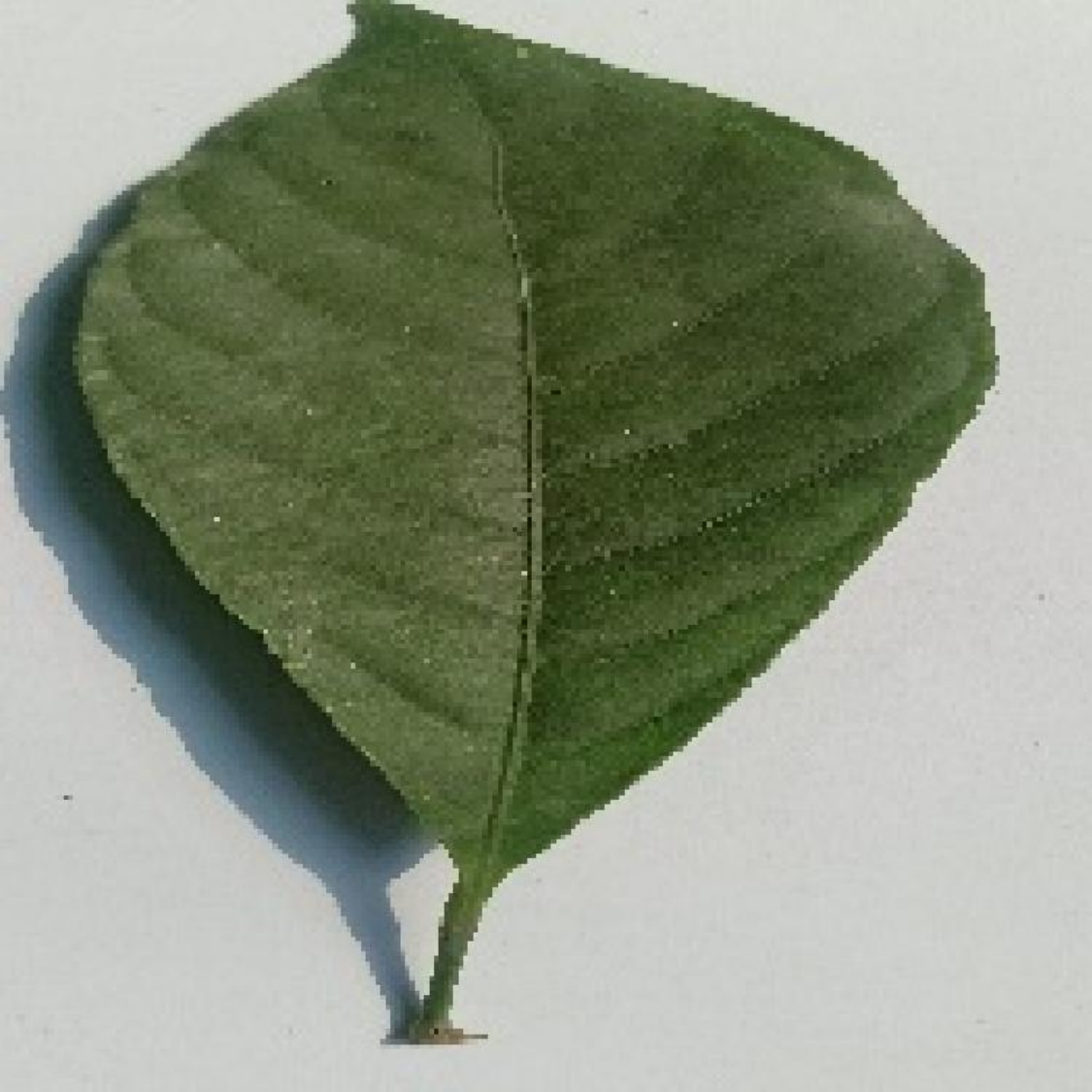} &
\includegraphics[width=0.23\textwidth]{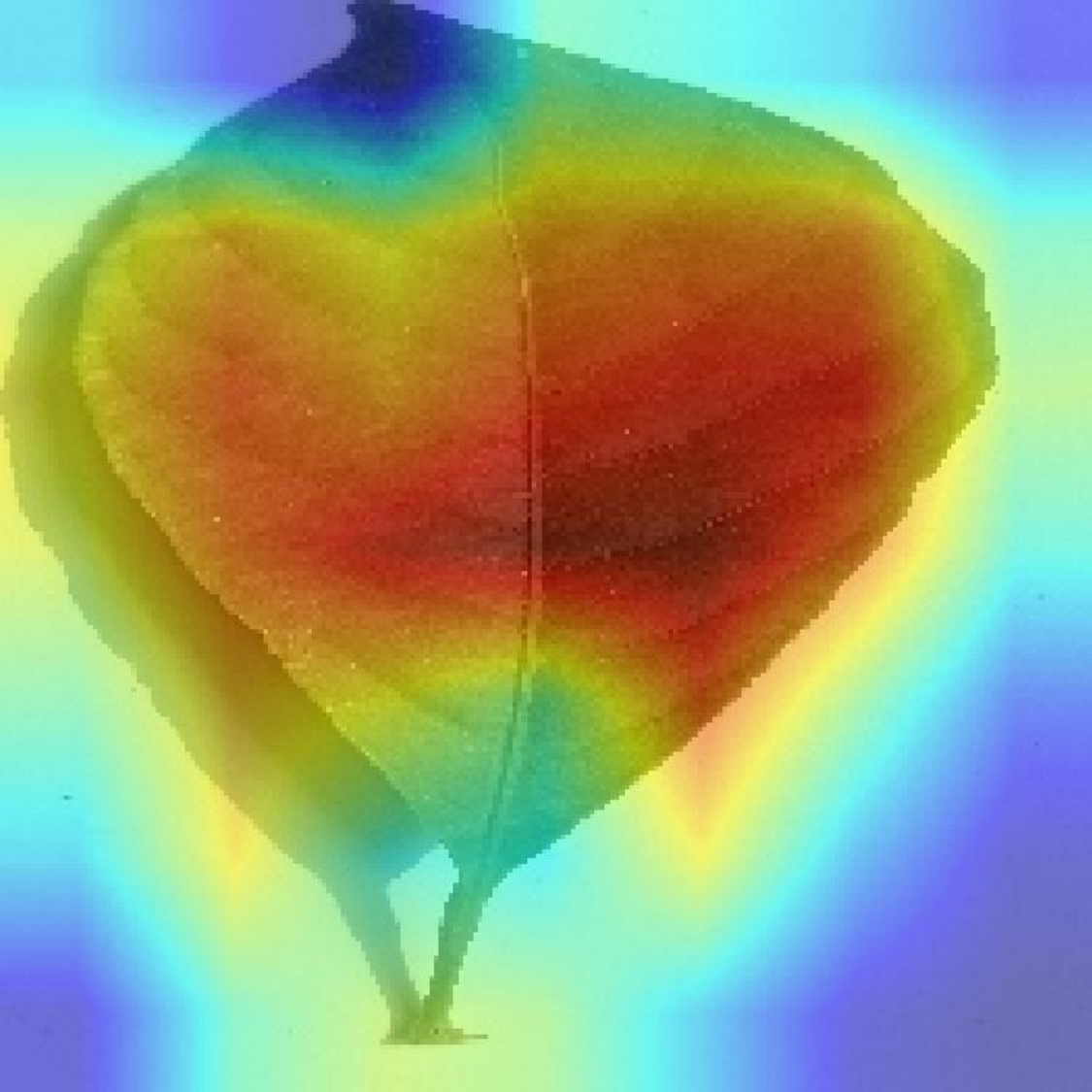} &
\includegraphics[width=0.23\textwidth]{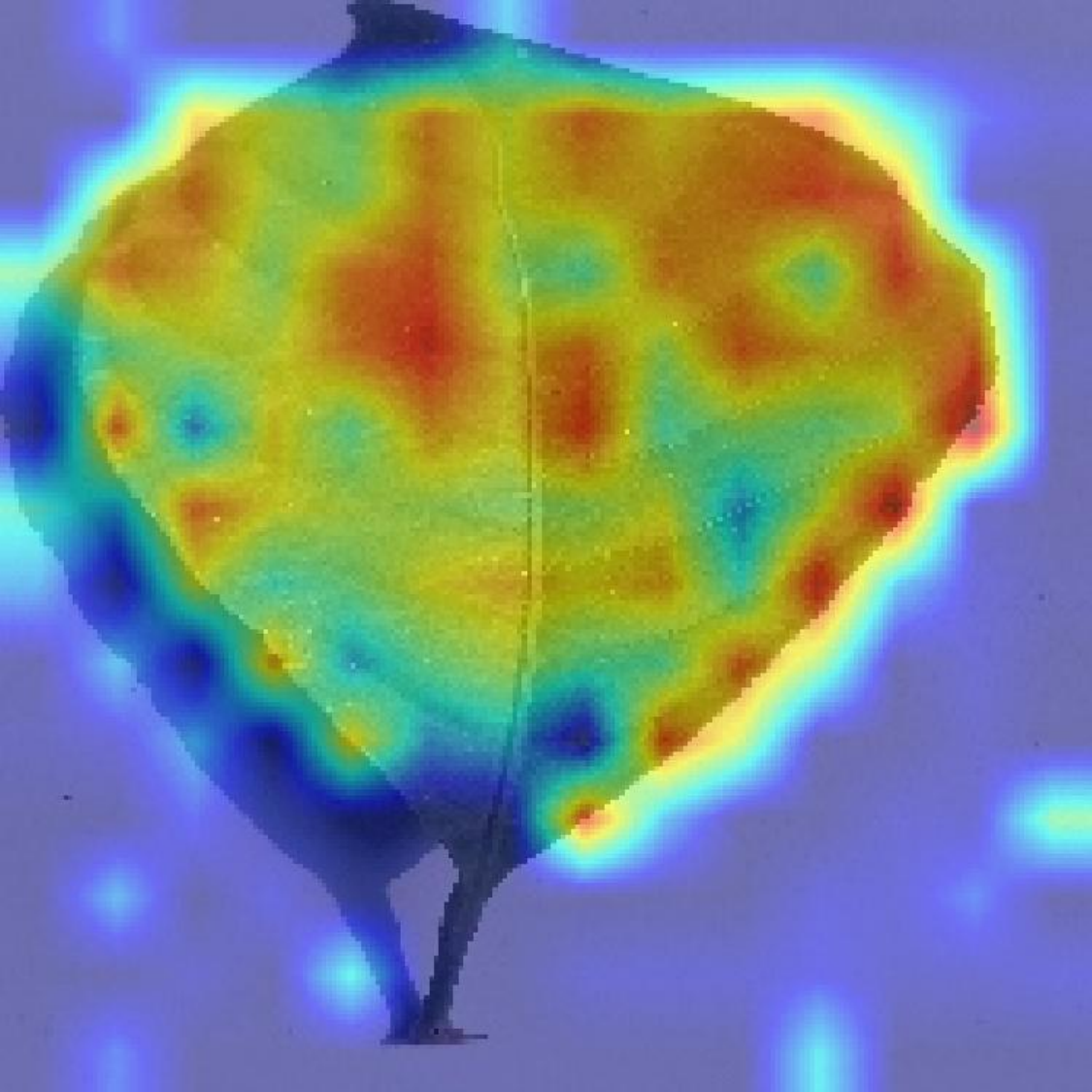} &
\includegraphics[width=0.23\textwidth]{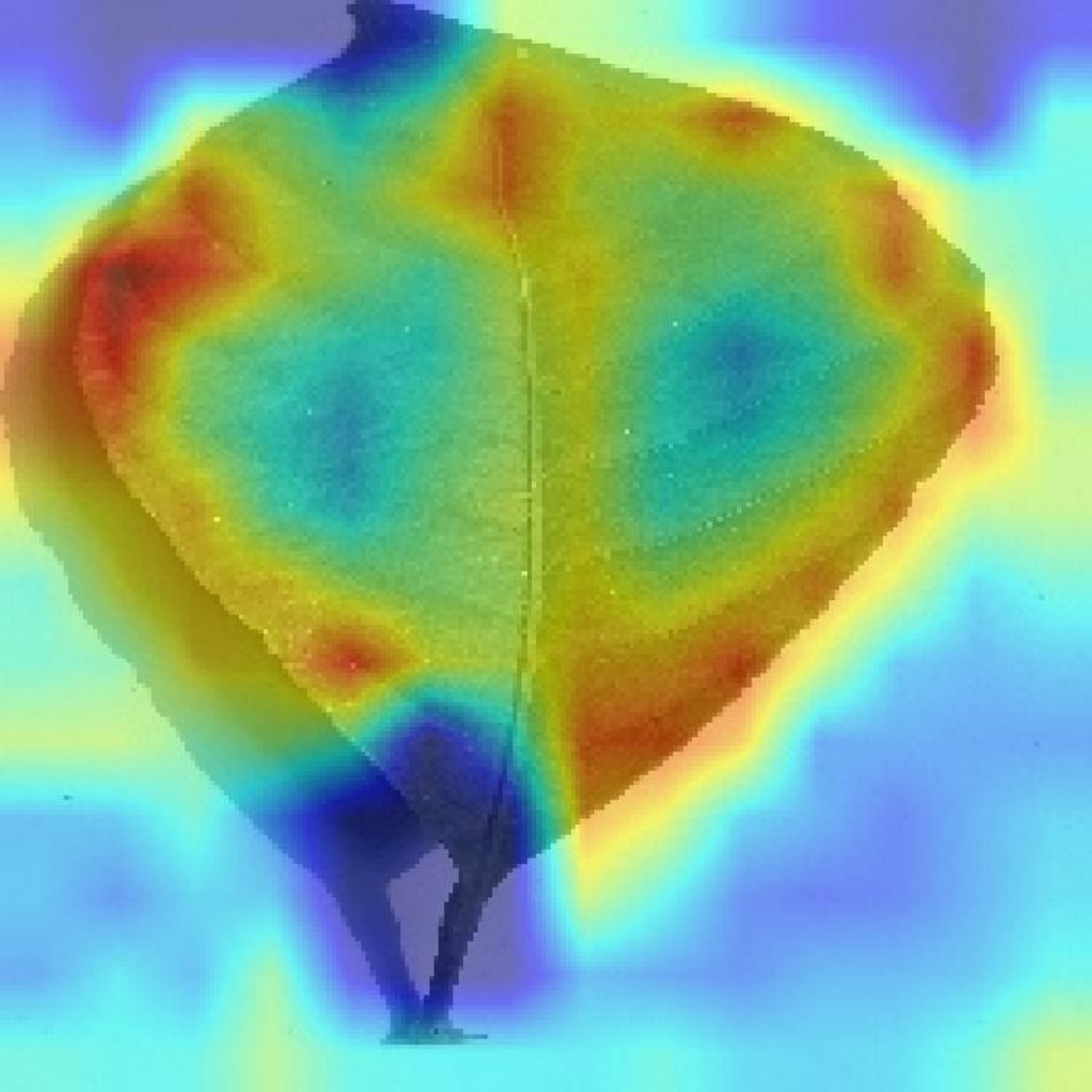} \\[-2pt]
\small Original (Healthy) & \small MobileNetV2 & \small Teacher & \small Student
\end{tabular}
\caption{Grad-CAM on Burmese Grape dataset}
\label{fig:gradcam_burmese}
\end{figure*}

Figure~\ref{fig:gradcam_burmese} compares activation maps of the baseline MobileNetV2, the ViT teacher, and the distilled student on representative samples on Burmese dataset.

On diseased samples (e.g., Powdery Mildew), the baseline CNN captures the lesion region but produces relatively diffuse attention.
The ViT teacher, while encoding global context, exhibits scattered and noisy activations across the leaf surface.
In contrast, the distilled student focuses more precisely on the infected regions, producing compact and well-localised attention maps.

On healthy samples, the baseline model shows uneven attention distribution, while the teacher again produces overly dispersed responses.
The distilled student, however, attends to the leaf structure in a more coherent and structured manner, covering relevant regions without introducing spurious activations.

These observations suggest that distillation improves not only predictive performance but also spatial interpretability.
In particular, the student learns to combine the global contextual knowledge of the teacher with the locality bias of CNNs, resulting in more accurate and diagnostically relevant attention.
This behaviour is consistent with the ablation findings, where logits-based supervision provides strong predictive guidance, while relation- and feature-based objectives refine structural and spatial alignment.

\vspace{0.2cm}
\textbf{{ii) Grad-CAM analysis on Tomato dataset}}
\vspace{0.2cm}

\begin{figure*}[h]
\centering
\setlength{\tabcolsep}{3pt}
\renewcommand{\arraystretch}{1.2}
\begin{tabular}{cccc}
\includegraphics[width=0.23\textwidth]{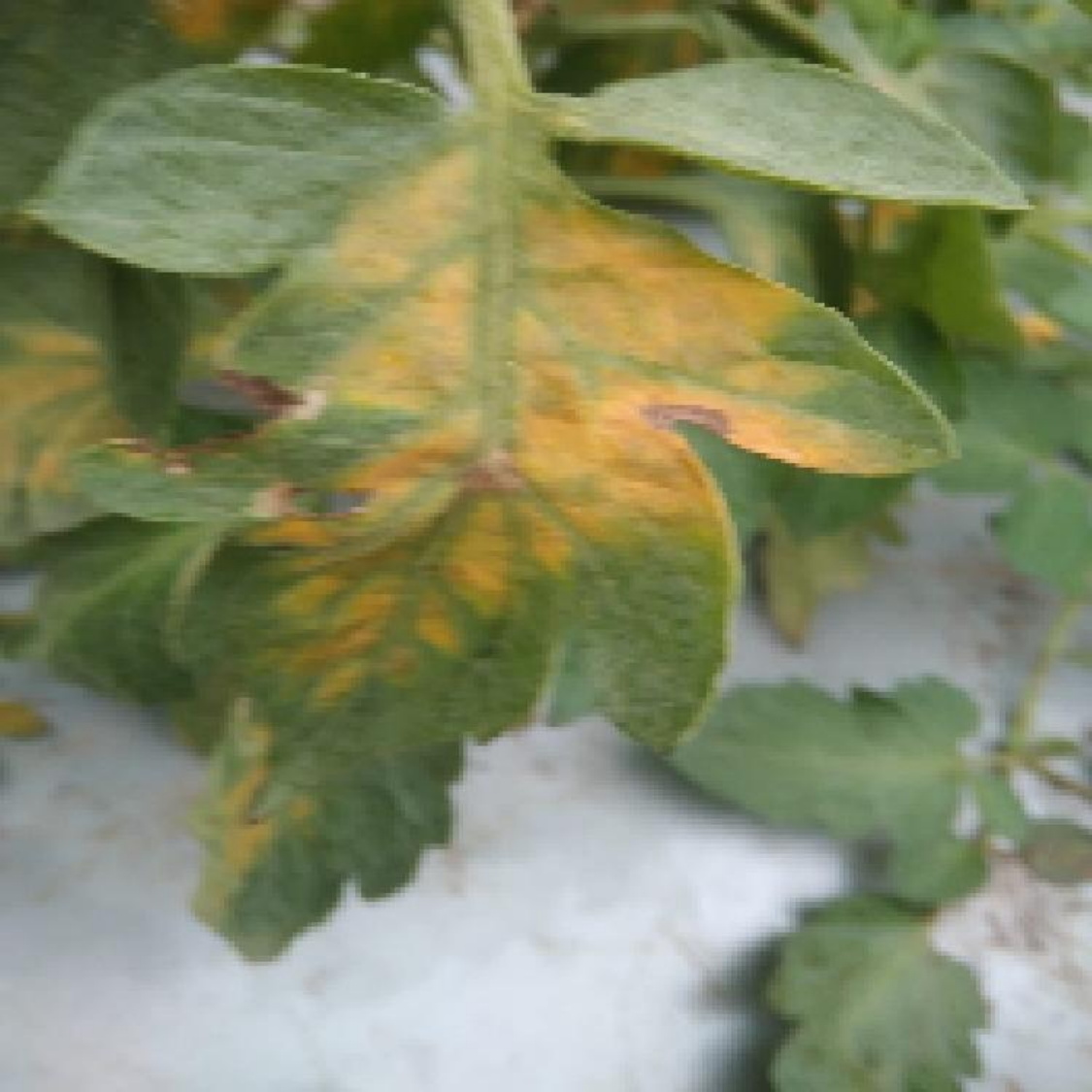} &
\includegraphics[width=0.23\textwidth]{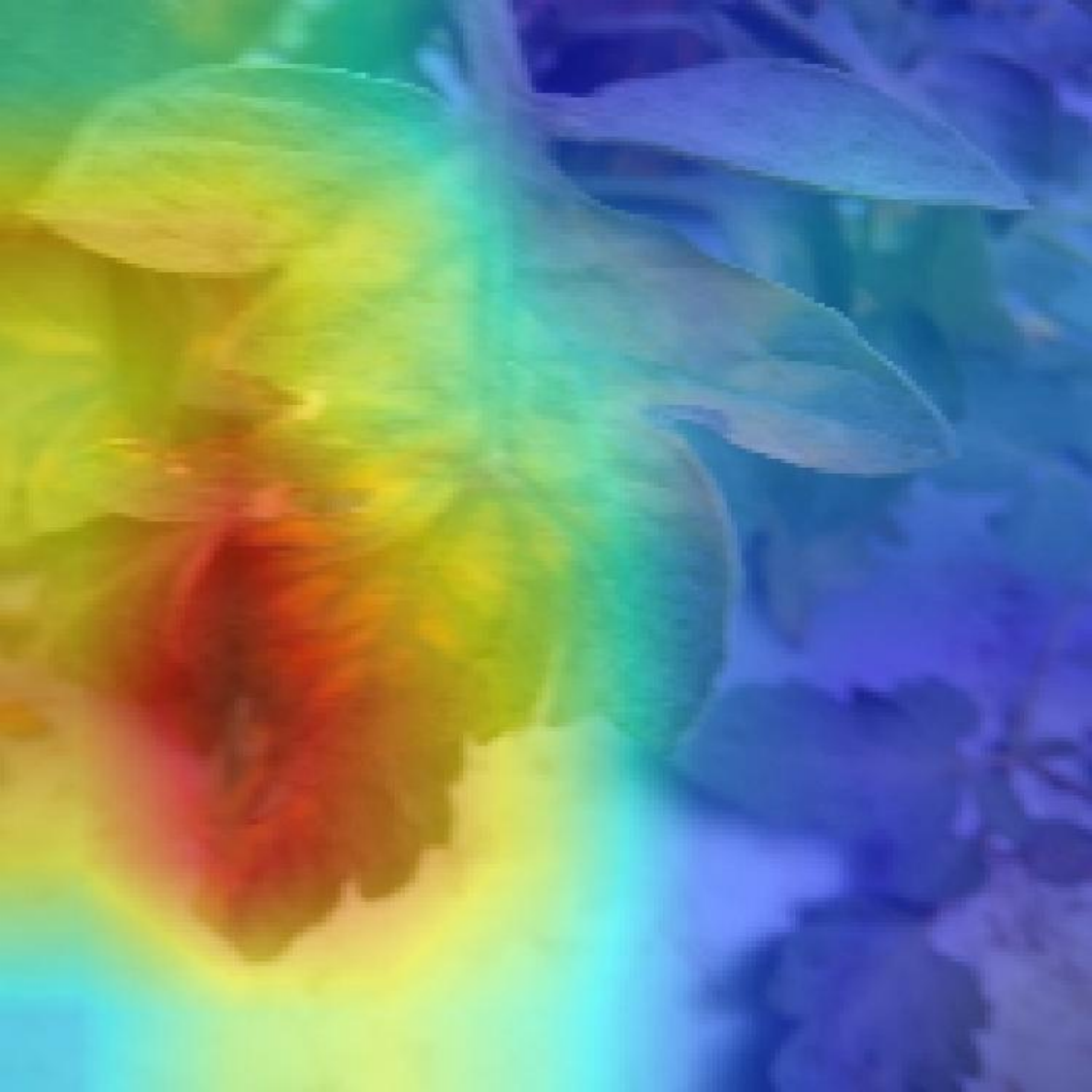} &
\includegraphics[width=0.23\textwidth]{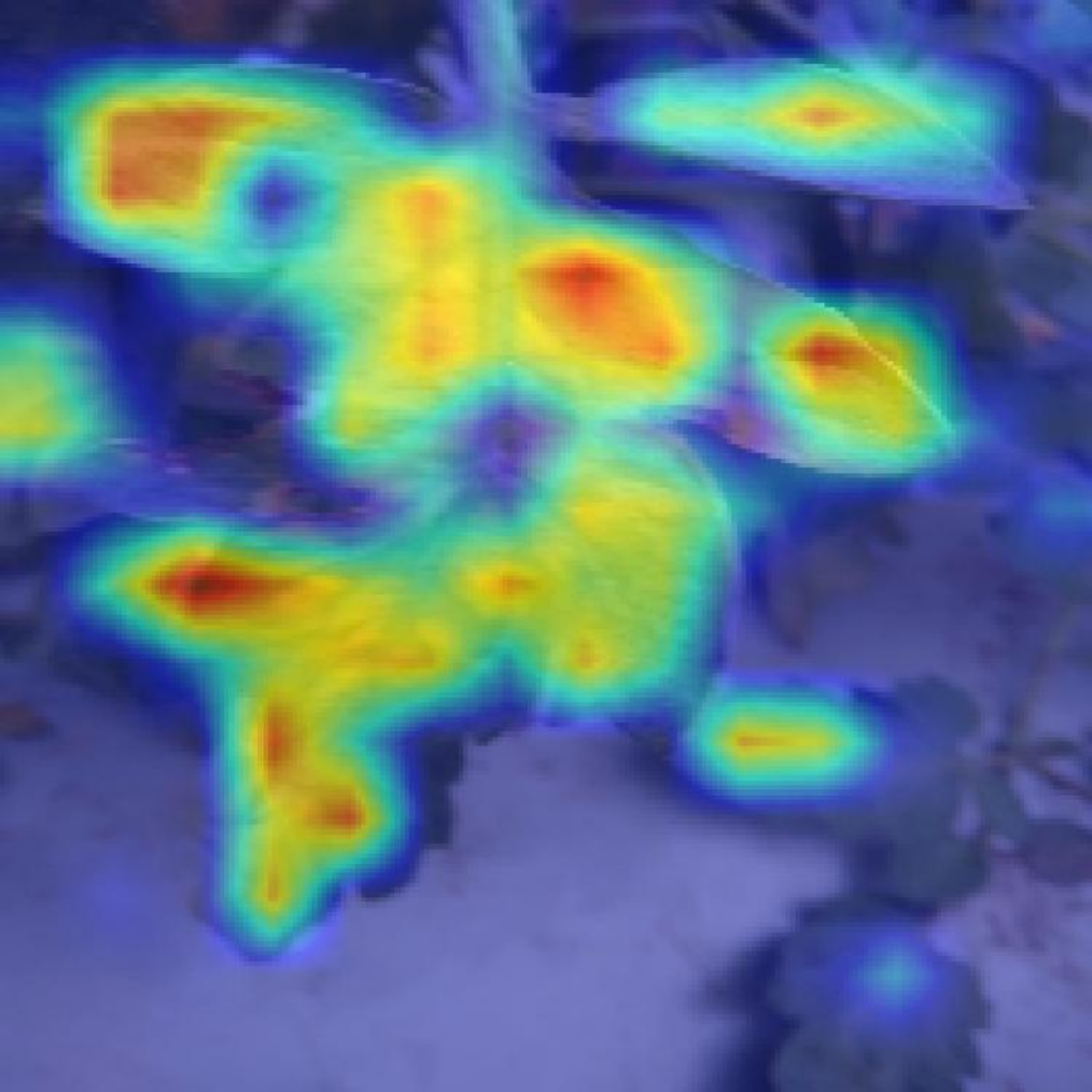} &
\includegraphics[width=0.23\textwidth]{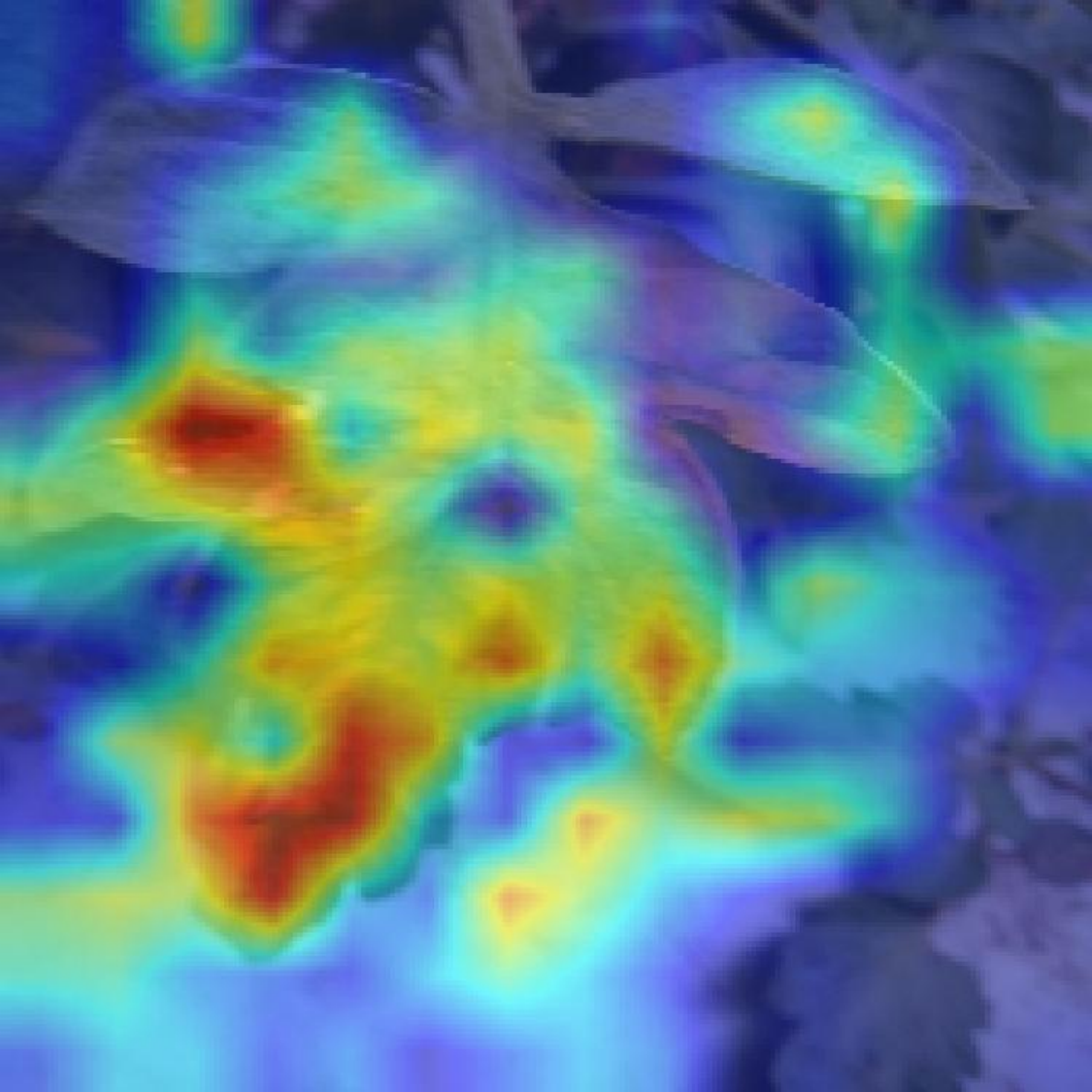} \\[-2pt]
\small Original (Bacterial Spot) & \small MobileNetV2 & \small Teacher & \small Student \\[8pt]
\includegraphics[width=0.23\textwidth]{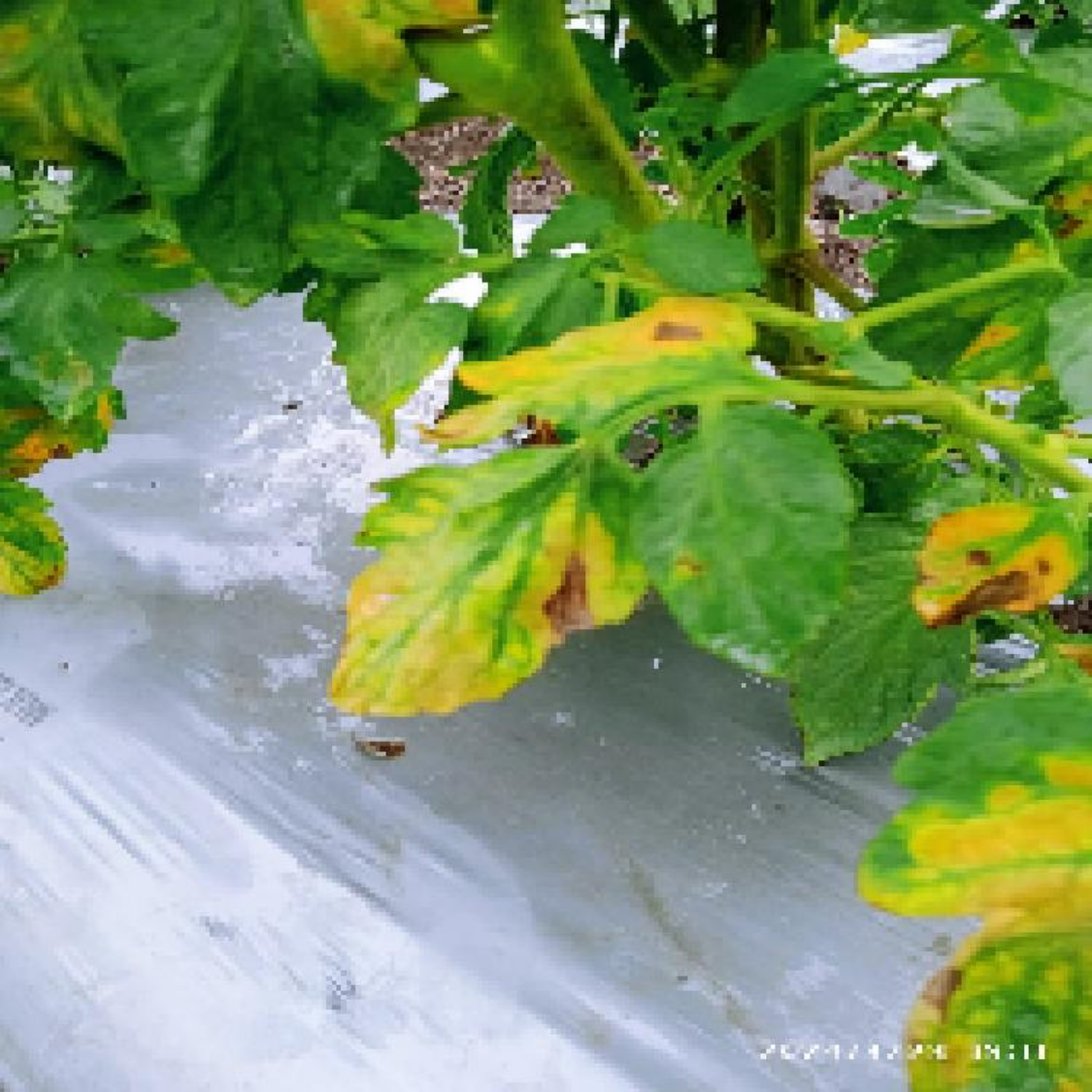} &
\includegraphics[width=0.23\textwidth]{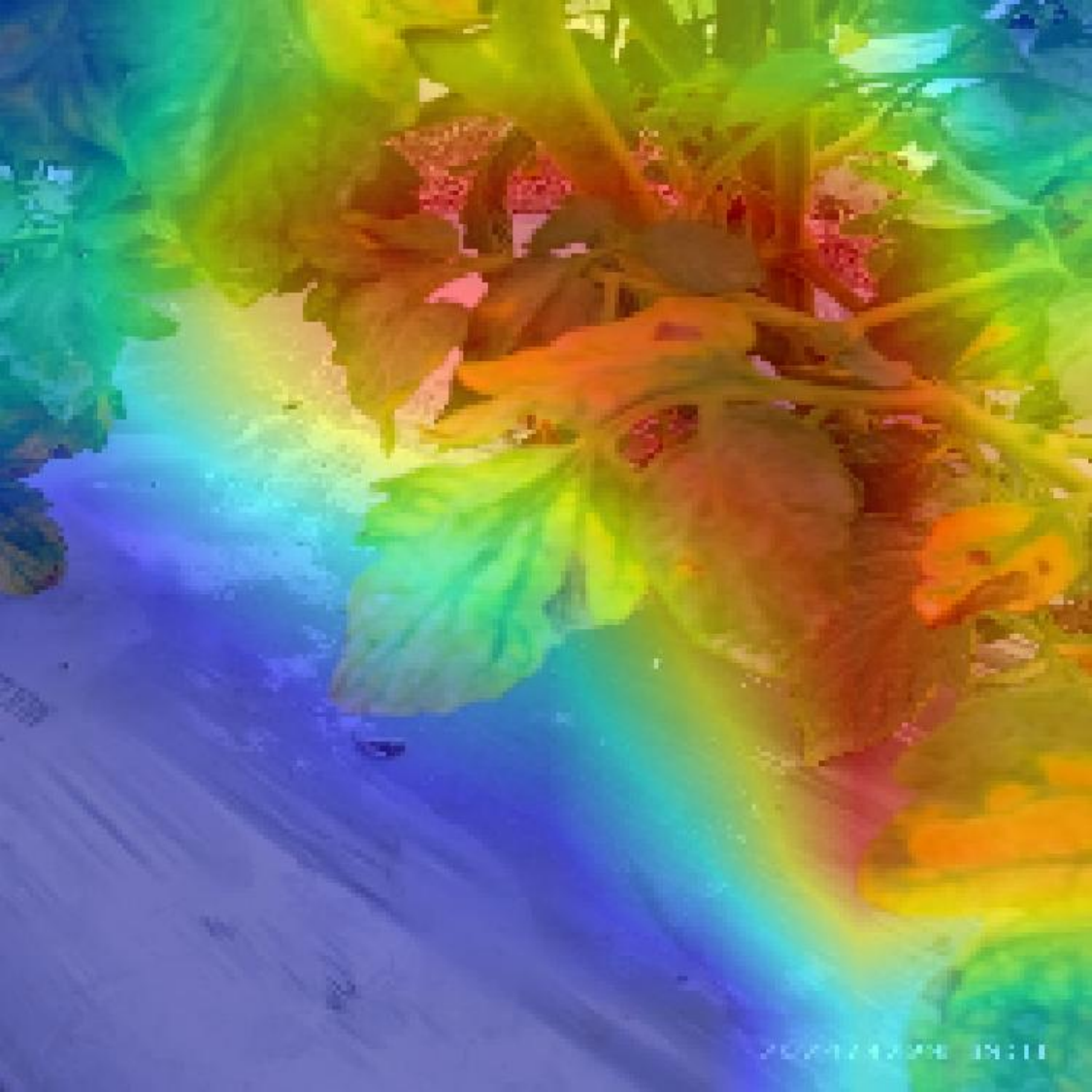} &
\includegraphics[width=0.23\textwidth]{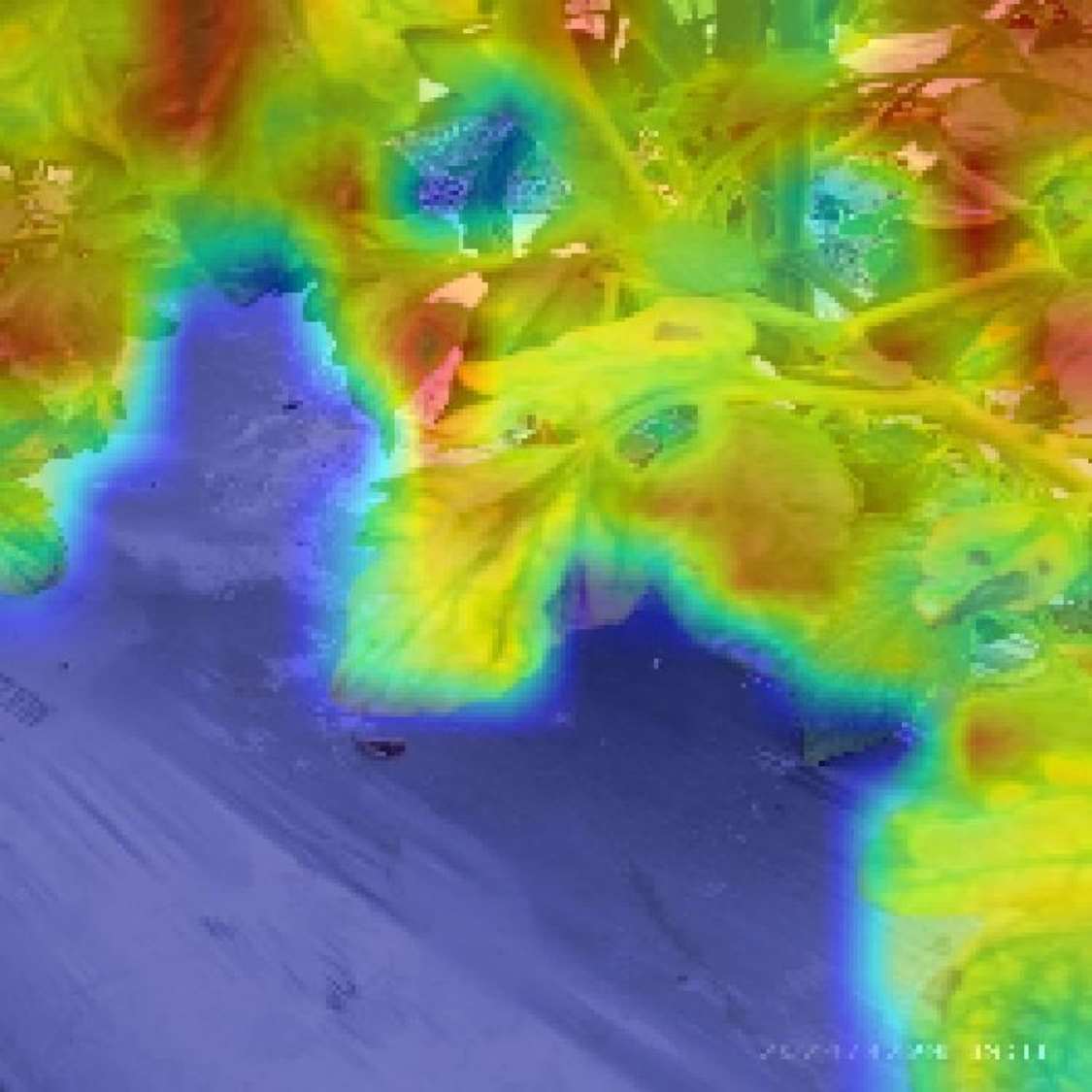} &
\includegraphics[width=0.23\textwidth]{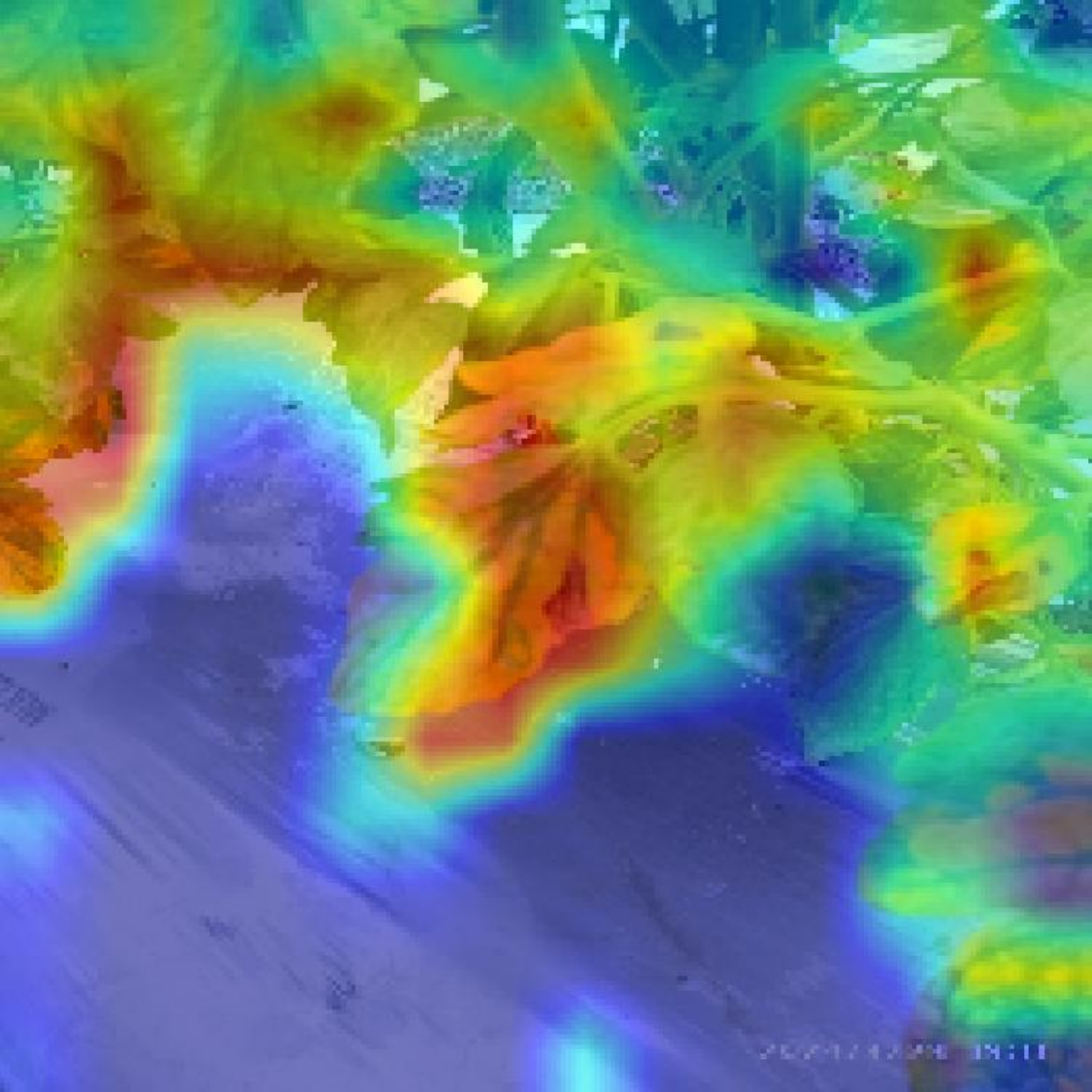} \\[-2pt]
\small Original (Leaf Mold) & \small MobileNetV2 & \small Teacher & \small Student
\end{tabular}
\caption{Grad-CAM on Tomato dataset}
\label{fig:gradcam_tomato}
\end{figure*}

Figure~\ref{fig:gradcam_tomato} presents Grad-CAM visualisations on representative Tomato samples.
On the Bacterial Spot case, the baseline CNN focuses on a single dominant region, missing several smaller lesions, while the ViT teacher produces fragmented and noisy attention across the leaf.
In contrast, the distilled student captures multiple infected regions with more coherent and selective attention, avoiding excessive activation on irrelevant areas.

A similar pattern is observed for Leaf Mold.
The baseline model exhibits biased and incomplete attention, whereas the teacher attends broadly, including background regions.
The distilled student achieves a better balance by covering the diseased leaf regions while suppressing spurious responses outside the target area.

These results indicate that distillation improves both coverage and selectivity of attention.
Compared to the CNN baseline, the student captures more comprehensive disease patterns, and compared to the ViT teacher, it produces more focused and less noisy activations.
This behaviour suggests that the student effectively integrates global contextual knowledge from the teacher with the locality bias of convolutional architectures.

\vspace{0.2cm}
\textbf{{iii) Grad-CAM analysis on Potato dataset}}
\vspace{0.2cm}

\begin{figure*}[h]
\centering
\setlength{\tabcolsep}{3pt}
\renewcommand{\arraystretch}{1.2}
\begin{tabular}{cccc}
\includegraphics[width=0.23\textwidth]{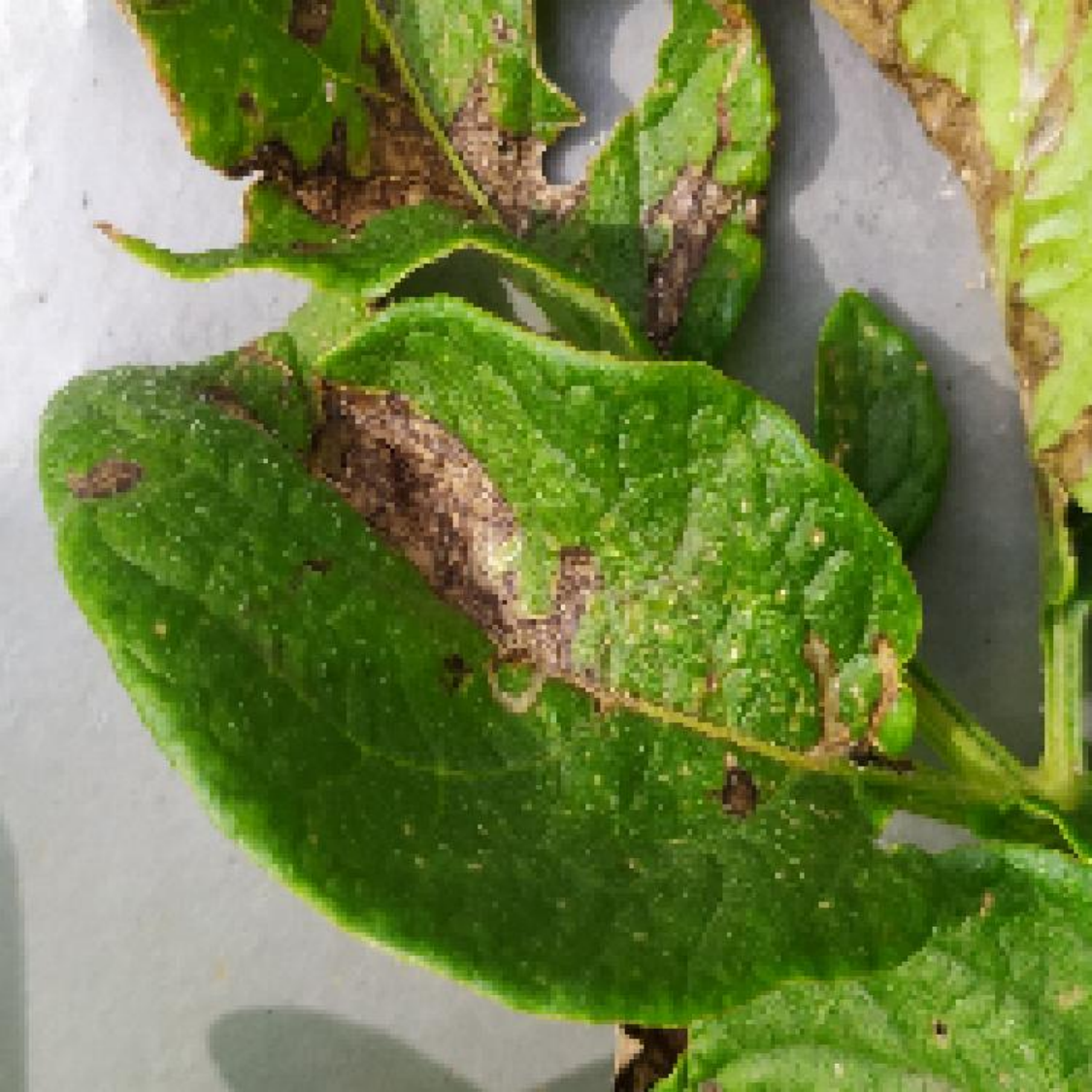} &
\includegraphics[width=0.23\textwidth]{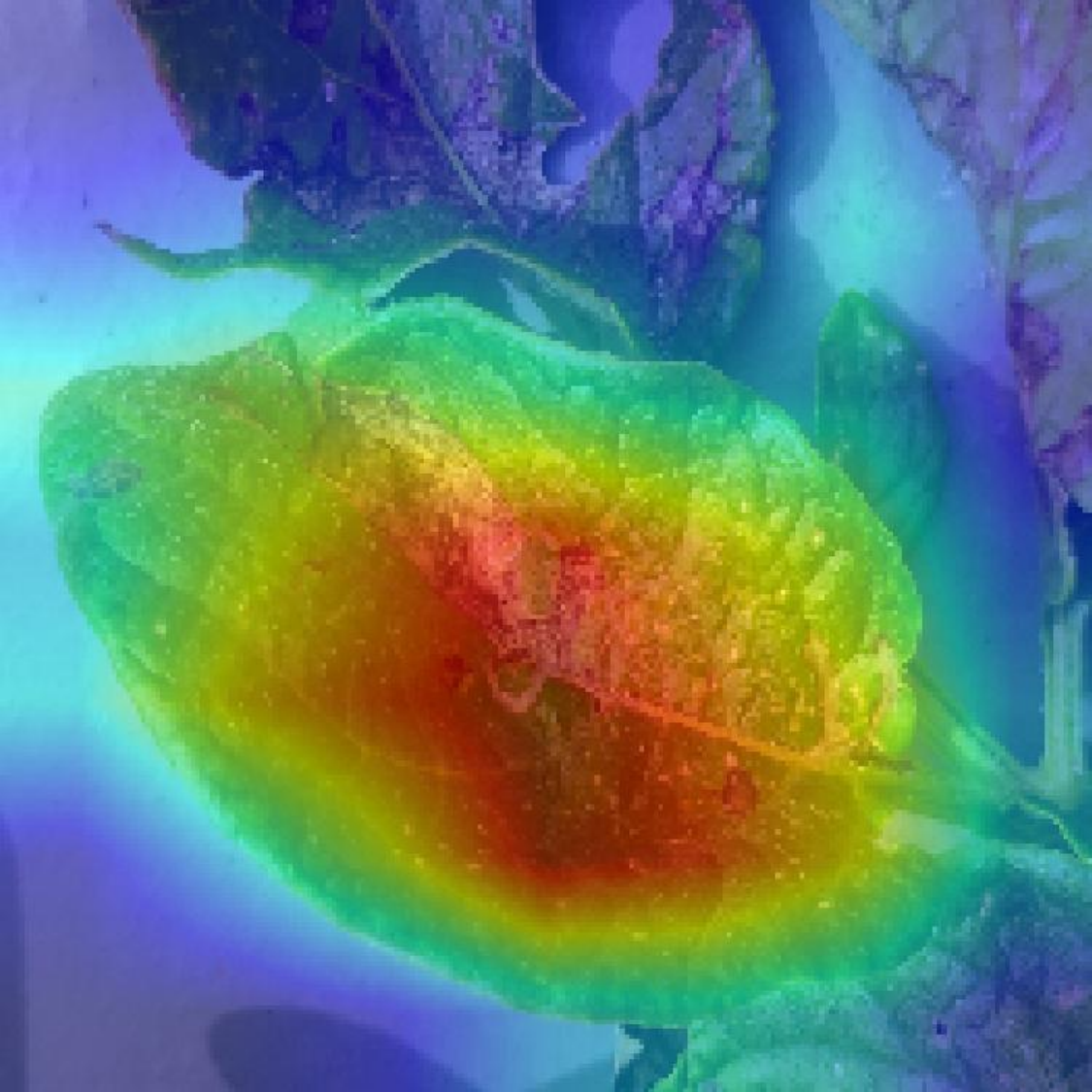} &
\includegraphics[width=0.23\textwidth]{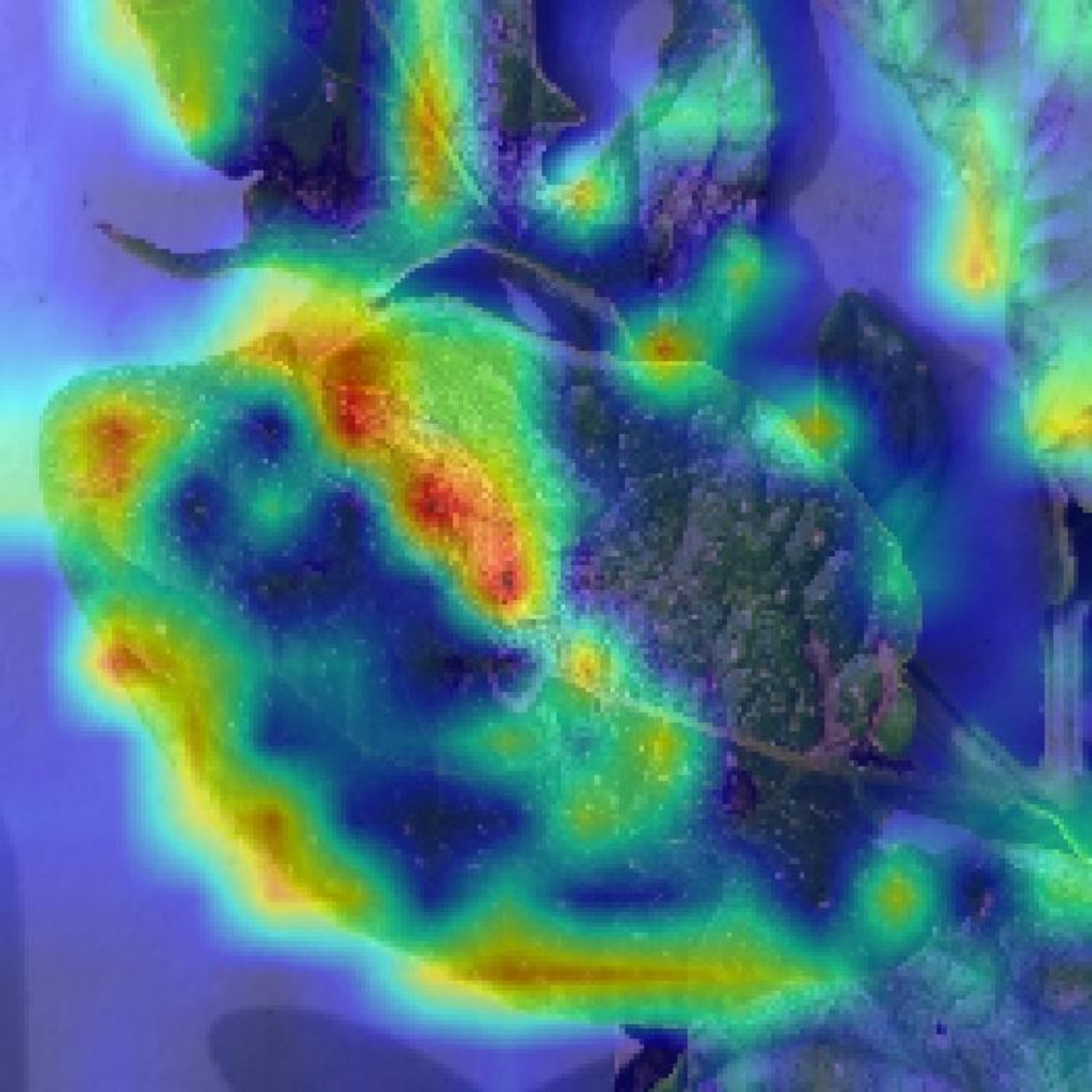} &
\includegraphics[width=0.23\textwidth]{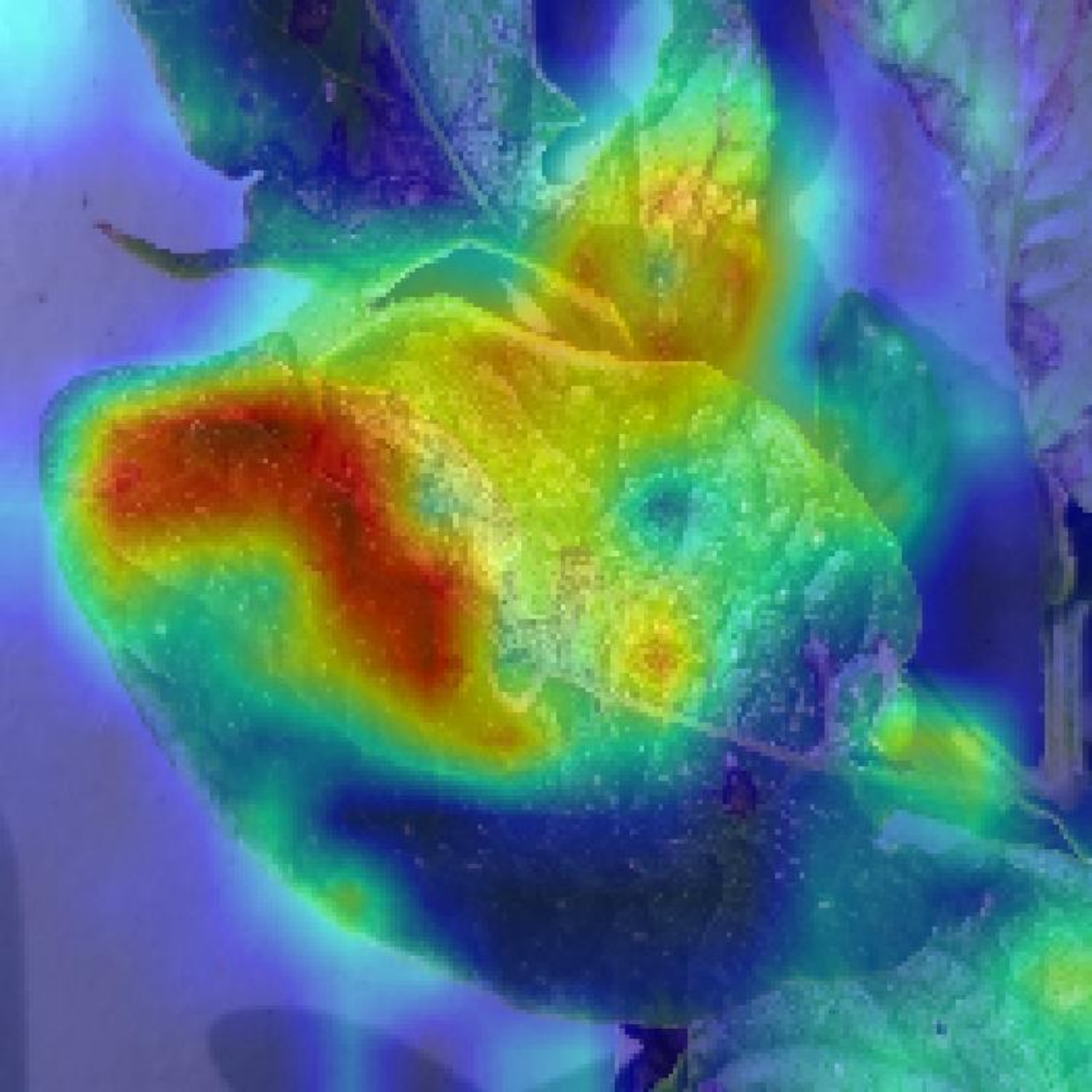} \\[-2pt]
\small Original (Fungi) & \small MobileNetV2 & \small Teacher & \small Student \\[8pt]
\includegraphics[width=0.23\textwidth]{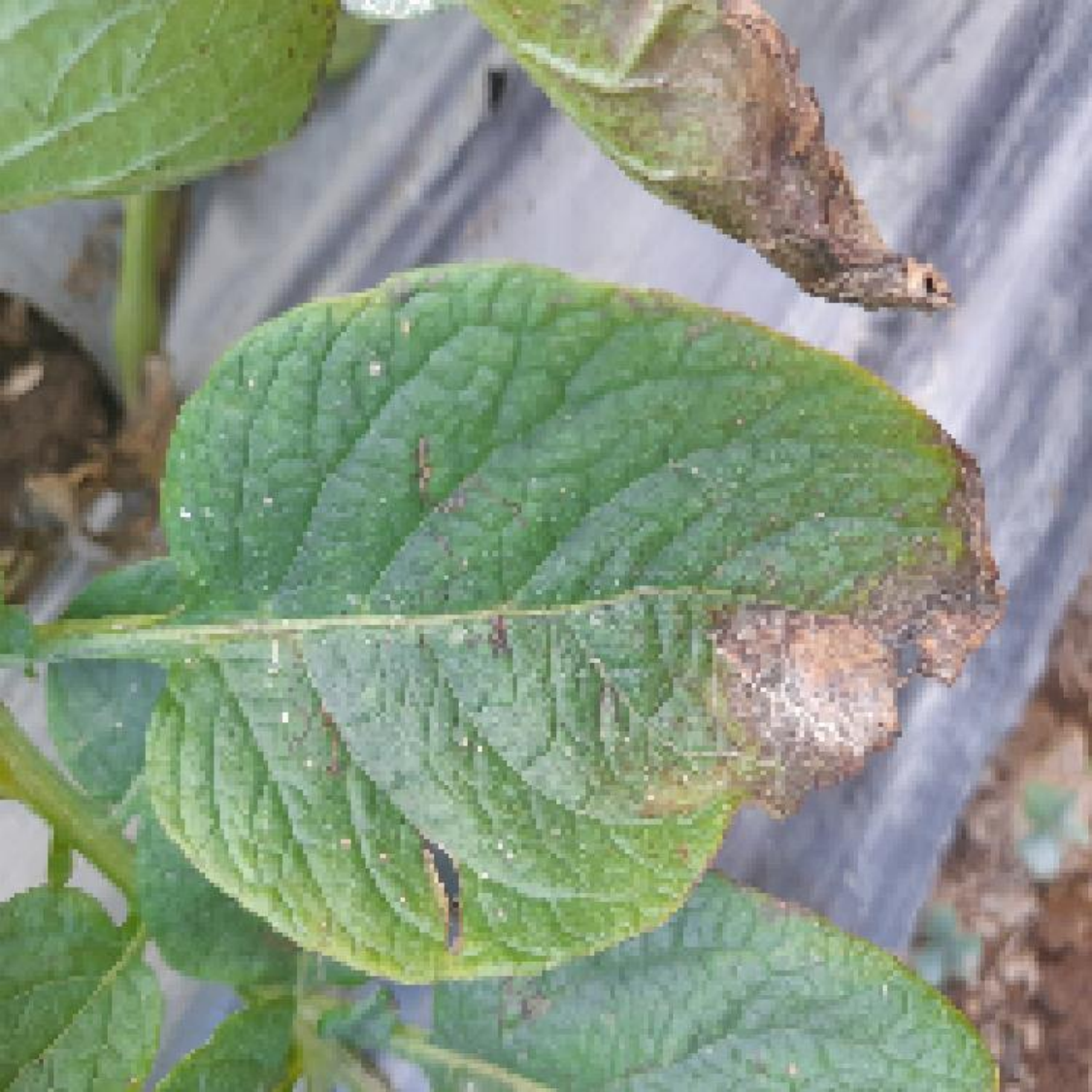} &
\includegraphics[width=0.23\textwidth]{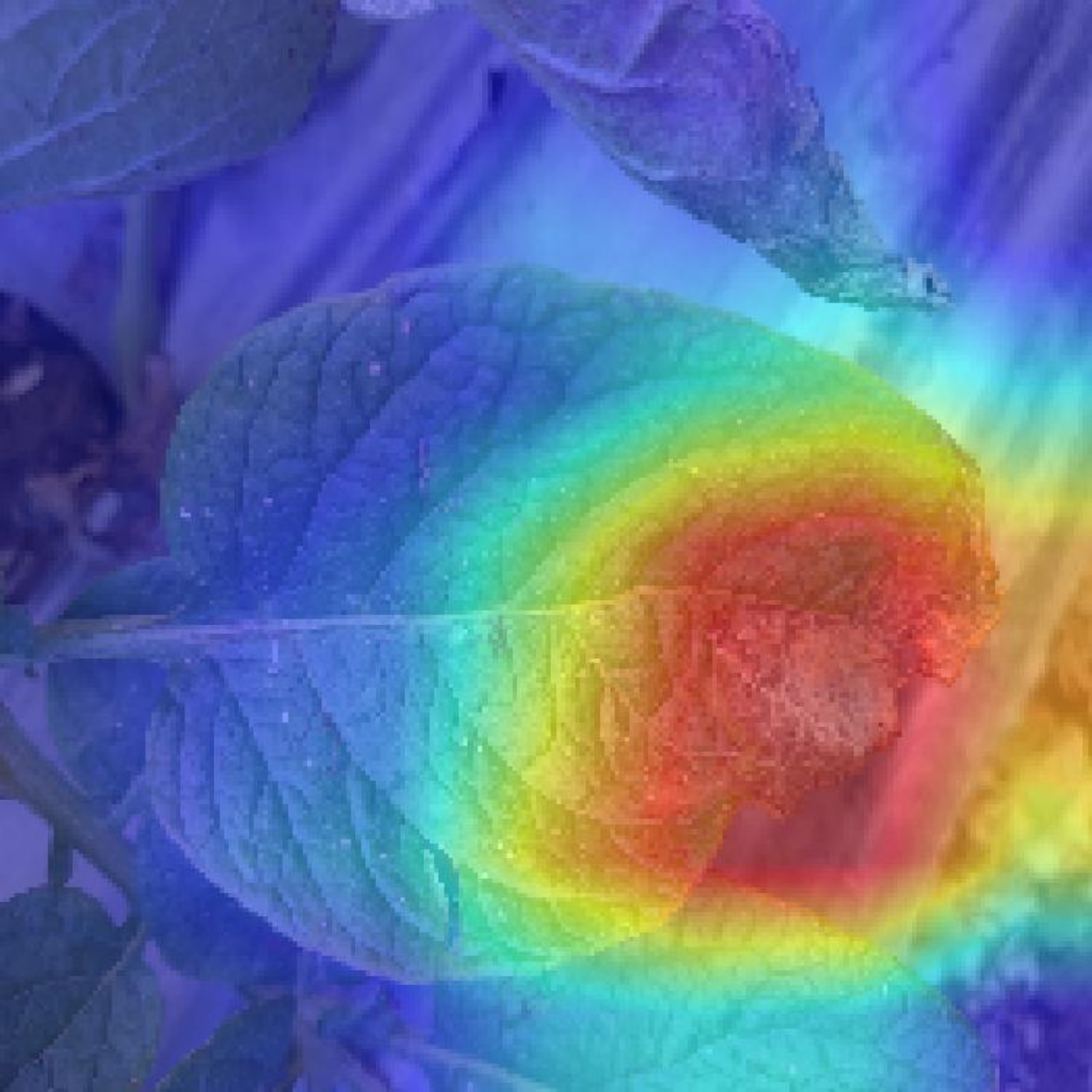} &
\includegraphics[width=0.23\textwidth]{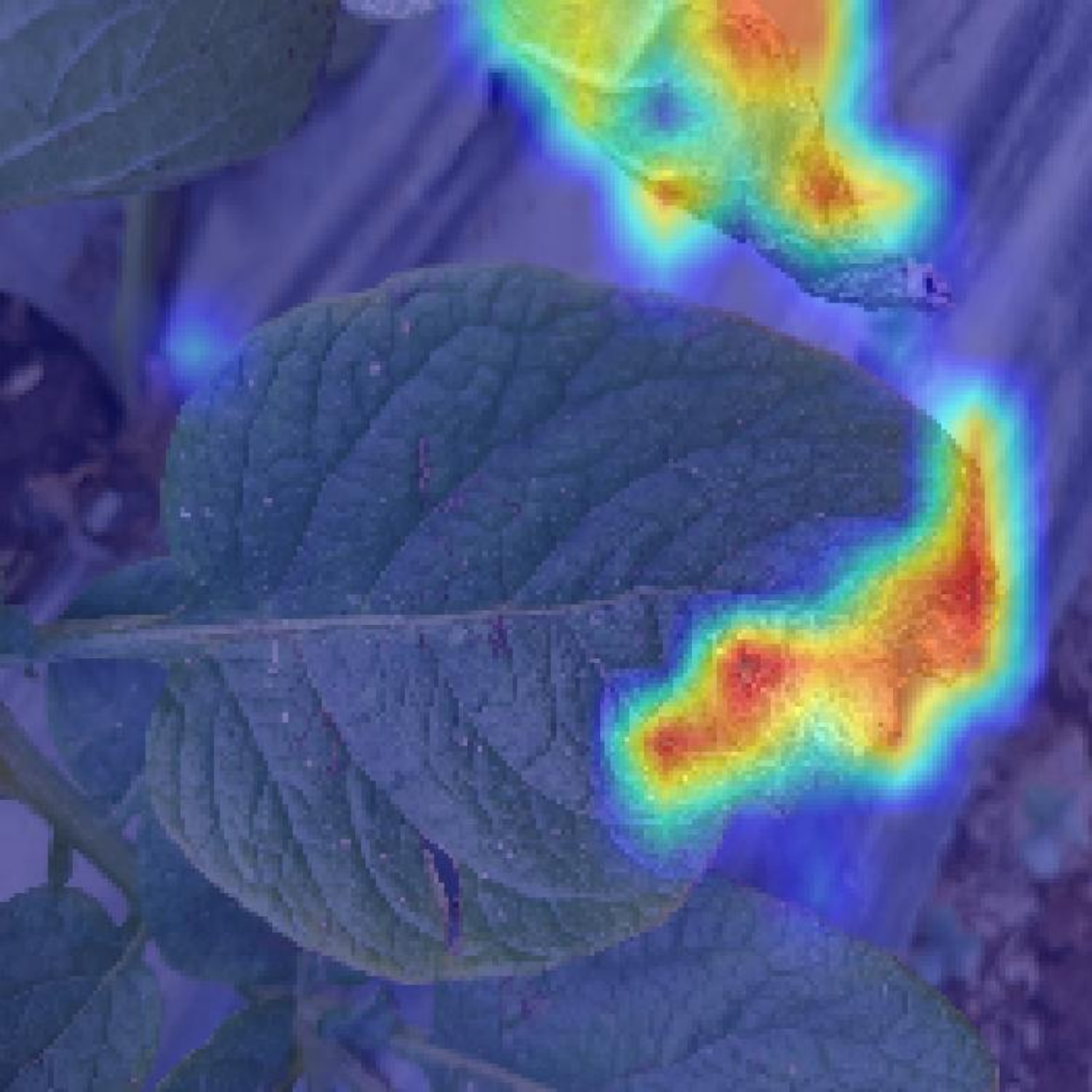} &
\includegraphics[width=0.23\textwidth]{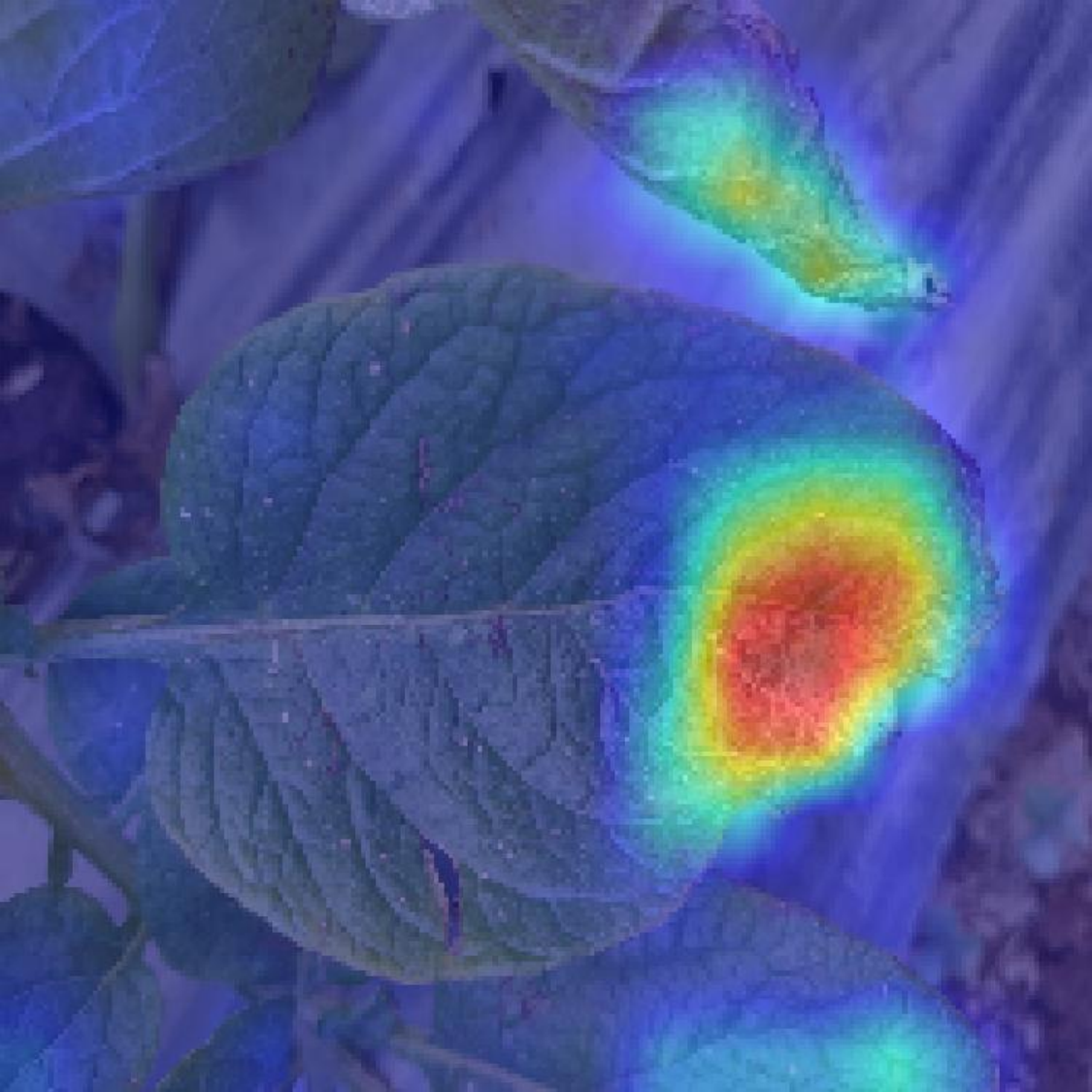} \\[-2pt]
\small Original (Phytophthora) & \small MobileNetV2 & \small Teacher & \small Student
\end{tabular}
\caption{Grad-CAM on Potato dataset}
\label{fig:gradcam_potato}
\end{figure*}

Figure~\ref{fig:gradcam_potato} presents Grad-CAM visualisations on the more challenging Potato dataset, which exhibits stronger class imbalance and higher visual variability.

On the Fungi case, the baseline CNN produces coarse and over-smoothed attention, failing to capture fine-grained lesion patterns, while the ViT teacher shows highly fragmented and noisy activations.
The distilled student, in contrast, produces more coherent and structured attention, capturing relevant regions with reduced noise.

On the Phytophthora case, the baseline model focuses on a dominant region but deviates from the true lesion area, and the teacher attends to elongated but misaligned regions.
The distilled student achieves more accurate localisation, concentrating on the actual infected region with clearer and more compact activations.

These results highlight that distillation improves robustness under challenging conditions.
Compared to the CNN baseline, the student captures more detailed and complete disease patterns, and compared to the ViT teacher, it suppresses spurious activations and improves localisation accuracy.
This suggests that the student effectively integrates global contextual knowledge with spatial selectivity, leading to more reliable representations in imbalanced and complex datasets.

The Grad-CAM results consistently show that the distilled student produces more focused and coherent attention than both the baseline CNN and the ViT teacher.
This indicates that multi-level distillation improves not only predictive performance but also the quality and reliability of learned representations.

\FloatBarrier

\section{Comparison with previous studies}
\label{sec:Comparative}



\begin{figure}[H]
\centering
\includegraphics[width=\linewidth]{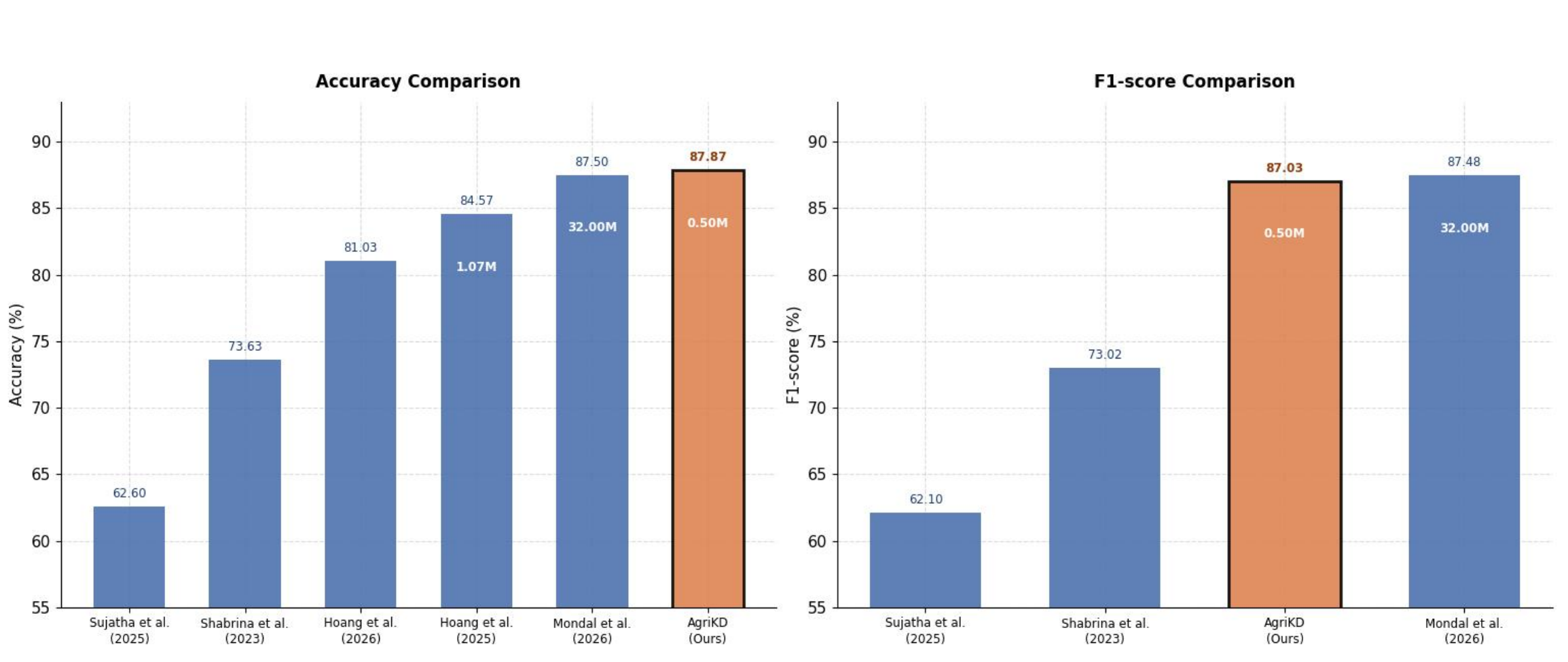}
\caption{Performance comparison of existing approaches on the Potato Leaf Disease dataset}
\label{fig:comparative_study_potato}
\end{figure}

Figure~\ref{fig:comparative_study_potato} presents a comparison between AgriKD and existing approaches on the Potato Leaf Disease dataset in terms of accuracy and F1-score. AgriKD achieves the highest accuracy of 87.87\%, while maintaining a competitive F1-score of 87.03\%.

Although~\citep{mondal2026hybrid} achieved a slightly higher F1-score (87.48\%), their model was substantially larger (32M vs. 0.50M parameters), indicating a clear efficiency advantage for AgriKD. In addition, compared to ~\citep{hoang2025multi}, AgriKD improved accuracy from 84.57\% to 87.87\% while using fewer parameters, further reinforcing its effectiveness–efficiency trade-off.
The handcrafted-feature-based LightGBM model of ~\citep{10.1007/978-981-95-4960-3_20} achieves 81.03\% accuracy, outperforming earlier methods such as \citep{sujatha2025advancing} and \citep{Shabrina2023Potato} (below 75\%), but still trailing behind deep learning approaches. This pattern suggests that while feature engineering improves representation quality over traditional methods, recent gains are primarily driven by learned representations.

The results indicate that performance improvements are not solely associated with increasing model size. Instead, AgriKD achieves comparable or better results than larger models, suggesting a more effective use of model capacity through knowledge distillation rather than scale.

\section{Deployment Validation}
\label{sec:deployment}
\subsection{Real-World Deployment Demonstration}
\label{subsec:deployment-demo}

To demonstrate the practical applicability of the proposed framework beyond benchmark evaluation, the distilled student model was deployed in real-world inference scenarios on both edge devices and mobile platforms.
In both settings, the inference pipeline follows a consistent procedure: the exported model (TFLite Float16 for mobile and TensorRT FP16 for edge devices) processes input images resized to $224 \times 224$ and normalized identically to the test configuration, before producing the predicted disease class and confidence score.

\begin{figure}[H]
\centering
\includegraphics[width=1\linewidth]{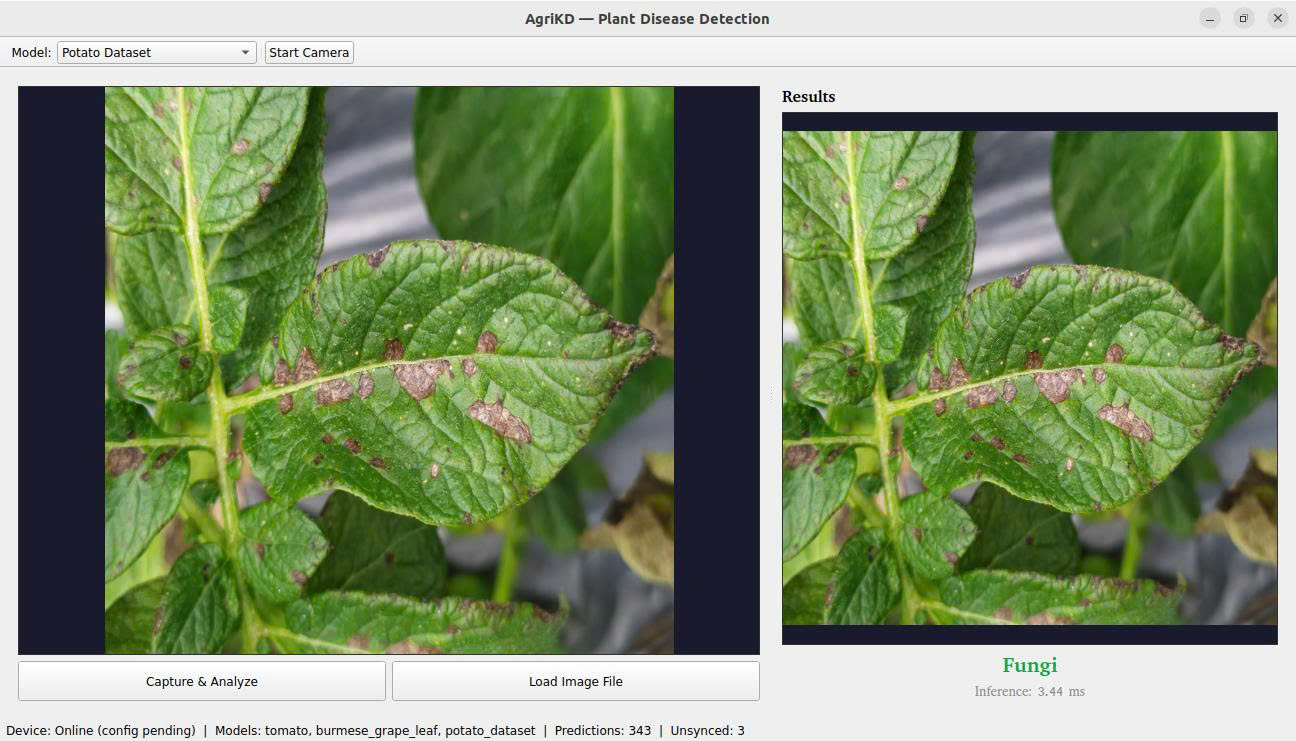}
\caption{Inference on NVIDIA Jetson Orin Nano using TensorRT FP16: Potato datasets.}
\label{fig:jetson-demo}
\end{figure}

\textbf{Edge device deployment.}
The TensorRT FP16 model was deployed on an NVIDIA Jetson Orin Nano device. Figure~\ref{fig:jetson-demo} shows representative inference results, confirming real-time inference with ultra-low latency (1.9--2.2 ms), consistent with the quantitative results reported in Table~\ref{tab:format-all}.

\begin{figure}[H]
\centering
\includegraphics[width=0.32\linewidth]{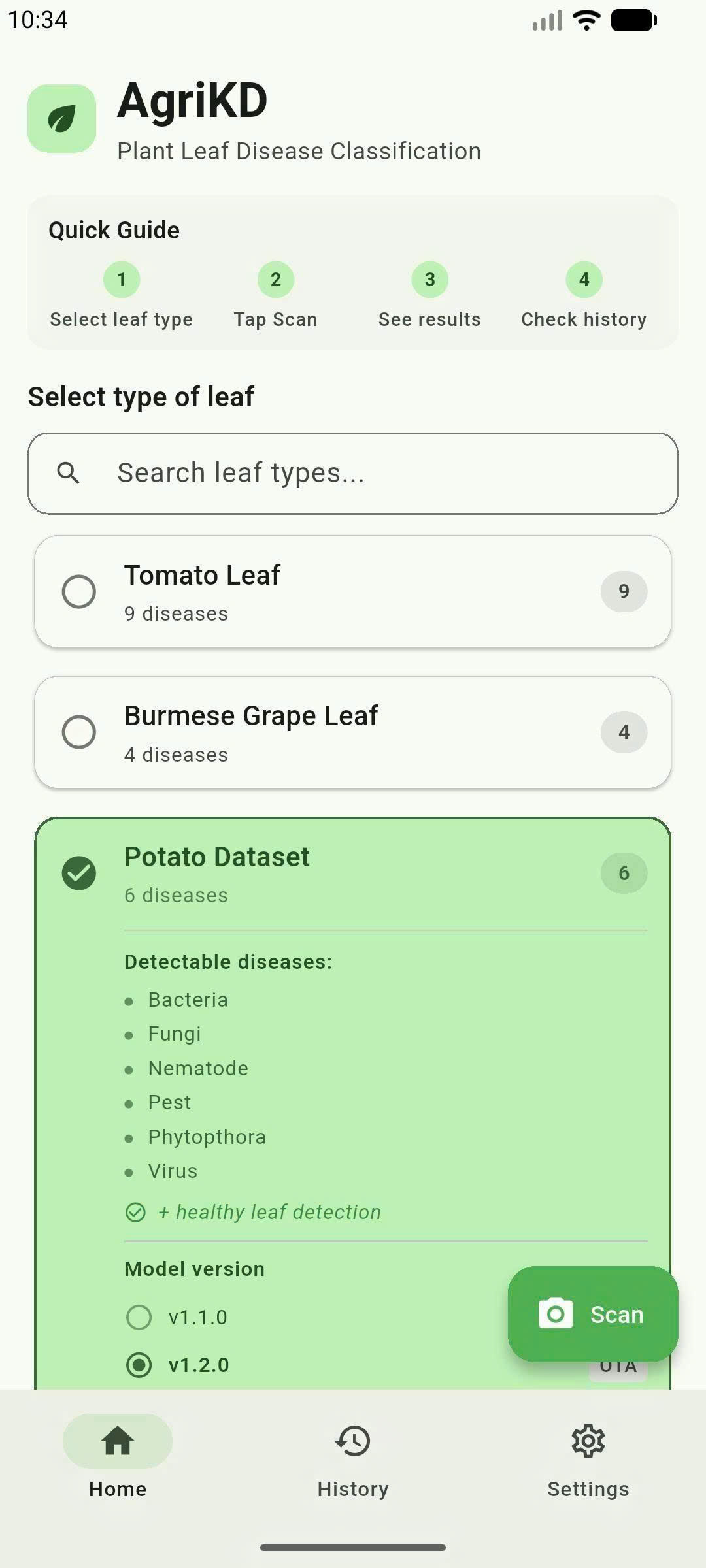}
\includegraphics[width=0.32\linewidth]{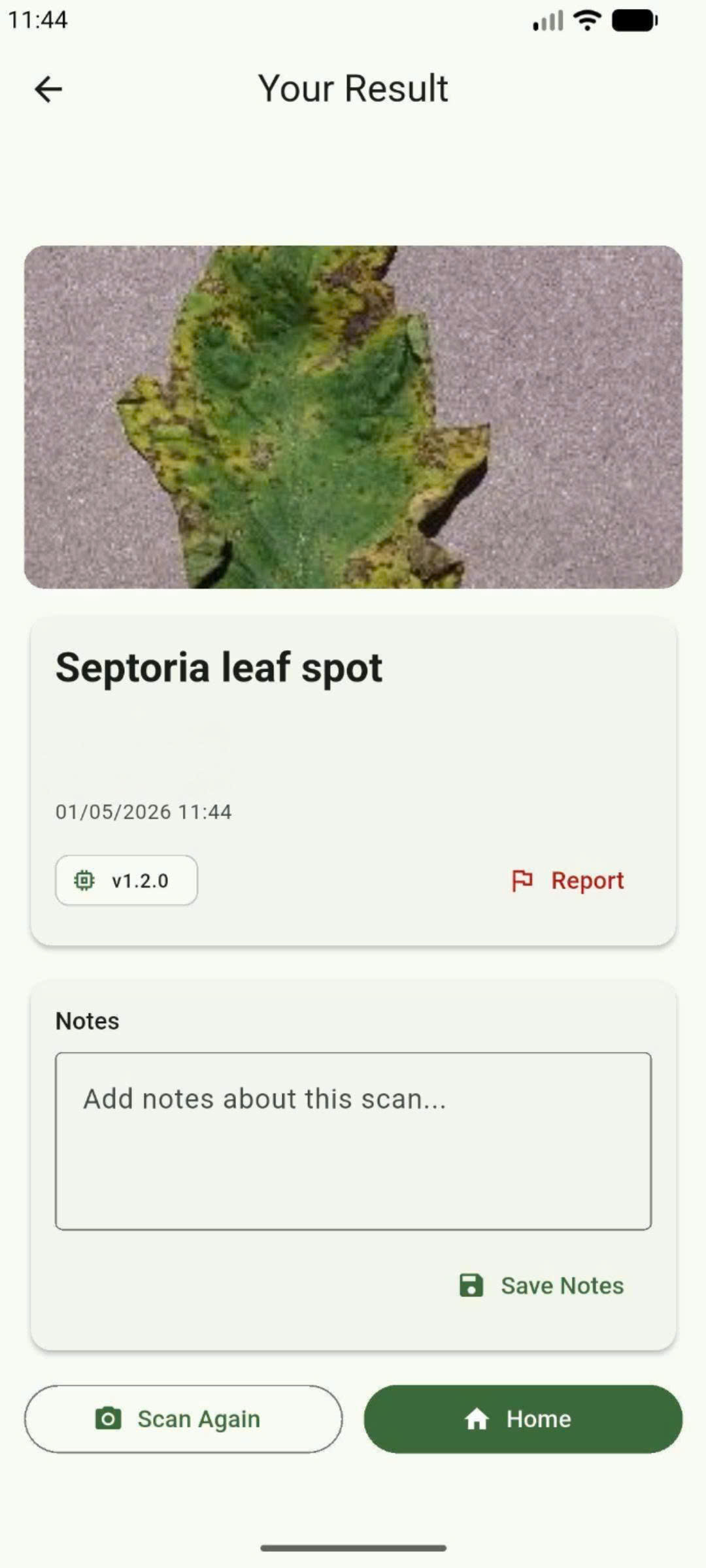}
\includegraphics[width=0.32\linewidth]{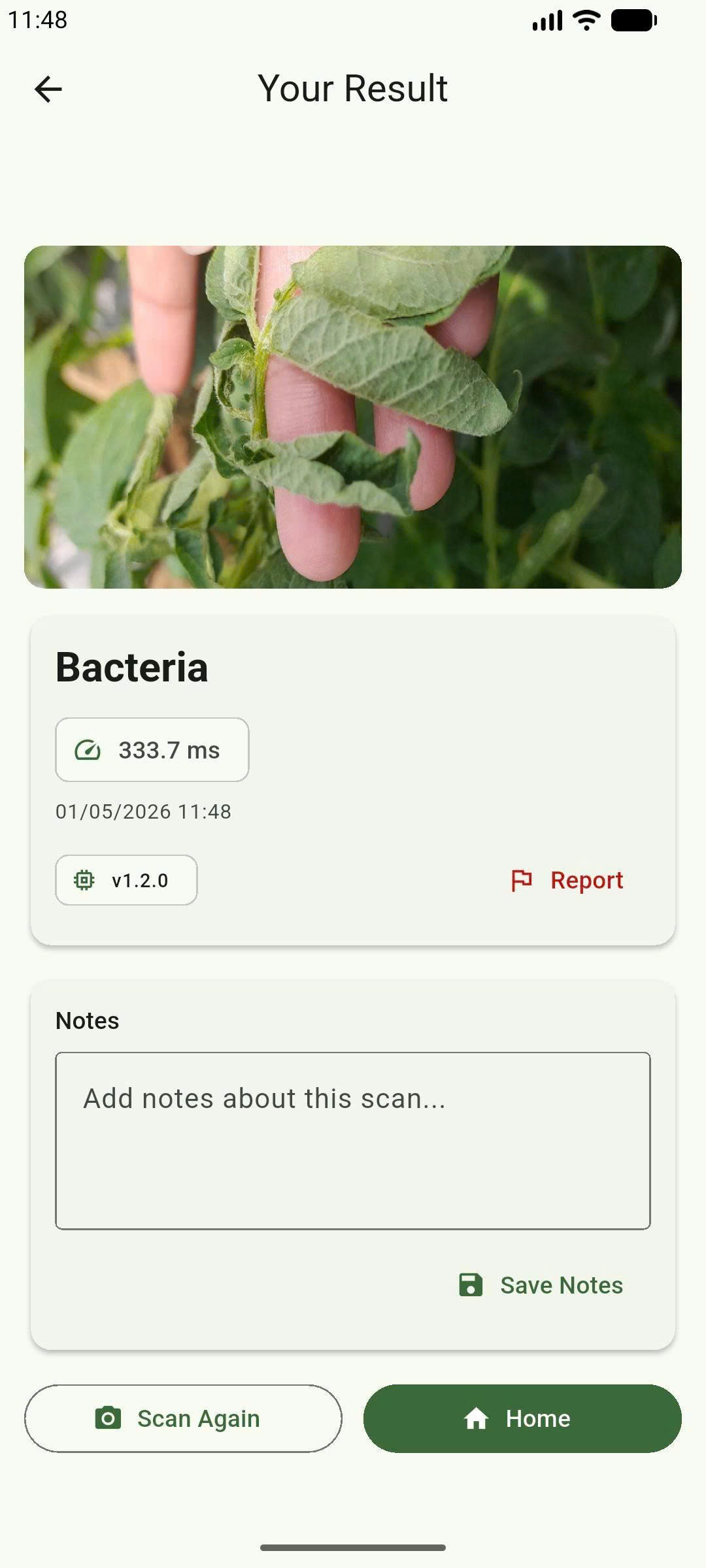}
\caption{Mobile application using TFLite Float16: application menu (left),
Tomato inference (center), and Potato inference (right).}
\label{fig:mobile-demo}
\end{figure}

\textbf{Mobile application deployment.}
The TFLite Float16 model was integrated into a mobile application. Figure~\ref{fig:mobile-demo} illustrates the application interface, including the crop selection menu (left) and representative inference results on Tomato and Potato datasets (center and right). The model runs entirely on-device without server-side computation, enabling low-latency inference while preserving user data privacy, and maintaining a compact sub-1 MB model footprint. This design is particularly suitable for real-world agricultural applications in remote or connectivity-limited environments.

\subsection{Cross-Format Deployment Performance}
\label{subsec:format-validation}

To validate deployment readiness, the final distilled student models were exported from PyTorch to ONNX, TFLite Float16, and TensorRT FP16 formats. Despite potential numerical differences introduced by graph optimization and reduced-precision computation, the results showed that classification fidelity was consistently preserved across all formats (see Tables~\ref{tab:format-all})

Across the three datasets, performance remained stable after conversion. On Tomato (Table~\ref{tab:format-all}), all formats achieved identical accuracy and F1-score. On Burmese Grape (Table~\ref{tab:format-all}), only a minor decrease (0.16\% in accuracy and 0.20\% in F1-score) was observed. On Potato (Table~\ref{tab:format-all}), the differences were negligible, with at most 0.01\% variation in accuracy. These results indicate that format conversion introduces minimal impact on predictive performance.

In contrast, deployment efficiency improved substantially. As shown in Tables~\ref{tab:format-all}, ONNX and TFLite reduced CPU latency by more than 2× compared to PyTorch, while TFLite Float16 compressed the model size to approximately 0.95--0.96 MB. TensorRT FP16 achieved the best performance, reaching 1.9--2.2 ms latency and up to 523 FPS on the Jetson edge device.

These findings suggest that the distilled student model can be reliably deployed across multiple formats with negligible accuracy loss, while enabling real-time inference and significant reductions in computational and storage costs.

\begin{center}
\captionof{table}{Cross-format deployment evaluation across datasets}
\label{tab:format-all}
\resizebox{\columnwidth}{!}{%
\begin{tabular}{l l c c c c c}
\hline\hline
\textbf{Dataset} & \textbf{Format} & \textbf{Acc (\%)} & \textbf{F1 (\%)} &
\textbf{Latency (ms)} & \textbf{FPS} & \textbf{Size (MB)} \\
\hline

\multirow{4}{*}{Tomato}
& PyTorch        & 88.16 & 90.13 & 11.4 &  87 & 1.99 \\
& ONNX           & 88.16 & 90.13 &  5.0 & 199 & 1.87 \\
& TFLite FP16    & 88.16 & 90.13 &  4.9 & 204 & 0.96 \\
& TensorRT FP16$^{\dagger}$ & 88.16 & 90.13 & 2.2 & 461 & 1.23 \\
\hline

\multirow{4}{*}{Burmese Grape}
& PyTorch        & 86.57 & 86.14 & 12.3 &  82 & 1.98 \\
& ONNX           & 86.57 & 86.14 &  3.9 & 256 & 1.87 \\
& TFLite FP16    & 86.41 & 85.94 &  4.8 & 207 & 0.95 \\
& TensorRT FP16$^{\dagger}$ & 86.41 & 85.94 & 1.9 & 523 & 1.22 \\
\hline

\multirow{4}{*}{Potato}
& PyTorch        & 87.81 & 85.86 & 14.4 &  70 & 1.98 \\
& ONNX           & 87.81 & 85.86 &  3.4 & 292 & 1.87 \\
& TFLite FP16    & 87.81 & 85.86 &  5.1 & 198 & 0.96 \\
& TensorRT FP16$^{\dagger}$ & 87.80 & 85.86 & 2.0 & 498 & 1.25 \\
\hline\hline
\end{tabular}}
\vspace{2pt}
\footnotesize{$^{\dagger}$TensorRT FP16 is evaluated on the Jetson edge device with GPU acceleration; all other formats are evaluated on a CPU-only GitHub Actions runner.}
\end{center}

\section{Discussion}
\label{sec:discussion}

\subsection{Dataset Difficulty}
The three datasets span a spectrum of acquisition difficulty that directly influences model behavior. Burmese Grape is collected under controlled conditions with foreground extraction, resulting in stable performance despite moderate imbalance (3.4:1). Tomato represents a semi-controlled setting with a higher number of classes and variable acquisition conditions, leading to increased cross-fold variation and moderate imbalance (4.04:1). In contrast, Potato is the most challenging dataset, with open-field images, no foreground extraction, and severe imbalance (11:1), making it difficult for both generalization and spatial feature learning.

\subsection{Effect of Knowledge Distillation Across Datasets}
Across all datasets, the distilled student achieves macro F1-scores that are competitive with or higher than the teacher while substantially reducing model complexity, indicating an improved performance--efficiency trade-off. The student outperforms the teacher on Burmese Grape and Tomato (+1.30\% and +1.25\% macro F1), but remains slightly below on Potato ($-1.03\%$).

This behavior aligns with the regularizing effect of knowledge distillation. Under controlled or moderately complex conditions, the teacher’s softened outputs provide richer supervision and help reduce overfitting. In contrast, the open-field conditions of Potato introduce background variability that weakens spatial alignment between student features and teacher representations, reducing the contribution of projection-based losses. Nevertheless, Grad-CAM results (Section~\ref{subsec:gradcam}) show that the distilled student still produces more coherent lesion localization than the baseline MobileNetV2.

\subsection{Generalization Insight}
Three consistent patterns emerge. First, logit-based and relation-based distillation provide the primary performance gains across datasets, indicating that prediction-level supervision is the most robust component. Second, projection-based losses offer smaller but consistent improvements, acting as a refinement objective rather than a primary driver of performance. Third, the benefit of logit distillation increases with the number of classes, as richer inter-class relationships provide more informative supervision (e.g., +6.97\% on Tomato vs. +2.89\% on Potato).

\subsection{Limitations}
Several limitations should be noted. The evaluation is limited to three datasets and may not fully reflect real-world variability. The framework is tested with a single teacher--student architecture pair, and comparisons with alternative distillation methods are not included. In addition, the choice of truncation depth is not systematically validated, and energy consumption on edge devices is not evaluated, limiting the completeness of the deployment analysis.

\section{Conclusion}
\label{sec:conclusion}
This paper presents AgriKD, a deployment-oriented cross-architecture knowledge distillation framework that transfers knowledge from a ViT-Base teacher to a compact MobileNetV2 student for efficient leaf disease classification. By systematically integrating multiple complementary distillation objectives, the proposed approach enables the lightweight student to achieve strong classification performance while significantly reducing model size, computational cost, and inference latency.
Experiments on three datasets demonstrate that the distilled student
achieves a favorable performance--efficiency trade-off. The full
distillation objective consistently outperforms cross-entropy-only
training (e.g., +9.12\% F1 on Tomato and +4.54\% on Potato), while
reducing model complexity by over two orders of magnitude.
Deployment validation further confirms that the model can be reliably
converted to ONNX, TFLite Float16, and TensorRT FP16 with negligible
accuracy loss. In particular, TFLite reduces model size to below 1 MB,
and TensorRT enables real-time inference (1.9--2.2 ms latency) on an
edge device, demonstrating practical deployment readiness.
Future work will explore leveraging the ViT \texttt{[CLS]} token as an
additional global semantic representation, evaluating on more challenging real-world
datasets, and extending the framework to diverse teacher--student
architectures. Overall, AgriKD provides an effective step toward
compact and deployable AI solutions for crop disease diagnosis.

\section*{Funding:} This research received no funding.

\section*{Data and Code Availability}

The datasets used in this study are publicly available from the following sources:

\begin{itemize}
\item Burmese dataset: \href{https://data.mendeley.com/datasets/k6gy38xv89/1}{Burmese datase}
\item Cropped Burmese dataset: \href{https://www.kaggle.com/datasets/leminhdungk18dn/burmese-crop}{Cropped Burmese}
\item Potato dataset: \href{https://data.mendeley.com/datasets/ptz377bwb8/1}{Potato }
\item Tomato dataset: \href{https://data.mendeley.com/datasets/zfv4jj7855/1}{Tomato}
\end{itemize}

\begin{samepage}
The research code and software implementation are available at:

\begin{itemize}
\item Research code: \href{https://github.com/BerserkerFPT/Capstone_KD/tree/main}{GitHub repository}
\item Software/mobile application: \href{https://github.com/zoe141004/agrikd-app}{Mobile app repository}
\end{itemize}
\end{samepage}

\section*{Author contribution:}

\textbf{Minh-Dung Le}: Writing - review \& editing, Writing - original draft, Data curation,S Visualization, Validation, Methodology, Investigation. \textbf{Hoang-Vu Truong}: Software, Data curation, Investigation, Validation, Methodology, Software. \textbf{Minh-Duc Hoang}: Formal analysis, Validation, Investigation, Visualization, Methodology, Software.
\textbf{Thi-Thu-Hong Phan}: Conceptualization, Methodology, Supervision, Validation, Writing - original draft, Writing - review \& editing.

\section*{Declaration of interests:}

The authors declare that they have no known competing financial interests or personal relationships that could have appeared to influence the work reported in this paper.

\section*{Declaration of generative AI and AI-Assisted technologies in the writing process:}

During the preparation of this manuscript, Grammarly and generative large language models were employed to improve grammar and refine wording for clarity. The authors subsequently reviewed and revised the content as necessary and take full responsibility for the final version of the manuscript.

\bibliographystyle{elsarticle-harv}
\bibliography{agrikd-refs}

@inproceedings{sandler2018mobilenetv2,
  title     = {MobileNetV2: Inverted Residuals and Linear Bottlenecks},
  author    = {Sandler, Mark and Howard, Andrew and Zhu, Menglong and Zhmoginov, Andrey and Chen, Liang-Chieh},
  booktitle = {Proceedings of the IEEE Conference on Computer Vision and Pattern Recognition (CVPR)},
  pages     = {4510--4520},
  year      = {2018},
  note      = {PLACEHOLDER - verify DOI}
}

@inproceedings{lin2017focal,
  title     = {Focal Loss for Dense Object Detection},
  author    = {Lin, Tsung-Yi and Goyal, Priya and Girshick, Ross and He, Kaiming and Doll{\'a}r, Piotr},
  booktitle = {Proceedings of the IEEE International Conference on Computer Vision (ICCV)},
  pages     = {2980--2988},
  year      = {2017},
  note      = {PLACEHOLDER - verify DOI}
}

@article{abade2021plant,
  title={Plant diseases recognition on images using convolutional neural networks: A systematic review},
  author={Abade, Andr{\'e} and Ferreira, Paulo Afonso and de Barros Vidal, Flavio},
  journal={Computers and Electronics in Agriculture},
  volume={185},
  pages={106125},
  year={2021},
  publisher={Elsevier}
}

@article{dhaka2021survey,
  title={A survey of deep convolutional neural networks applied for prediction of plant leaf diseases},
  author={Dhaka, Vijaypal Singh and Meena, Sangeeta Vaibhav and Rani, Geeta and Sinwar, Deepak and Ijaz, Muhammad Fazal and Wo{\'z}niak, Marcin},
  journal={Sensors},
  volume={21},
  number={14},
  pages={4749},
  year={2021},
  publisher={MDPI}
}

@article{hinton2015distilling,
  title={Distilling the knowledge in a neural network},
  author={Hinton, Geoffrey and Vinyals, Oriol and Dean, Jeff},
  journal={arXiv preprint arXiv:1503.02531},
  year={2015}
}

@article{dosovitskiy2020image,
  title={An image is worth 16x16 words: Transformers for image recognition at scale},
  author={Dosovitskiy, Alexey and Beyer, Lucas and Kolesnikov, Alexander and Weissenborn, Dirk and Zhai, Xiaohua and Unterthiner, Thomas and Dehghani, Mostafa and Minderer, Matthias and Heigold, Georg and Gelly, Sylvain and others},
  journal={arXiv preprint arXiv:2010.11929},
  year={2020}
}

@article{anis2026survey,
  title={A survey of deep learning techniques for image-based disease detection in dicot plants},
  author={Anis, Adeeba and Wang, Penghao and Li, Chengdao and Sohel, Ferdous},
  journal={Information Processing in Agriculture},
  year={2026},
  publisher={Elsevier}
}

@InProceedings{Ren_2022_CVPR,
    author    = {Ren, Sucheng and Gao, Zhengqi and Hua, Tianyu and Xue, Zihui and Tian, Yonglong and He, Shengfeng and Zhao, Hang},
    title     = {Co-Advise: Cross Inductive Bias Distillation},
    booktitle = {Proceedings of the IEEE/CVF Conference on Computer Vision and Pattern Recognition (CVPR)},
    month     = {June},
    year      = {2022},
    pages     = {16773-16782}
}

@misc{mugisha2025hybridknowledgetransferattention,
      title={Hybrid Knowledge Transfer through Attention and Logit Distillation for On-Device Vision Systems in Agricultural IoT}, 
      author={Stanley Mugisha and Rashid Kisitu and Florence Tushabe},
      year={2025},
      eprint={2504.16128},
      archivePrefix={arXiv},
      primaryClass={cs.CV},
      url={https://arxiv.org/abs/2504.16128}, 
}

@misc{phan2025multi,
      title={Multi-objective hybrid knowledge distillation for efficient deep learning in smart agriculture}, 
      author={Phi-Hung Hoang and Nam-Thuan Trinh and Van-Manh Tran and Thi-Thu-Hong Phan},
      year={2025},
      eprint={2512.22239},
      archivePrefix={arXiv},
      primaryClass={cs.CV},
      url={https://arxiv.org/abs/2512.22239}, 
}

@article{mehdipour2026novel,
  title={A novel lightweight hybrid CNN--ViT for maize leaf disease classification},
  author={Mehdipour, Saber and Mirroshandel, Seyed Abolghasem and Tabatabaei, Seyed Amirhossein},
  journal={Scientific Reports},
  year={2026},
  publisher={Nature Publishing Group UK London}
}

@misc{Liu2022CAKD,
      title={Cross-Architecture Knowledge Distillation}, 
      author={Yufan Liu and Jiajiong Cao and Bing Li and Weiming Hu and Jingting Ding and Liang Li},
      year={2022},
      eprint={2207.05273},
      archivePrefix={arXiv},
      primaryClass={cs.CV},
      url={https://arxiv.org/abs/2207.05273}, 
}

@misc{Huang2022DIST,
      title={Knowledge Distillation from A Stronger Teacher}, 
      author={Tao Huang and Shan You and Fei Wang and Chen Qian and Chang Xu},
      year={2022},
      eprint={2205.10536},
      archivePrefix={arXiv},
      primaryClass={cs.CV},
      url={https://arxiv.org/abs/2205.10536}, 
}

@article{jlassi2024potato,
  title={Potato leaf disease classification using transfer learning and reweighting-based training with imbalanced data},
  author={Jlassi, Amal and Elaoud, Amani and Ghazouani, Haythem and Barhoumi, Walid},
  journal={SN Computer Science},
  volume={5},
  number={8},
  pages={987},
  year={2024},
  publisher={Springer}
}

@article{lu2023improved,
  title={Improved MobileNetV2 crop disease identification model for intelligent agriculture},
  author={Lu, Jianbo and Liu, Xiaobin and Ma, Xiaoya and Tong, Jin and Peng, Jungui},
  journal={PeerJ Computer Science},
  volume={9},
  pages={e1595},
  year={2023},
  publisher={PeerJ Inc.}
}

@article{borhani2022deep,
  title={A deep learning based approach for automated plant disease classification using vision transformer},
  author={Borhani, Yasamin and Khoramdel, Javad and Najafi, Esmaeil},
  journal={Scientific Reports},
  volume={12},
  number={1},
  pages={11554},
  year={2022},
  publisher={Nature Publishing Group UK London}
}

@article{thai2023formerleaf,
  title={FormerLeaf: An efficient vision transformer for Cassava Leaf Disease detection},
  author={Thai, Huy-Tan and Le, Kim-Hung and Nguyen, Ngan Luu-Thuy},
  journal={Computers and Electronics in Agriculture},
  volume={204},
  pages={107518},
  year={2023},
  publisher={Elsevier}
}

@article{Solapure2024Tomato,
  author = {Vaibhav Solapure and SmartAgroTech DY and Anish Jawale},
  title = {Tomato Leaf Disease Dataset},
  year = {2024},
  doi = {10.17632/zfv4jj7855.1},
  journal = {Mendeley Data},
  volume = {1}
}

@article{Rahman2025Burmese,
  author = {Salman Af Rahman and Md. Nafiz Imtiaz and Naima Ahmed and Md Hasan Imam Bijoy},
  title = {Burmese Grape Leaf Disease Dataset for Computer Vision-Based Plant Health Diagnosis},
  year = {2025},
  doi = {10.17632/k6gy38xv89.1},
  journal = {Mendeley Data},
  volume = {1}
}

@article{Shabrina2023Potato,
  author = {Nabila Husna Shabrina and Siwi Indarti and Rina Maharani and Dinar Ajeng Kristiyanti and Irmawati Irmawati and Niki Prastomo and Tika Adillah M},
  title = {A novel dataset of potato leaf disease in uncontrolled environment},
  journal = {Data in Brief},
  volume = {52},
  pages = {109955},
  year = {2024},
  doi = {10.1016/j.dib.2023.109955}
}

@article{hossain2024deep,
  title={Deep learning for mango leaf disease identification: A vision transformer perspective},
  author={Hossain, Md Arban and Sakib, Saadman and Abdullah, Hasan Muhammad and Arman, Shifat E},
  journal={Heliyon},
  volume={10},
  number={17},
  year={2024},
  publisher={Elsevier}
}

@article{duhan2025rtr_lite_mobilenetv2,
  title={RTR\_Lite\_MobileNetV2: A lightweight and efficient model for plant disease detection and classification},
  author={Duhan, Sangeeta and Gulia, Preeti and Gill, Nasib Singh and Narwal, Ekta},
  journal={Current Plant Biology},
  volume={42},
  pages={100459},
  year={2025},
  publisher={Elsevier}
}

@article{noman2025vix,
  title={ViX-MangoEFormer: An Enhanced Vision Transformer--EfficientFormer and Stacking Ensemble Approach for Mango Leaf Disease Recognition with Explainable Artificial Intelligence},
  author={Noman, Abdullah Al and Hossain, Amira and Sakib, Anamul and Debnath, Jesika and Fardin, Hasib and Sakib, Abdullah Al and Haque, Rezaul and Ahmed, Md Redwan and Reza, Ahmed Wasif and Dewan, M Ali Akber},
  journal={Computers},
  volume={14},
  number={5},
  pages={171},
  year={2025},
  publisher={MDPI}
}

@article{hao2023one,
  title={One-for-all: Bridge the gap between heterogeneous architectures in knowledge distillation},
  author={Hao, Zhiwei and Guo, Jianyuan and Han, Kai and Tang, Yehui and Hu, Han and Wang, Yunhe and Xu, Chang},
  journal={Advances in Neural Information Processing Systems},
  volume={36},
  pages={79570--79582},
  year={2023}
}

@inproceedings{huang2017densely,
  title={Densely connected convolutional networks},
  author={Huang, Gao and Liu, Zhuang and Van Der Maaten, Laurens and Weinberger, Kilian Q},
  booktitle={Proceedings of the IEEE conference on computer vision and pattern recognition},
  pages={4700--4708},
  year={2017}
}

@inproceedings{tan2019efficientnet,
  title={Efficientnet: Rethinking model scaling for convolutional neural networks},
  author={Tan, Mingxing and Le, Quoc},
  booktitle={International conference on machine learning},
  pages={6105--6114},
  year={2019},
  organization={PMLR}
}

@article{simonyan2014very,
  title={Very deep convolutional networks for large-scale image recognition},
  author={Simonyan, Karen and Zisserman, Andrew},
  journal={arXiv preprint arXiv:1409.1556},
  year={2014}
}

@inproceedings{he2016deep,
  title={Deep residual learning for image recognition},
  author={He, Kaiming and Zhang, Xiangyu and Ren, Shaoqing and Sun, Jian},
  booktitle={Proceedings of the IEEE conference on computer vision and pattern recognition},
  pages={770--778},
  year={2016}
}

@article{sujatha2025advancing,
  title={Advancing plant leaf disease detection integrating machine learning and deep learning},
  author={Sujatha, R and Krishnan, Sushil and Chatterjee, Jyotir Moy and Gandomi, Amir H},
  journal={Scientific Reports},
  volume={15},
  number={1},
  pages={11552},
  year={2025},
  publisher={Nature Publishing Group UK London}
}

@article{hoang2025multi,
  title={Multi-objective hybrid knowledge distillation for efficient deep learning in smart agriculture},
  author={Hoang, Phi-Hung and Trinh, Nam-Thuan and Tran, Van-Manh and Phan, Thi-Thu-Hong},
  journal={arXiv preprint arXiv:2512.22239},
  year={2025}
}

@InProceedings{10.1007/978-981-95-4960-3_20,
author="Hoang, Phi-Hung
and Phan, Thi-Thu-Hong",
editor="Quan, Thanh Tho
and Sombattheera, Chattrakul
and Pham, Hoang-Anh
and Tran, Ngoc Thinh",
title="Potato Leaf Disease Classification in Uncontrolled Environments: Leveraging the Synergy of Handcrafted Features",
booktitle="Multi-disciplinary Trends in Artificial Intelligence",
year="2026",
publisher="Springer Nature Singapore",
address="Singapore",
pages="247--259",
isbn="978-981-95-4960-3"
}

@article{mondal2026hybrid,
  title={A hybrid CNN-transformer model with adaptive activation function for potato leaf disease classification},
  author={Mondal, Ayan and Chatterjee, Ayan and Avazov, Nurilla},
  journal={Scientific Reports},
  year={2026},
  publisher={Nature Publishing Group UK London}
}






\end{document}